\PassOptionsToPackage{
  colorlinks=true,
  linkcolor=vtProp,
  citecolor=vtString,
  urlcolor=vtLiteral,
  filecolor=vtLiteral,
  anchorcolor=vtProp
}{hyperref}
\documentclass[preprint,3p,nonatbib]{elsarticle}
\makeatletter
\let\c@author\relax
\makeatother
\usepackage{geometry}

\usepackage{xcolor}
\definecolor{darkolive}      {HTML}{393a34}  
\definecolor{offwhite}      {HTML}{faf9f5}  
\definecolor{sage} {HTML}{248459}  
\definecolor{olive}{HTML}{6a8c1f  }  
\definecolor{terracotta}  {HTML}{c67b5d}  
\definecolor{warmgrey} {HTML}{998f84}  
\definecolor{teal} {HTML}{3a9c9b}  
\definecolor{rose} {HTML}{a14f55}  
\definecolor{sienna}    {HTML}{ad502b}  
\definecolor{siennasatured}{HTML}{bd491b}
\definecolor{parchment}  {HTML}{f0ece3}  
\definecolor{tan}    {HTML}{c8bfb0}  
\definecolor{teal2}{HTML}{217169}
\definecolor{darkpink}{HTML}{a9305a}
\definecolor{darkpurple}{HTML}{8721ba}
\usepackage{hyperref}
\hypersetup{
  colorlinks=true,
}
\usepackage{xurl}
\usepackage{amsmath,amssymb,amsfonts}
\usepackage{mathrsfs}
\usepackage{dsfont}
\usepackage{bm}
\usepackage{adjustbox} 
\usepackage{lipsum}
\usepackage{needspace}
\usepackage{orcidlink}

\newcommand{\orcidauthor}[1]{\hspace{2pt}\orcidlink{#1}}

\usepackage{xcolor}
\definecolor{vtFg}      {HTML}{393a34}  
\definecolor{vtBg}      {HTML}{faf9f5}  
\definecolor{vtKeyword} {HTML}{248459}  
\definecolor{vtFunction}{HTML}{6a8c1f}  
\definecolor{vtString}  {HTML}{c67b5d}  
\definecolor{vtComment} {HTML}{998f84}  
\definecolor{vtLiteral} {HTML}{3a9c9b}  
\definecolor{vtDeleted} {HTML}{a14f55}  
\definecolor{vtProp}    {HTML}{ad502b}  
\definecolor{vtHeader}  {HTML}{f0ece3}  
\definecolor{vtRule}    {HTML}{c8bfb0}  

\definecolor{darkgreen}{HTML}{3c5800}

\colorlet{kwBlue}  {vtKeyword}
\colorlet{kwTeal}  {vtFunction}
\colorlet{kwAmber} {vtString}
\colorlet{kwGray}  {vtComment}
\colorlet{indOne}  {vtProp}
\colorlet{indTwo}  {vtLiteral}
\colorlet{indThree}{vtDeleted}
\colorlet{algHeader}{vtHeader}

\usepackage{tikz}              
\usepackage{fontspec}          
\usepackage{algorithm}
\usepackage[
  indLines       = true,
  noEnd          = true,
  italicComments = true,
  rightComments  = true,
  commentColor   = vtComment,
]{algpseudocodex}

\newfontfamily\juliamono{JuliaMono}[
  Path = ./JuliaMono/,       
  Extension = .ttf,          
  UprightFont = *-Regular,   
  BoldFont = *-Bold,
  FontFace = {mb}{n}{*-Medium},   
  FontFace = {sb}{n}{*-SemiBold},
  FontFace = {bold}{n}{*-Bold},
  FontFace = {black}{n}{*-Black},
  FontFace = {regular}{n}{*-Regular},
  FontFace = {exbold}{n}{*-ExtraBold},
  Scale = 1.0
]

\newcommand{\juliasemibold}{\juliamono\fontseries{sb}\selectfont}

\newcommand{\juliaextrabold}{\juliamono\fontseries{exbold}\selectfont}

\makeatletter
\tikzset{
  algpxIndentLine/.style={
    draw,
    line width=0.55pt,
    /utils/exec={%
      \ifnum\value{ALG@nested}=1 \pgfkeysalso{color=indOne!60}%
      \else\ifnum\value{ALG@nested}=2 \pgfkeysalso{color=indTwo!60}%
      \else \pgfkeysalso{color=indThree!60}%
      \fi\fi
    }
  }
}
\makeatother

\algrenewcommand\algorithmicif      {\textcolor{vtKeyword} {\textbf{if}}}
\algrenewcommand\algorithmicthen    {\textcolor{vtKeyword} {\textbf{then}}}
\algrenewcommand\algorithmicelse    {\textcolor{vtKeyword} {\textbf{else}}}
\algrenewcommand\algorithmicfor     {\textcolor{vtKeyword} {\textbf{for}}}
\algrenewcommand\algorithmicforall  {\textcolor{vtKeyword} {\textbf{for all}}}
\algrenewcommand\algorithmicdo      {\textcolor{vtKeyword} {\textbf{do}}}
\algrenewcommand\algorithmicwhile   {\textcolor{vtKeyword} {\textbf{while}}}
\algrenewcommand\algorithmicrepeat  {\textcolor{vtKeyword} {\textbf{repeat}}}
\algrenewcommand\algorithmicuntil   {\textcolor{vtKeyword} {\textbf{until}}}
\algrenewcommand\algorithmicreturn  {\textcolor{vtProp}  {\textbf{return}}}
\algrenewcommand\algorithmicoutput  {\textcolor{vtString}  {\textbf{output}}}
\algrenewcommand\algorithmicprocedure{\textcolor{vtFunction}{\textsc{procedure}}}
\algrenewcommand\algorithmicfunction {\textcolor{vtFunction}{\textsc{function}}}

\newcommand{\proc}[1]{\textnormal{\textcolor{vtFunction}{\juliaextrabold #1}}}

\newcommand{\proctwo}[1]{\textnormal{\textcolor{black}{\small\juliasemibold #1}}}

\newcommand{\kw}  [1]{\textcolor{vtKeyword}{\textbf{#1}}}

\usepackage{booktabs}
\usepackage{tabularx}
\newenvironment{algIO}{%
  \vspace{3pt}%
  \small\color{vtFg}%
  \setlength{\tabcolsep}{4pt}%
  \tabularx{\dimexpr\linewidth-8pt\relax}{@{\hspace{4pt}} l @{\hspace{6pt}} l @{\hspace{4pt}---\hspace{6pt}} X @{}}%
}{%
  \endtabularx%
  \vspace{2pt}%
  \par\noindent\textcolor{vtRule}{\rule{\linewidth}{0.4pt}}%
  \vspace{3pt}%
}
\newcommand{\algRequire}[2]{\textcolor{vtProp}{\sffamily\bfseries Input:} & $#1$ & #2 \\[1pt]}
\newcommand{\algEnsure}[2]{\textcolor{vtProp}{\sffamily\bfseries Output:} & $#1$ & #2 \\[1pt]}

\usepackage{tcolorbox}
\tcbuselibrary{breakable, skins}

\newcounter{algbox}

\newenvironment{algbox}[2]{%
  \refstepcounter{algbox}%
  \begin{tcolorbox}[
    enhanced,
    colframe         = vtRule,
    boxrule          = 0.6pt,
    arc              = 3pt,
    borderline north = {3pt}{0.5pt}{darkgreen},
    title = {%
      \vspace{6pt}%
      {\normalsize\sffamily\bfseries\textcolor{darkgreen}{Algorithm~\thealgbox}}%
      \enspace{\textcolor{vtRule}{\large\textbar}}\enspace
      {\normalsize\juliaextrabold\textcolor{vtFunction}{{#1}}}%
      \hfill
      {\normalsize\sffamily\textcolor{vtFunction}{{#2}}}%
      \vspace{3pt}%
    },
    coltitle         = vtFg,
    colbacktitle     = vtHeader,
    titlerule        = 0.4pt,
    titlerule style  = {vtRule},
    colback          = white,
    top              = 4pt,
    bottom           = 6pt,
    left             = 6pt,
    right            = 6pt,
  ]%
}{%
  \end{tcolorbox}%
}

\newcounter{example}

\definecolor{vtKeyword2} {HTML}{6d3b5e}  

\definecolor{lightred}{HTML}{c67b5d}
\definecolor{darkred}{HTML}{ad502b}

\newenvironment{example}[1]{%
  \refstepcounter{example}%
  
  \begin{tcolorbox}[
    enhanced,
    colframe         = vtRule,
    boxrule          = 0.6pt,
    arc              = 3pt,
    borderline north = {3pt}{0.5pt}{darkred},
    title = {%
      \vspace{6pt}%
      {\normalsize\sffamily\bfseries\textcolor{darkred}{Example~\theexample}}%
      \enspace{\textcolor{vtRule}{\large\textbar}}\enspace
      {\normalsize\sffamily\textcolor{lightred}{ #1 }}%
      \vspace{3pt}%
    },
    colbacktitle     = vtHeader,
    titlerule        = 0.4pt,
    titlerule style  = {vtRule},
    colback          = white,
    top              = 8pt,
    bottom           = 6pt,
    left             = 6pt,
    right            = 6pt,
  ]%
}{%
  \end{tcolorbox}
  \vspace{0.5em}
}

\newcounter{thtable}
\newenvironment{thtable}[2]{%
  \refstepcounter{thtable}%
  
  \begin{tcolorbox}[
    enhanced,
    colframe         = vtRule,
    boxrule          = 0.6pt,
    arc              = 3pt,
    borderline north = {3pt}{0.5pt}{teal2},
    title = {%
      \vspace{6pt}%
      {\normalsize\sffamily\bfseries\textcolor{teal2}{Table~\thethtable}}%
      \enspace{\textcolor{vtRule}{\large\textbar}}\enspace
      {\normalsize\sffamily\textcolor{teal}{#1}}%
      \hfill
      {\normalsize\sffamily\textcolor{teal}{#2}}%
      \vspace{3pt}%
    },
    coltitle         = vtFg,
    colbacktitle     = vtHeader,
    titlerule        = 0.4pt,
    titlerule style  = {vtRule},
    colback          = white,
    top              = 4pt,
    bottom           = 6pt,
    left             = 6pt,
    right            = 6pt,
  ]%
  \small\color{vtFg}%
  \setlength{\tabcolsep}{5pt}%
  \renewcommand{\midrule}  {%
    \arrayrulecolor{vtRule}\hline\arrayrulecolor{vtFg}}%
  \centering
}{%
  \end{tcolorbox}%
}

\newcommand{\colhead}[1]{%
  \textcolor{teal2}{\sffamily #1}}

\usepackage{changepage}

\definecolor{vtFigAccent}{HTML}{3d5a8a}
\definecolor{vtFigLight} {HTML}{6b86b5}

\newcounter{figbox}

\newlength{\figboxmargin}
\newenvironment{figbox}[2]{%
  \refstepcounter{figbox}%
  \begin{adjustwidth}{-\figboxmargin}{-\figboxmargin}%
  \begin{tcolorbox}[
    enhanced,
    colframe         = vtRule,
    boxrule          = 0.6pt,
    arc              = 3pt,
    borderline north = {3pt}{0.5pt}{vtFigAccent},
    title            = {%
      \vspace{6pt}%
      {\normalsize\sffamily\bfseries\textcolor{vtFigAccent}{Solution~\thefigbox}}%
      \enspace{\textcolor{vtRule}{\large\textbar}}\enspace
      {\normalsize\sffamily\textcolor{vtFigLight}{#1}}%
      \vspace{3pt}%
      \hfill
      {\normalsize\sffamily\textcolor{vtFigLight}{#2}}%
      \vspace{0pt}%
    },
    colbacktitle     = vtHeader,
    titlerule        = 0.4pt,
    titlerule style  = {vtRule},
    colback          = white,
    top              = 8pt,
    bottom           = 6pt,
    left             = 6pt,
    right            = 6pt,
  ]%
  \centering
}{%
  \end{tcolorbox}%
  \end{adjustwidth}%
  \vspace{0.5em}%
}

\usepackage{graphicx}
\usepackage{subcaption}
\usepackage{caption}
\usepackage{booktabs}
\usepackage{multirow}
\usepackage{array}
\usepackage{colortbl}
\usepackage{diagbox}
\usepackage{fontspec}
\newfontfamily\ipafont{CharisSIL}[
  Path = ./CharisSIL/,
  Extension = .ttf,
  UprightFont = *-Regular,
]

\usepackage{siunitx}
\usepackage{tcolorbox}
\tcbuselibrary{skins,theorems}
\usepackage[
  backend=biber,
  style=alphabetic,
  sorting=nyt,
  maxalphanames=1,
  minalphanames=1,
  maxbibnames=99,
  doi=true,
  url=true,
  eprint=false
]{biblatex}
\DeclareFieldFormat{doi}
  {\mkbibacro{DOI}\addcolon\space\href{https://doi.org/#1}{#1}}
\DeclareFieldFormat{url}{\url{#1}}
\usepackage[nameinlink]{cleveref}

\crefname{figbox}{Solution}{Solutions}
\crefalias{algbox}{algorithm}
\crefname{algorithm}{Algorithm}{Algorithms}

\crefname{example}{Example}{Examples}
\crefname{section}{Section}{Sections}
\crefname{subsection}{Section}{Sections}
\crefname{subsubsection}{Section}{Sections}
\crefname{appendix}{Appendix}{Appendices}
\crefname{figure}{Figure}{Figures}
\crefname{table}{Table}{Tables}
\crefname{equation}{Equation}{Equations}

\DeclareCaptionType[within=none]{figboxcap}[Solution][List of Solutions]

\newcommand{\bmtilde}[1]{\tilde{\bm{#1}}}

\newcommand{\Tp}[1]{{#1}^{\mathsf{T}}}
\DeclareMathOperator*{\argmin}{arg\,min}

\DeclareFieldFormat{doi}{%
  \mkbibacro{DOI}\addcolon\space
  \href{https://doi.org/#1}{\nolinkurl{#1}}}
  
\AtBeginDocument{
  \hypersetup{
    colorlinks=true,
    linkcolor=siennasatured,
    citecolor=darkpink,
    urlcolor=siennasatured,
    filecolor=offwhite,
    anchorcolor=offwhite
  }
}

\let\oldhref\href
\renewcommand{\href}[2]{\oldhref{#1}{{\fontseries{b}\selectfont#2}}}

\begin{document}
 
\begin{frontmatter}
 
\title{An Optimisation Framework for the Well-Conditioned Training of Physics-Informed Neural Networks}

\author[inst1]{Joseph Webb\corref{cor1}\orcidauthor{0000-0002-3473-5625}}
\ead{joseph.webb@worc.ox.ac.uk}
\author[inst1]{Sadok Jerad\orcidauthor{0000-0002-4892-0514}}
\ead{sadok.jerad@maths.ox.ac.uk}
\author[inst1]{Coralia Cartis\orcidauthor{0000-0002-0963-5550}}
\ead{coralia.cartis@maths.ox.ac.uk}

\cortext[cor1]{Corresponding author}

\affiliation[inst1]{organisation={Mathematical Institute, University of Oxford},
             city={Oxford},
             country={United Kingdom}}

\begin{abstract}
Physics-informed neural networks (PINNs) have emerged as a promising route to solve partial differential equations, 
yet they have struggled to reach the precision of classical solvers.
The obstacle is increasingly understood to be one of optimisation, owing to the severely ill-conditioned loss landscape. We present \proctwo{DSGNAR}: Doubly-Sketched Gauss--Newton with Adaptive Ratio, a scalable second-order optimisation framework that confronts this ill-conditioning and, in doing so, obtains unprecedented accuracy and speed.
\proctwo{DSGNAR} couples a doubly-sketched Gauss--Newton model with a novel strategy that carefully controls both regularisation and step length. 
Across a suite of problems spanning nonlinear, chaotic, multi-scale, high-dimensional, and Navier--Stokes, the framework greatly improves on the state of the art: able to attain relative $\ell_2$ errors as low as $3\times10^{-16}$ in double precision, improve contemporary results by five orders of magnitude on the canonical Burgers' equation, and as much as eight orders on a high-dimensional Poisson problem, while remaining markedly faster. We further show that, in single precision, solutions at the limit of round-off error can be obtained very quickly: Burgers' equation to $\ell_2^{\text{rel}} = 4.75 \times 10^{-7}$ in under ten seconds. The framework is also robust to the choice of architecture, arithmetic precision, and initial hyperparameters.

The code is available at \url{https://www.github.com/wephy/physics-informed-neural-networks}
\end{abstract}


\end{frontmatter}

\section{Introduction}\label{sec:intro}
Physics-informed neural networks (PINNs) have become one of the most active paradigms in scientific machine learning, offering a promising route to solving partial differential equations (PDEs). In the few years since their modern revival, they have found use across many fields of science and engineering: biomedicine, chemical engineering, dynamical systems, finance, fluid and solid mechanics, and geophysics \cite{Toscano2025FromLearning}. There are many motivations for PINNs, among them: (1) their \textit{mesh-free} approach, which sidesteps mesh generation, irregular geometries, the stability restrictions of finite-difference and finite-element schemes, and the curse of dimensionality \cite{Frey2000MeshGeneration}; (2) their \textit{ease of use}, since a problem can be solved purely by stating the governing equations, with little of the bespoke machinery a traditional solver demands; and (3) their \textit{natural data-assimilation}, fusing sparse or noisy observations with the governing equations and making them well-suited to inverse problems where the physics is only partially known \cite{Karniadakis2021Physics-informedLearning}.

The idea of casting a system of physical laws as an unconstrained minimisation problem dates to \textcite{Dissanayake1994Neural-Network-BasedEquations} and \textcite{Lagaris1998ArtificialEquations}; later, with the advent of modern machine learning tools and ideas, it was refined and popularised by \textcite{Raissi2019Physics-InformedEquations}. PINNs solve PDEs by approximating the solution with a neural network and iteratively evaluating the current violation of the governing equations at sampled points; using automatic differentiation for the required derivatives, steps are then taken to minimise these violations until the network satisfies the problem.

\begin{figure}[t]
    \centering
    \hspace{-1em}
    \includegraphics[scale=0.85]{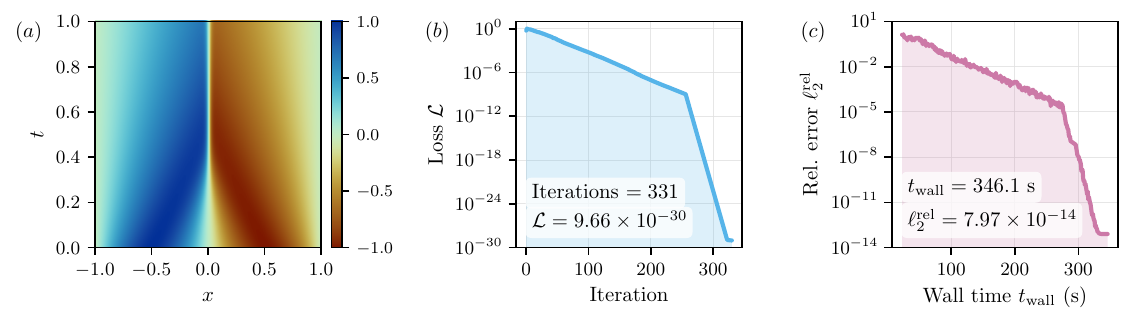}
    \caption{\textbf{A representative high-accuracy solve with our framework.}
    The viscous Burgers' equation (\cref{app:burgers_eq}) solved in double precision
    with a SIREN of $d_{\bm\theta}=11{,}285$ trainable parameters and sketch size $s=4{,}000$.
    \textbf{(a)} The recovered solution $u_{\bm\theta}(x,t)$, including the sharp internal
    layer that forms near $x=0$. \textbf{(b)} The training loss $\mathcal{L}$, driven below
    $10^{-29}$ in $331$ iterations. \textbf{(c)} The relative $\ell_2$ error against the
    reference solution, reaching $\ell_2^{\text{rel}}=7.97\times10^{-14}$ in $346.1$ seconds
    on a single NVIDIA H100 GPU\@ -- several orders of magnitude beyond the best previously
    reported accuracy for this benchmark (\cref{tab:pde_results}).}
    \label{fig:summary}
\end{figure}

PINNs are often regarded as a leading candidate for the next generation of scientific computing, yet it is increasingly clear from the existing literature that they have yet to truly rival traditional numerical methods in many areas. The major hurdle has been the inherent ill-conditioning of PINNs, detailed in \textcite{Rathore2024ChallengesPerspective}, alongside a growing consensus that the bottleneck lies in the optimiser \cite{Kiyani2025OptimizingNetworks}. PINNs differ considerably from traditional deep learning applications such as computer vision (CV) or natural language processing (NLP): their networks are typically far smaller -- tens of thousands of parameters, as seen in \cref{tab:pde_results}, against the millions or billions used in CV, NLP, and large language models (LLMs) -- yet, unlike these fields, PINNs demand high precision to be competitive with established classical solvers. It is worth noting that, although branded as \textit{machine learning}, PINNs can exist independent of real data, relying only on their own residual evaluations; a consequence of this is that overfitting is not a genuine concern, and that losses at the limit of machine precision are theoretically possible, and indeed welcome. Standard first-order optimisers (e.g. \cite{Kingma2015Adam:Optimization}) therefore often prove insufficient for training PINNs \cite{Rathore2024ChallengesPerspective,Urban2025UnveilingBe}.

Current state-of-the-art methods typically employ second-order techniques that approximate the Hessian. One line of research enhances quasi-Newton methods (such as BFGS \cite{Nocedal2006NumericalOptimization}) with additional scaling factors \cite{Al-Baali1998NumericalAlgorithms}; an effective hybrid strategy combines an initial phase of Adam with a later switch to a quasi-Newton variant, proving reliable across a broad spectrum of problems \cite{Kiyani2025OptimizingNetworks}. Other methods employ Kronecker-Factored Approximate Curvature (KFAC) \cite{Dangel2024Kronecker-FactoredNetworks,Wang2025GradientPerspective}, building on the seminal work of \textcite{Martens2015OptimizingCurvature,Martens2018Kronecker-FactoredNetworks}. A third line, closest to ours, adopts a Gauss--Newton framework \cite{Jnini2026Gauss-NewtonDynamics,Guzman-Cordero2025ImprovingRandomization,Muller2023AchievingDescent}, carefully deriving matrix-free implementations to reduce the cost of direct Jacobian calculations; for the PINN least-squares objective, the energy natural gradient in fact coincides with a Gauss--Newton step \cite{Muller2023AchievingDescent,Jnini2026Gauss-NewtonDynamics}. Tools from randomised linear algebra, sketching in particular \cite{Martinsson2020RandomizedAlgorithms,Halko2011FindingDecompositions}, are also applied to avoid ever forming the full Gauss--Newton matrix \cite{Rathore2024ChallengesPerspective,BestMckay2025Near-OptimalNetworks,Yang2022Sketch-BasedLearning}. Beyond the optimiser itself, the choice of globalisation scheme -- line search \cite{Armijo1966MinimizationDerivatives}, trust-region \cite{Conn2000TrustMethods}, or Levenberg--Marquardt \cite{Levenberg1944ASquares,Marquardt1963AnParameters} -- significantly affects the results \cite{Kiyani2025OptimizingNetworks}, as does the multi-scale nature of the PINN loss, whose dynamics, boundary, and initial-condition terms occupy disparate scales that must be balanced \cite{Wu2026ALearning,Anagnostopoulos2024Residual-BasedNetworks}.

The success of the Gauss--Newton method for nonlinear least squares and other supervised-learning contexts, alongside its relatively low-cost second-order approximation, motivates further exploration in the field of PINNs. We propose \proctwo{DSGNAR}\footnote{Code available at \url{https://www.github.com/wephy/physics-informed-neural-networks}} ({\ipafont /dɪˈzaɪnə/}, like ``designer''), a novel doubly-sketched Gauss--Newton method that alleviates the ill-conditioning of the PINN objective and permits aggressive sampling of points. It is further combined with a sophisticated regularisation scheme that aims to achieve richer exploration of the loss landscape. We summarise our contributions below.

\begin{figure}[t]
\begin{thtable}{Results}{Accuracies and runtimes for comprehensive suite of PDEs}
\begin{tabular}{l l l r l r l r r}
\multirow{2}{*}{\colhead{PDE}}
  & \multicolumn{3}{c}{\colhead{Benchmark}}
  & \multicolumn{2}{c}{\colhead{Ours}}
  & \multicolumn{3}{c}{\colhead{Architecture}} \\
\cmidrule(lr){2-4}\cmidrule(l){5-6}\cmidrule(lr){7-9}
  & \colhead{Ref.} & \colhead{$\ell_2^{\text{rel}}$ $\downarrow$} & \colhead{$t_{\text{wall}}$ (s) $\downarrow$}
  & \colhead{$\ell_2^{\text{rel}}$ $\downarrow$} & \colhead{$t_{\text{wall}}$ (s) $\downarrow$}
  & \colhead{Prec.} & \colhead{Params} & \colhead{Sketch $s$} \\[4pt]
\midrule\addlinespace[2pt]\midrule\addlinespace[4pt]
\multirow{2}{*}{\textbf{Burgers}}
  & \multirow{2}{*}{\cite{Kiyani2025OptimizingNetworks}} & $4.04\times10^{-5}$ & $179$
  & $\mathbf{4.75\times10^{-7}}$ & $\mathbf{9.8}$
  & Single & $1{,}447$ & $700$ \\[4pt]
  & & $1.62\times10^{-8}$ & $2{,}878$
  & $\mathbf{7.97\times10^{-14}}$ & $\mathbf{346.1}$
  & Double & $11{,}285$ & $4{,}000$ \\[4pt]
\midrule\addlinespace[4pt]
\textbf{KS}
  & \cite{Kiyani2025OptimizingNetworks} & $6.51\times10^{-4}$ & $130{,}222$
  & $\mathbf{5.12\times10^{-7}}$ & $\mathbf{5{,}226.6}$
  & Double & $15{,}265$ & $5{,}000$ \\[4pt]
\midrule\addlinespace[4pt]
\textbf{10D Poisson}
  & \cite{Guzman-Cordero2025ImprovingRandomization} & $\hspace{0.55em}{\sim}1\times10^{-5}$ & ${\sim}8{,}000$
  & $\mathbf{5.96\times10^{-12}}$ & $\mathbf{3{,}239.7}$
  & Double & $17{,}241$ & $5{,}000$ \\[4pt]
\midrule\addlinespace[4pt]
\multirow{2}{*}{\textbf{Navier--Stokes}}
  & \multirow{2}{*}{\cite{Chiu2026Scale-PINN:Correction}} & \multirow{2}{*}{$1.43\times10^{-2}$} & \multirow{2}{*}{${\sim}90$}
  & $\mathbf{1.13\times10^{-4}}$ & $402.9$
  & Double & $9{,}219$ & $4{,}000$ \\[4pt]
  & & &
  & $\mathbf{6.34\times10^{-4}}$ & $\mathbf{80.4}$
  & Double & $9{,}219$ & $1{,}000$ \\[4pt]
\midrule\addlinespace[2pt]\midrule\addlinespace[4pt]
\multirow{2}{*}{\textbf{Wave}}
  & \multirow{2}{*}{\cite{Dai2026TINNs:PDEs}} & \hfill $-$ & \hfill $-$
  & $\mathbf{5.41\times10^{-7}}$ & $\mathbf{70.1}$
  & Single & $5{,}085$ & $2{,}000$ \\[4pt]
  & & $6.71\times10^{-6}$ & $2{,}772$
  & $\mathbf{1.20\times10^{-15}}$ & $\mathbf{635.4}$
  & Double & $11{,}225$ & $4{,}000$ \\[4pt]
\midrule\addlinespace[4pt]
\multirow{2}{*}{\textbf{KdV}}
  & \multirow{2}{*}{\cite{Dai2026TINNs:PDEs}} & \hfill $-$ & \hfill $-$
  & $\mathbf{9.55\times10^{-7}}$ & $\mathbf{275.3}$
  & Single & $5{,}121$ & $3{,}000$ \\[4pt]
  & & $1.53\times10^{-4}$ & $2{,}412$
  & $\mathbf{8.24\times10^{-11}}$ & $\mathbf{1{,}209.7}$
  & Double & $11{,}285$ & $4{,}000$ \\[4pt]
\midrule\addlinespace[4pt]
\multirow{2}{*}{\textbf{Multi-scale}}
  & \multirow{2}{*}{\cite{Anderson2026ELM-FBPINNs:Method}} & \hfill $-$ & \hfill $-$ 
  & $\mathbf{4.83\times10^{-7}}$ & $\mathbf{46.5}$
  & Single & $3{,}215$ & $1{,}200$ \\[4pt]
  & & \hspace{0.55em}${\sim}1\times10^{-3}$ & $\textbf{62}$
  & $\mathbf{4.46\times10^{-14}}$ & $281.0$
  & Double & $11{,}825$ & $4{,}000$ \\[4pt]
\midrule\addlinespace[4pt]
\textbf{5D Poisson}
  & \cite{Guzman-Cordero2025ImprovingRandomization} & $\hspace{0.55em}{\sim}1\times10^{-7}$ & ${\sim}7{,}000$
  & $\mathbf{3.03\times10^{-16}}$ & $\mathbf{406.7}$
  & Double & $13{,}313$ & $4{,}000$ \\[4pt]
\end{tabular}
\captionof{table}{%
  Comparison of our framework against state-of-the-art benchmarks across a suite of PDEs.
  The upper group lists the four problems discussed in the main text; the lower group the
  further benchmarks reported in \cref{app:extra_results}.
  \textbf{Benchmark} and \textbf{Ours} columns report the relative $\ell_2$ error,
  $\ell_2^{\text{rel}}$, and wall-clock training time, $t_{\text{wall}}$, in seconds.
  Bold entries indicate the better result in each metric pair.
  Benchmark figures are taken from the cited contemporaneous works. \textbf{Architecture} columns report the floating-point precision (Single/Double),
  the number of trainable parameters, and the sketch size $s$ used during training.
  All experiments in \textbf{Ours} were run on a single NVIDIA H100 GPU\@, and benchmark results report the $t_\text{wall}$ values from the respective papers. Architectures were chosen mostly with a focus on $\ell_2^{\text{rel}}$ error, but owing to the interplay between accuracy and $t_\text{wall}$, more performance can be gained in either direction.
}
\label{tab:pde_results}
\end{thtable}
\end{figure}

\begin{itemize}
    \item \textbf{A conditioning-first principle for the step.} We argue that a PINN's accuracy ceiling is set by how well-conditioned the final region of the iterate is, tured by the \emph{effective regularisation} -- the amount of regularisation a Gauss--Newton step needs before it yields a decrease. We turn this into a concrete rule by letting \textit{both} the regularisation and step length be implicit, fixing each step by comparing the objective against its Gauss--Newton model to hit a target decrease ratio. This target is chosen dynamically, promoting early exploration before shifting to aggressive descent once minimal regularisation is achieved (\cref{sec:overview}).

    \item \textbf{A doubly-sketched Gauss--Newton model.} We sketch the Jacobian on both sides, using CountSketch \cite{Charikar2004FindingStreams} on the rows and a subsampled randomised cosine transform \cite{Martinsson2020RandomizedAlgorithms,Halko2011FindingDecompositions} on the columns. In contrast to the matrix-free line of work \cite{Jnini2026Gauss-NewtonDynamics,Guzman-Cordero2025ImprovingRandomization}, we build the sketched Jacobian explicitly, doing so quickly and memory-efficiently via the proposed methods (\cref{sec:build_jac}). The resulting approximation is small, so its \proctwo{SVD} yields inexpensive candidate steps.
    
    \item \textbf{Cheap, dense residual sampling.} As the Jacobian sketch is \textit{accumulated}, notably over low-memory batches, we are able to sample collocation points far more aggressively than is typically feasible (\cref{tab:sketch_timings}). This sidesteps the residual-selection problem that is prevalent in the PINN literature \cite{Wu2023ANetworks}.

    \item \textbf{Comprehensive, state-of-the-art results.} We show that \proctwo{DSGNAR} solves a large suite of PDE problems to precisions that rival classical solvers, in few iterations and little compute time. We provide extensive and highly detailed numerics, outperforming state-of-the-art PINNs by orders of magnitude across nonlinear, chaotic, multi-scale, high-dimensional, and Navier--Stokes problems, while remaining robust to architecture, precision, and initial hyperparameters. \cref{fig:summary} previews a representative solve, and \cref{tab:pde_results} presents the full set of results.
\end{itemize}

The rest of the paper is organised as follows. \cref{sec:pinn_setup} formulates the PINN problem as a weighted sum of squares and introduces the running example. \cref{sec:gn_background} recalls the Gauss--Newton method, along with the Levenberg--Marquardt and trust-region machinery we build on, situating our approach in the literature. \cref{sec:method} sets out the optimisation philosophy, how we build the doubly-sketched Jacobian, solve the subproblem through a single \proctwo{SVD}, and select each step to obtain the target decrease ratio. \cref{sec:impdetails} gives the implementation details, \cref{sec:experiments} reports results across the suite of problems in \cref{tab:pde_results}, and \cref{sec:conclusion} concludes with discussion. The appendices collect the remaining algorithms, architectural details, problem definitions, and the full solution figures, showcasing all available metrics from training the PINNs.

\section{Background}
\label{sec:background}
\subsection{Problem setup: a physics-informed neural network}
\label{sec:pinn_setup}

For our purposes, we consider an initial-boundary value problem on a space-time domain $\Omega \times [0, T]$ as a triplet of operators $(\bm{\mathcal{P}}, \bm{\mathcal{I}}, \bm{\mathcal{B}})$ acting on a solution $\bm{u}$ in a suitable function space $\mathcal{V}$. For a system of $L$ equations, the problem is expressed as:
\begin{align}
    \bm{\mathcal{P}}\bm{u} &= \bm{0}, & (\bm{x}, t) &\in \Omega \times [0, T], \label{eq:pde} \\
    \bm{\mathcal{I}}\bm{u} &= \bm{0}, & (\bm{x}, t) &\in \Omega \times \{0\}, \label{eq:ic} \\
    \bm{\mathcal{B}}\bm{u} &= \bm{0}, & (\bm{x}, t) &\in \partial\Omega \times [0, T], \label{eq:bc}
\end{align}
with $\bm{u}: \Omega \times [0, T] \to \mathbb{R}^L$. By defining a composite operator $\bm{\mathcal{F}}$:
\begin{equation}
    \bm{\mathcal{F}}\bm{u} = \begin{pmatrix}\bm{\mathcal{P}}\bm{u} \\
    \bm{\mathcal{I}}\bm{u} \\
    \bm{\mathcal{B}}\bm{u}
\end{pmatrix},\end{equation}
the solution $\bm{u}$ satisfies the operator identity $\bm{\mathcal{F}}\bm{u} = \bm{0}$, namely, $\bm{u} \in \ker(\bm{\mathcal{P}}) \cap \ker(\bm{\mathcal{I}}) \cap \ker(\bm{\mathcal{B}})$.
Physics-informed neural networks (PINNs) aim to approximate the solution with a neural network $\bm{u}_{\bm\theta}$ by solving $\bm{\mathcal{F}} \bm{u}_{\bm\theta}(\bm{x}, t) = \bm{0}$, where $\bm{\theta} \in \mathbb{R}^{d_{\bm\theta}}$ collects \emph{all} of the network's trainable parameters, flattened into a single vector of dimension $d_{\bm\theta}$. These are precisely the quantities the optimiser updates, while the inputs $(\bm{x},t)$ and the architecture itself are held fixed. The explicit parameter set for each architecture used in this work is stated in \cref{app:architectures}. The training objective is\footnote{The factor $\tfrac12$ is the conventional least-squares scaling.}:
\begin{equation}
    \mathcal{L}(\bm{\theta}) := \dfrac{1}{2} \lVert \bm{\mathcal{F}}\bm{u}_{\bm\theta} \rVert ^2_{\bm{\mathcal{X}}, \bm{w}},
    \tag{PINN objective}
    \label{eq:simple_objective}
\end{equation}
with a weighted $\ell^2$ norm given by:
\begin{equation}
    \lVert \bm{\mathcal{F}}\bm{u}_{\bm\theta} \rVert ^2_{\bm{\mathcal{X}}, \bm{w}} := \sum_{m=1}^{M} \frac{w_m}{|{\mathcal{X}}_m|} \sum_{(\bm{x}, t) \in \mathcal{X}_m}  \big[ \bm{\mathcal{F}}_m\bm{u}_{\bm\theta}(\bm{x}, t) \big]^2.
    \label{eq:objective}
\end{equation}
Here, $M$ is the total number of conditions; each condition $m$ has an associated weight $w_m$ and a set of collocation points $\mathcal{X}_m$ used to compute residuals, giving $N=\sum_m |\mathcal{X}_m|$ collocation points in total. For a system of $L$ equations, we may expect $L$ conditions for each operator, but there may be more; for example, a single PDE that is second-order in time (e.g. the Wave equation) will require two initial conditions, $\bm{\mathcal{I}}_1$ and $\bm{\mathcal{I}}_2$. Fewer conditions are also possible: for incompressible Navier--Stokes, only pressure \textit{gradients} appear, so a condition can be dropped (\cref{ex:multidim}); conditions can also be dropped when they are analytically enforced into the network architecture, referred to as hard constraints (\cref{ex:hard_constraints}). This formulation extends naturally to other problem types, such as time-independent problems, with numerous examples shown later in this paper.
\begin{figure}[t]
\begin{example}{Systems of equations}
\label{ex:multidim}
    When solving a system of equations, the network will have multiple outputs, yielding more conditions to be solved. For example, the steady-state incompressible Navier--Stokes equations require three outputs: the horizontal velocity $u$, the vertical velocity $v$, and the fluid pressure $p$. As the focus is on a steady-state solution, there is no need for initial conditions $\bm{\mathcal{I}}$. Furthermore, as only pressure gradients appear in our equations, we can avoid an associated boundary condition by allowing the pressure field to take an arbitrary constant shift. Namely, we remove the condition which typically \textit{pins} the value of pressure at some point. The resulting $\bm{\mathcal{F}}$ for a particular lid-driven cavity problem has $M=5$ conditions and takes the form:
    \begin{equation*}
\bm{\mathcal{F}} \bm{u} = \left( \begin{array}{l@{\quad}l}
u u_x + v u_y + p_x - \frac{1}{\text{Re}}(u_{xx} + u_{yy}) & \forall (x,y) \in \Omega \\
u v_x + v v_y + p_y - \frac{1}{\text{Re}}(v_{xx} + v_{yy}) & \forall (x,y) \in \Omega \\
u_x + v_y & \forall (x,y) \in \Omega \\
u(x, y) - \mathbb{I}_{\{y=1\}} \cdot 4x(1-x) & \forall (x,y) \in \partial\Omega \\
v(x, y) - 0 & \forall (x,y) \in \partial\Omega
\end{array} \right) = \begin{pmatrix}
{\mathcal{F}}_1 \bm{u} \\
{\mathcal{F}}_2 \bm{u} \\
{\mathcal{F}}_3 \bm{u} \\
{\mathcal{F}}_4 \bm{u} \\
{\mathcal{F}}_5 \bm{u}
\end{pmatrix} = \bm{0}.
\end{equation*}
\end{example}
\end{figure}

\subsection{Gauss--Newton methods and relevant safeguards}
\label{sec:gn_background}
We first recall the Gauss--Newton model and the trust-region/Levenberg--Marquardt machinery, and then situate the method relative to the literature it both draws from and departs from.
Gauss--Newton methods are employed to solve sum-of-squares objectives:
\begin{equation}
    \mathcal{L}(\bm\theta) = \frac{1}{2}\sum_{j=1}^{N}r_j^{2}(\bm\theta),
\end{equation}
where each $r_j: \mathbb{R}^{d_{\bm\theta}} \to \mathbb{R}$ is referred to as a residual, whose value is determined by a given choice of the trainable parameters. We assemble these $N$ residuals into a \textit{residual function}, $\bm{r}: \mathbb{R}^{d_{\bm\theta}} \to \mathbb{R}^N$, with its corresponding Jacobian $\bm{J} \in \mathbb{R}^{N \times {d_{\bm\theta}}}$ defined as:
\begin{equation}
    \bm{r}(\bm\theta) = \begin{pmatrix} r_1(\bm\theta) \\ r_2(\bm\theta) \\ \vdots \\ r_N(\bm\theta) \end{pmatrix},
    \qquad
    \bm{J}(\bm\theta) = \begin{bmatrix}
    \Tp{\nabla r_1(\bm\theta)} \\
    \Tp{\nabla r_2(\bm\theta)} \\
    \vdots \\
    \Tp{\nabla r_N(\bm\theta)}
    \end{bmatrix}.
\end{equation}
The gradient and Hessian of $\mathcal{L}$ are therefore written as:
\begin{equation}
\nabla \mathcal{L}(\bm\theta) = \Tp{\bm{J}(\bm\theta)} \bm{r}(\bm\theta), \qquad
\nabla^2 \mathcal{L}(\bm\theta) = \Tp{\bm{J}(\bm\theta)} \bm{J}(\bm\theta) + \sum_{j=1}^{N} r_j(\bm\theta) \nabla^2 r_j(\bm\theta).
\label{eq:gradandhess}
\end{equation}
The Gauss--Newton approximation replaces the Hessian in \cref{eq:gradandhess} with $\Tp{\bm{J}(\bm\theta)} \bm{J}(\bm\theta)$, discarding the second-order term $\sum_j r_j \nabla^2 r_j$ and retaining only the first-order term, which is positive semi-definite by construction. For the PINN least-squares objective, this model is especially natural, and the energy natural gradient direction of \textcite{Muller2023AchievingDescent} coincides with the parameter-space Gauss--Newton step when the same collocation points are used \cite{Jnini2026Gauss-NewtonDynamics}.
When using a weighted objective, such as the one defined in \cref{eq:objective}, we consider the residual function to absorb the respective weights and normalisations for all of the conditions. However, in practice, it is more convenient to build components from pure, unweighted residuals, $\bm{\mathcal{F}}_m \bm{u}_{\bm{\theta}} (\bm{x}_i, t_i)$. The numerous conditions naturally lead to a structured block Jacobian, where each block $\bm{J}_m \in \mathbb{R}^{|\mathcal{X}_m| \times d_{\bm\theta}}$ has entries:
\begin{equation}
    \left( \bm{J}_m \right)_{i, \cdot} = \sqrt{\frac{w_m}{|\mathcal{X}_m|}} \nabla_{\bm{\theta}} \left[ \bm{\mathcal{F}}_m \bm{u}_{\bm{\theta}} (\bm{x}_i, t_i) \right].
\end{equation}
Similarly, the residual of the $m$-th condition, at the $i$-th collocation point $(\bm{x}_i, t_i) \in \mathcal{X}_m$, is given by:
\begin{equation}
    \left(\bm{r}_m\right)_{i} = \sqrt{\frac{w_m}{|\mathcal{X}_m|}} \bm{\mathcal{F}}_m
    \bm{u}_{\bm{\theta}} (\bm{x}_i, t_i).
\end{equation}
The Levenberg--Marquardt method \cite{Levenberg1944ASquares, Marquardt1963AnParameters} updates parameters $\bm\theta$ using a Gauss--Newton approximation alongside Tikhonov regularisation \cite{Tikhonov1963RegularizationProblems}, yielding an update step:
\begin{equation}
    \bm{p}_k^{\textsf{LM}}(\lambda) = -\left[\Tp{\bm{J}}_k \bm{J}_k + \lambda \bm{I}\right]^{-1} \Tp{\bm{J}}_k \bm{r}_k,
    \label{eq:lm}
\end{equation}
where $\bm{r}_k = \bm{r}(\bm\theta_k)$ and $\bm{J}_k = \bm{J}(\bm\theta_k)$. The regularisation, with an appropriate $\lambda$ update strategy, ensures global convergence to a first-order stationary point from any starting point under standard assumptions \cite{Conn2000TrustMethods}. In contrast, unregularised Gauss--Newton has no such global guarantee; locally it converges quadratically only in the zero-residual limit, degrading to linear convergence for small but non-zero residuals -- at a rate that worsens as the residual at the solution grows -- and may fail to converge altogether for large-residual problems \cite[Ch.~10]{Nocedal2006NumericalOptimization}. Note that the step corresponds to the minimiser of a regularised quadratic model:
\begin{equation}
    m^\mathsf{LM}_k(\bm{p}; \lambda) := \frac{1}{2} \lVert \bm{r}_k + \bm{J}_k \bm{p} \rVert^2 + \frac{\lambda}{2} \lVert \bm{p} \rVert^2.
    \label{eq:LM_model}
\end{equation}
From the trust-region perspective, and by duality of the trust-region subproblem \cite[Chapter~5]{Conn2000TrustMethods}, the step $\bm{p}_k$ is equivalently the minimiser of the unregularised quadratic model, $m^\mathsf{Q}_k(\bm{p}) := \frac{1}{2} \lVert \bm{r}_k + \bm{J}_k \bm{p} \rVert^2$, subject to a constraint on the step length:
\begin{equation}
    \bm{p}_k^{\textsf{TR}}(\Delta) = \argmin_{\bm{p}} m^\mathsf{Q}_k(\bm{p}) \qquad \text{subject to}\ \lVert \bm{p} \rVert \le \Delta.
\end{equation}
The regularisation parameter $\lambda$ here serves as the Lagrange multiplier for the constraint, establishing an implicit mapping $\lambda(\Delta)$, or $\Delta(\lambda)$ in Levenberg--Marquardt, such that $\lambda \cdot \left(\Delta - \lVert \bm{p}_k \rVert\right) = 0$ with $\lambda \geq 0$.
Often, a scale-invariant metric for the quality of a local model is required. A popular choice is the \textit{decrease ratio}, hereafter referred to simply as the \textit{ratio}, considered in various nonlinear optimisation schemes \cite{Conn2000TrustMethods,Nocedal2006NumericalOptimization}, obtained by comparing the actual reduction in the objective to the reduction predicted by the model:
\begin{equation}
   \frac{\mathcal{L}(\bm{\theta}_k) - \mathcal{L}(\bm{\theta}_k + \bm{p}_k)}{\mathcal{L}(\bm{\theta}_k) - m(\bm{p}_k)}.
   \label{eq:true_rho}
\end{equation}
A ratio $\varrho_k \approx 1$ indicates that the model $m$ is a good local proxy for the true objective, and $\varrho_k \approx 0$ indicates that it predicted a significantly larger decrease than was actually achieved. A low decrease ratio can be interpreted as having `trusted' the model over too large a region; in a trust-region method, we would subsequently reduce the radius to ensure sufficient decrease is obtained \cite{Conn2000TrustMethods}.

A direct Levenberg--Marquardt or trust-region solve is, however, out of reach for reasonably sized PINNs: a naive implementation of \cref{eq:lm} costs $\mathcal{O}(N d_{\bm\theta}^2 + d_{\bm\theta}^3)$, and even storing the Gauss--Newton matrix $\Tp{\bm{J}}\bm{J}$ is infeasible at the problem sizes involved (\cref{sec:build_jac}).
Further enhancements of Gauss--Newton have been proposed using momentum and by solving a smaller linear system \cite{Guzman-Cordero2025ImprovingRandomization}. Our work differs from the literature in both our sketching methodology, which acts on both rows and columns (laid out in \proctwo{Sketch}, \cref{alg:sketch}), and a novel method that removes the need to hand-tune the regularisation $\lambda$, deriving it instead via the principled approach detailed in \cref{alg:solve_subproblem}. These methods require further considerations, for which we provide accommodations, such as the a tailored  per iteration condition weight update in \cref{eq:variation_objective} (\cref{alg:update_weights}). Two questions therefore remain, which the rest of the paper answers. \emph{How do we obtain a step cheaply, and in reasonable memory?} We replace $\bm{J}$ with a small, explicitly built sketch, so that a single \proctwo{SVD} cheaply yields the Levenberg--Marquardt step for any desired $\lambda$ -- the subject of \cref{sec:build_jac}
\emph{And how do we choose among the many possible steps?} This is governed by the philosophy we set out next, and realised by the algorithm of \cref{sec:algorithm_overview}. Our trial step is itself a sketched Gauss--Newton step, and so sits within the line of randomised second-order methods initiated by the Newton Sketch of \textcite{Pilanci2017NewtonConvergence}.

\section{Method: the optimisation framework}\label{sec:method}
\subsection{Overview and key ideas}\label{sec:overview}

Our proposed method, \proctwo{DSGNAR}: Doubly-Sketched Gauss--Newton with Adaptive Ratio, is built around a new optimisation philosophy. When using the Gauss--Newton approximation, one traditionally controls the step by choosing either the regularisation parameter $\lambda$ within Levenberg--Marquardt, or the radius $\Delta$ within trust-region. We argue that doing either is insufficient, and can in fact be detrimental in practice. The unintuitive point is that a very ``successful'' step, one that decreases the objective considerably (perhaps due to a good local model, with $\varrho \approx 1$), can directly lead to a poor solution, especially if taken too early, before the conditioning of the problem has been addressed. The loss landscape is complex, non-convex, and high-dimensional, and largely unknown to the optimiser; a large decrease in the objective might therefore \textit{seem} beneficial, yet an optimal (or good) PINN solution may lie far away, and even the best nearby minimum may still be poor. Instead, we propose observing the current \emph{effective regularisation}: what scale of regularisation is required for the associated step, from \cref{eq:lm}, to yield a decrease at all.
The regularisation plays a major role: it simultaneously controls the step length, the interplay between pure Gauss--Newton and gradient descent, and, crucially, the extent to which the minimised objective departs from the true PINN problem. When aiming to solve PDE problems to as high an accuracy as possible, ideally to the limits of machine precision, one cannot afford to solve a shifted objective, in which the fidelity of the underlying physics is compromised. After all, in a PINN problem the governing equations are known exactly, and zero residuals (or as close as the network's expressivity allows) are achievable in principle. To accommodate this perspective, we propose moving away from controlling either $\lambda$ or $\Delta$ directly, and instead analysing a scale-independent quantity: the ratio $\varrho$, defined in \cref{eq:true_rho}. Using this ratio, we can effectively control each step: values in $(0,1)$ can be chosen to determine the step's \emph{aggressiveness}, agnostic to step length, regularisation, and complexities of the current landscape.

In summary, rather than chasing decreases in the objective directly, we navigate the loss landscape by: (1) taking large steps; (2) combating ill-conditioning by remaining in better-behaved regions; and (3) searching for a region capable of solving the true problem as closely as possible. Taking steps as large as possible corresponds to keeping the regularisation as small as possible (\cref{sec:gn_background}). We aim to keep the effective regularisation as small as possible throughout the first stage of training; decreases in it indicate that the inherent conditioning of the problem is becoming less severe. This idea -- to first seek out a region of parameter space where minimal regularisation is required, and only then descend -- manifests as a \textit{two-stage optimisation method}. The first stage adopts a small target ratio, $\varrho^\textsf{Stage 1} \le 0.2$, which is conservative and promotes a decrease in the regularisation; the second stage adopts a larger target, $\varrho^\textsf{Stage 2} \ge 0.5$, which is more aggressive and promotes a decrease in the objective. This leaves two tasks: building a cheap, explicit model for the step, since this philosophy requires repeated access to the \proctwo{SVD} for many candidate $\lambda$ (\cref{sec:build_jac}); and finding, from that model, the step that achieves the desired target ratio (\cref{sec:algorithm_overview}).

\label{sec:algorithm_overview}
We combine the solutions to these tasks, together with supporting routines, into a single algorithm, \proctwo{DSGNAR} (\cref{alg:dsgnar}). Every iteration, the sketch operators are formed, and subsequently used in \proctwo{Sketch} in \cref{alg:sketch}, producing the sketched Jacobian $\tilde{\bm{J}}$, sketched residual $\tilde{\bm{r}}$, and full residual $\bm{r}$ (\cref{sec:sketched_model}). The algorithm then takes the \proctwo{SVD} of $\tilde{\bm{J}}$ and, from it, computes the step whose ratio matches the current target $\varrho^\star$ (\crefrange{sec:subproblem}{sec:finding_step}) via \proctwo{LambdaSolve}. The step is evaluated in the full parameter space and accepted only if it yields a decrease, with the trust-region radius updated accordingly. Training halts once that radius falls below a tolerance $\Delta_{\min}$. Two quantities are then refreshed before the next iteration: the condition weights, via \proctwo{UpdateWeights} (\cref{sec:adaptive_weights}), which keep the residuals of different conditions on a similar scale; and the target ratio, via \proctwo{UpdateTargetRatio} (\cref{sec:stage_switch}), which advances from its first- to second-stage value once the regularisation has bottomed out. We frame \proctwo{DSGNAR} as a trust-region method, despite its non-standard step selection, so as to inherit the associated convergence machinery: a step is taken on success (sufficient decrease); otherwise, the radius centring the next iteration's probes is contracted.

\begin{figure}[t]
\begin{algbox}{DSGNAR: Doubly-Sketched Gauss–Newton with Adaptive Ratio}{
}
\label{alg:dsgnar}
\begin{algIO}
  \algRequire{\bm{\mathcal{F}}}{the $M$ residual conditions of the problem}
  \algRequire{\bm{\mathcal{X}}}{the $M$ sets of collocation points for each condition}
  \algRequire{\bm{u}_{\bm{\theta}}}{the neural network parameterised by $\bm{\theta}$}
  \algRequire{\bm{\theta}_{0},\;\Delta_{0},\;\bm{w}_{0},\;\varrho_{0}}{initial parameters, trust-region radius, weights, and target ratio}
  \algEnsure{\bm{\theta}^{\star}}{the fully trained network parameters}
\end{algIO}
\begin{algorithmic}[1]
  \Statex \Comment{Hyperparameters: sketch rank $s$, convergence tolerance $\Delta_{\min}$}
  \While{$k = 0, 1, \dots$}
  \State $\bm{C},\;\bm{\Omega},\;\bm{S}
         \leftarrow \proctwo{GetSketchOperators}(d_{\bm{\theta}},\,s)$
    \Comment{$\bm{C}\!\in\!\mathbb{R}^{s\times N},\;
              \bm{\Omega}\!\in\!\mathbb{R}^{d_{\bm{\theta}}\times d_{\bm{\theta}}},\;
              \bm{S}\!\in\!\mathbb{R}^{d_{\bm{\theta}}\times s}$}
    \Statex
    \State $\tilde{\bm{J}}_k,\;\tilde{\bm{r}}_k,\;\bm{r}_k
           \leftarrow \proc{Sketch}(\bm{\mathcal{F}},\;\bm{\mathcal{X}},\;
                                    \bm{C},\;\bm{\Omega},\;\bm{S},\;
                                    \bm{u}_{\bm{\theta}_k},\;\bm{\theta}_k,\;\bm{w}_k)$
      \Comment{$\tilde{\bm{J}}_k\!\in\!\mathbb{R}^{s\times s},\;
                \tilde{\bm{r}}_k\!\in\!\mathbb{R}^{s},\;
                \bm{r}_k\!\in\!\mathbb{R}^{N}$}
    \Statex
    \State $\bm{U}_k,\;\bm{\Sigma}_k,\;\bm{V}_k^{\!\top}
           \leftarrow \proctwo{SVD}\!\left(\tilde{\bm{J}}_k\right)$
    \Statex
    \State $\Delta^{\star},\;\lambda_k
           \leftarrow \proc{LambdaSolve}\!\left(
             \bm{\Omega},\;\bm{S},\;
             \bm{U}_k,\;\bm{\Sigma}_k,\;\bm{V}_k^{\!\top},\;
             \tilde{\bm{r}}_k,\;\bm{r}_k,\;
             \bm{u}_{\bm{\theta}},\;\bm{\theta}_k,\;
             \Delta_k,\;\varrho_k\right)$
      \Comment{$\Delta^{\star}\!\in\!\bigl[\tfrac{1}{3}\Delta_k,\;3\Delta_k\bigr]$}
    \Statex
    \State $\tilde{\bm{p}}_k
           \leftarrow
           -\bm{V}_k\;
            \mathrm{diag}\!\left(
              \frac{(\sigma_k)_i}{(\sigma_k)_i^{2}+\lambda_k}
            \right)
            \bm{U}_k^{\!\top}\,\tilde{\bm{r}}_k$
      \Comment{Step in sketch space}
    \State $\bm{p}_k \leftarrow \bm{\Omega}\,\bm{S}\,\tilde{\bm{p}}_k$
      \Comment{Lift to full parameter space}
    \Statex
    \If{$\mathcal{L}_k(\bm{\theta}_k+\bm{p}_k) < \mathcal{L}_k(\bm{\theta}_k)$}
      \State $\bm{\theta}_{k+1} \leftarrow \bm{\theta}_k + \bm{p}_k, \quad
             \Delta_{k+1} \leftarrow \Delta^{\star}$
        \Comment{Accept step and trust-region radius}
    \Else
      \State $\bm{\theta}_{k+1} \leftarrow \bm{\theta}_k, \quad
             \Delta_{k+1} \leftarrow \tfrac{1}{3}\,\Delta_k$
        \Comment{Reject step; shrink trust-region radius}
    \EndIf
    \Statex
    \If{$\Delta_{k+1} < \Delta_{\min}$}
      \State \Return $\bm{\theta}_{k+1}$
        \Comment{Convergence criterion met; terminate}
    \EndIf
    \Statex
    \State $\bm{w}_{k+1}
           \leftarrow \proc{UpdateWeights}(\bm{r}_k,\;\bm{w}_k)$
    \State $\varrho_{k+1}
           \leftarrow \proc{UpdateTargetRatio}\!\left(\{\lambda_i\}_i,\;\varrho_k\right)$
  \EndWhile
\end{algorithmic}
\end{algbox}
\end{figure}

\subsection{The doubly-sketched Gauss--Newton model}
\label{sec:sketched_model}
\paragraph{Novelty} Prior second-order methods for PINNs are largely matrix-free, applying the Gauss--Newton operator only through matrix--vector products \cite{Jnini2026Gauss-NewtonDynamics}. Where such methods do randomise, the sketch acts on the matrix $\bm{J}\bm{J}^\top+\lambda\bm{I}$, as in the randomised Nystr\"om method of \textcite{Guzman-Cordero2025ImprovingRandomization}. We instead form a Jacobian sketch \emph{explicitly}, never storing it in full (though we can, in fact, build it fully at no extra cost if it fits in memory). To sketch the Jacobian, we apply compression on \emph{both} sides: the residual dimension \emph{and} the parameter dimension. The sketch is accumulated over successive row batches (and per problem condition, see \cref{sec:pinn_setup}), which removes memory constraints and makes dense sampling possible, further detailed in \cref{sec:countsketch}.

The residual vector $\bm{r} \in \mathbb{R}^{N}$ and its Jacobian $\bm{J} \in \mathbb{R}^{N \times d_{\bm{\theta}}}$ are the central objects in \proctwo{DSGNAR}: once obtained, they determine the quadratic model in \cref{eq:LM_model} and the Levenberg--Marquardt step in \cref{eq:lm}, for a chosen $\lambda$. For the problem sizes of interest both $N$ and $d_{\bm{\theta}}$ can be large, so even storing the Gauss--Newton matrix $\Tp{\bm{J}}\bm{J}$ is infeasible. We therefore reduce $\bm{J}$ to a small square sketch $\tilde{\bm{J}} \in \mathbb{R}^{s \times s}$ of dimension $s$. A square shape is chosen for the fastest possible \proctwo{SVD}, with added stability gained from aggregating $N$ residuals down to $s \ll N$, which suppresses noise and reflects the over-determination of the system. Solving through the \proctwo{SVD} sidesteps any concerns particular to square systems. Empirically, we observed that taller sketches provided no accuracy benefit, only extra computational cost. The construction of this sketch remains cheap, even under exhaustive dense sampling of residuals (\cref{tab:sketch_timings}).

\paragraph{Choosing the sketch size} The sketch dimension $s$ is the primary control for this framework, trading the fidelity of the Gauss--Newton model against the cost of the \proctwo{SVD}. As a robust default we recommend taking $s$ between a third and a half of the parameter count, $s \in [\,\lfloor d_{\bm\theta}/3\rfloor,\ \lfloor d_{\bm\theta}/2 \rfloor \,]$. Intuitively, $s$ should comfortably exceed the numerical rank of $\bm{J}$, so that the dominant singular values and right singular vectors are faithfully preserved. Values much smaller risk discarding curvature directions, while values approaching $d_{\bm\theta}$ forfeit the speed advantage of sketching for little accuracy gain.
These design choices lead to a scalable framework, as we can avoid any memory issues associated with the sizes of $N$ and $d_{\bm\theta}$, where all costly mechanisms scale with the sketch dimension $s$, which can be chosen independently. Further, experiments in this paper highlight that the proposed framework lends itself to few, highly effective iterations; thus, even with a very large sketch dimension, and a consequentially expensive \proctwo{SVD}, these methods may still be preferred over other optimisers. In \cref{tab:sketch_timings} the computation of full and sketched Jacobians with various methods is explored, highlighting that our proposed method allows large problems to be tackled, where other implementations completely fail.
\begin{figure}
\begin{thtable}{Building full and sketched Jacobians}{Comparison of different methods}
\begin{tabular}{l l l l l l l l l l}
\multicolumn{2}{c}{\colhead{Dimensions}}
  & \multicolumn{2}{c}{\colhead{Na\"ive}}
  & \multicolumn{2}{c}{\colhead{$b' = 4$}}
  & \multicolumn{3}{c}{\colhead{$b' = 4$, $b = 2^{12}$}}
  & \colhead{} \\
\cmidrule(lr){1-2}\cmidrule(lr){3-4}\cmidrule(lr){5-6}\cmidrule(lr){7-9}
  \colhead{$N$} & \colhead{$d_{\bm{\theta}}$}
  & \colhead{fwd} & \colhead{rev}
  & \colhead{fwd} & \colhead{rev}
  & \colhead{fwd} & \colhead{rev}
  & \colhead{\shortstack{CS + rev}} \\[4pt]
\midrule\addlinespace[4pt]
$2^{12}$ & $1{,}447$ & 0.019s & 0.101s & 0.021s & 0.001s & 0.021s & 0.001s & 0.001s \\[4pt]
$2^{14}$ & $1{,}447$ & OOM    & OOM    & 0.077s & 0.002s & 0.080s & 0.003s & 0.003s \\[4pt]
$2^{16}$ & $1{,}447$ & OOM    & OOM    & OOM    & 0.006s & 0.320s & 0.010s & 0.011s \\[4pt]
$2^{18}$ & $1{,}447$ & OOM    & OOM    & OOM    & 0.030s & 1.302s & 0.039s & 0.041s \\[4pt]
$2^{20}$ & $1{,}447$ & OOM & OOM & OOM & OOM & 5.205s & 0.153s & 0.163s \\[4pt]
\midrule\addlinespace[1pt]\midrule\addlinespace[4pt]
$2^{20}$ & $104{,}003$ & OOM    & OOM    & OOM  & OOM & OOM & OOM & 3.127s \\[4pt]
$2^{20}$ & $252{,}503$ & OOM    & OOM    & OOM  & OOM & OOM & OOM & 7.799s \\[4pt]
\end{tabular}
\captionof{table}{%
  Wall-clock time to build Jacobians of size $N \times d_{\bm{\theta}}$ when solving the Burgers equation.
  \textbf{Na\"ive} computes the full Jacobian using \texttt{jax.jacfwd} or \texttt{jax.jacrev};
  $b' = 4$ sub-batches the computation, computing $4$ columns or rows at at time, depending on the autodiff method;
  $b' = 4$, $b = 2^{12}$ additionally adds memory considerations, adding macro-batches; and finally,
  the addition of CountSketch (CS) applies row-compression.
  \textbf{CS + rev} uses our sketch approach detailed in \cref{alg:sketch}, with $\bm{\Omega} = \bm{S} = \bm{I}$,
  combining CountSketch with considered reverse-mode automatic differentiation.
  OOM indicates an out-of-memory failure on the test device.
  All timings were measured on a single NVIDIA H100 GPU\@, averaged over 10 measurements. Results on Jacobians up to a size of $1{,}048{,}576 \times 252{,}503$ (using a sketch size of $s=10{,}000$) show fast computations which avoid memory constraints.
}
\label{tab:sketch_timings}
\end{thtable}
\end{figure}
\subsubsection{CountSketch}
\label{sec:countsketch}
The step $\bm{p}_k$ lives in parameter space, so the quality of a sketch $\tilde{\bm{J}}$ is judged solely by how faithfully it preserves the singular values and \emph{right} singular vectors of $\bm{J}$, which also live in $\mathbb{R}^{d_{\bm\theta}}$ and together determine the step (\cref{sec:subproblem}). The \emph{left} singular vectors live in the residual space $\mathbb{R}^{N}$ and carry no geometric meaning for the step, and no quantity is ever lifted back to $\mathbb{R}^{N}$: the row dimension may therefore be compressed aggressively, by a cheap, non-invertible operation that need not preserve geometry exactly.
Further, we wish to have the property:
\begin{equation}
\label{eq:batchable}
    \bm{C}\bm{J} = \sum_{\text{batches}} \bm{C}\bm{J}_{\text{batch}},
\end{equation}
which is the linearity of the sketch $\bm{C}$. A suitable choice is then CountSketch \cite{Charikar2004FindingStreams}, defined by $K$ independent hash functions $h_k: \{1,\dots,N\} \to \{1,\dots,s\}$, and sign vectors $\bm\varepsilon_k \in \{\pm1\}^{N}$, drawn uniformly at random. The CountSketch matrix $\bm{C} \in \mathbb{R}^{s \times N}$ is:
\begin{equation}
    \bm{C}_{ij} = \dfrac{1}{\sqrt{K}} \displaystyle\sum_{k=1}^{K} (\bm\varepsilon_k)_j\cdot\mathds{1}[h_k(j) = i].
\end{equation}
This has the effect that each row $j$ of $\bm{J}$ is multiplied by $(\bm\varepsilon_k)_j$, and output to row $h_k(j)$. To reduce variance, and following theoretical development \cite{Meister2019TightKernels,Kar2012RandomKernels}, this process happens $K$ times for each row of $\bm{J}$. This has negligible overhead, as $\bm{C}$ is extremely sparse, containing only $K$ non-zeros per column; as such, $\bm{C}$ belongs to the OSNAP family of sparse oblivious subspace embeddings \cite{Nelson2012OSNAP:Embeddings}, with the single-hash case $K=1$ recovering the original CountSketch \cite{Charikar2004FindingStreams,Clarkson2017Low-RankTime}.\footnote{A single hash is an isometry only \emph{in expectation}, $\mathbb{E}[\Tp{\bm{C}}\bm{C}] = \bm{I}_N$, with substantial per-draw variance from hash collisions; such embeddings concentrate around this expectation, preserving the geometry of a fixed $d$-dimensional subspace to relative accuracy $\varepsilon$ with high probability once $K>1$ hashes are used and the sketch has $\mathcal{O}(d/\varepsilon^2)$ rows. While these results do not directly cover our square, doubly-sketched construction (\cref{sec:conclusion}), they motivate both our use of multiple hashes for variance reduction and the expectation that a moderate $s$ suffices.}
In \cref{alg:sketch}, we detail the process by which the sketched Jacobian is constructed. There, the application of $\bm{C}$ occurs on a column-sketched sub-batch Jacobian $\hat{\bm{J}}_{\text{batch}}$ of size $b \times s$ (defined formally in \cref{sec:build_jac}), where $b$ is the batch size (the number of collocation points evaluated). This incurs a negligible cost of $\mathcal{O}(K \cdot s \cdot b )$ per batch, compared to $\mathcal{O}(s^2 \cdot b)$ for a dense embedding. Note that $K=2$ or $K=4$ is typical.

\paragraph{A novel consequence} This particular use of CountSketch provides desirable properties for PINN problems that, to our knowledge, have not previously been explored in the literature. With a fixed sketch size $s$, the expensive mechanisms in the optimiser are fixed, leaving us free to sample as many residuals as we like. Owing to this low cost, we can assemble residuals from very dense sampling into the $s$ buckets. This causes important, or troublesome, regions of the PDE to naturally dominate their buckets, as these have a higher associated value $\bm{\mathcal{F}}_m \bm{u}_{\bm{\theta}} (\bm{x}_i, t_i)$. The sheer number of residuals aggregated into each bucket also averages out noise from the system, yielding a better-conditioned problem to solve. This effectively removes the usual PINN problem of where best to place (a typically small set of) collocation points.

\subsubsection{SRCT}
\label{sec:srct}

The residual dimension (\cref{sec:countsketch}) is compressed once and never reconstructed. The parameter dimension is different: the step is solved in a sketched space and must be lifted back through the embedding to update $\bm\theta$, so its compression must instead be a near-isometry, so that this lift is faithful. To reduce the dimension $d_{\bm\theta}$, we therefore use a random subspace embedding, compressing columns of the Jacobian and thus alleviating the computational cost of Gauss--Newton steps. Historically, Gaussian embeddings were favoured, as they come with strong theoretical guarantees \cite{Martinsson2020RandomizedAlgorithms}. In our work, we instead use subsampled randomised trigonometric transforms (SRTTs), further detailed in \cite[Subsection\,9]{Martinsson2020RandomizedAlgorithms}. These have been observed to match the practical performance of Gaussian embeddings, while additionally admitting a Fast Fourier Transform implementation, as pioneered in SRFT \cite{Ailon2009TheNeighbors}; this gives an improved computational complexity over Gaussian embeddings, $\mathcal{O}(d_{\bm\theta} \log d_{\bm\theta})$ in our case. As a subspace embedding, an SRTT preserves all singular values of a fixed $d$-dimensional subspace to a desired relative accuracy, provided enough columns are retained \cite{Martinsson2020RandomizedAlgorithms}.

SRTTs are composed of a random sign flip $\bm{D} \in \mathbb{R}^{d_{\bm\theta}\times d_{\bm\theta}}$, a random permutation $\bm{\Pi} \in \mathbb{R}^{d_{\bm\theta}\times d_{\bm\theta}}$, a unitary trigonometric transform $\bm{F} \in \mathbb{R}^{d_{\bm\theta}\times d_{\bm\theta}}$ -- applied analogously to the FFT -- and finally, a random restriction $\bm{S} \in \mathbb{R}^{d_{\bm\theta}\times s}$, giving the embedding
\begin{equation}
    \bm{\Omega}\,\bm{S} = (\bm{D}\, \bm{\Pi}\, \bm{F})\, \bm{S}, \qquad \bm{\Omega} := \bm{D}\, \bm{\Pi}\, \bm{F}.
\end{equation}

As a product of orthogonal factors, $\bm\Omega \in \mathbb{R}^{d_{\bm\theta}\times d_{\bm\theta}}$ is itself orthogonal (unitary). We choose the subsampled randomised cosine transform (SRCT) for $\bm{F}$, as we deal with real-valued matrices, opting for the discrete cosine type-II variant. Together with the construction of $\bm{C}$ in \cref{sec:countsketch}, the initialisation of $\bm\Omega$ and $\bm{S}$ constitutes \proctwo{InitSketch} in \cref{alg:dsgnar}, meaning a single, consistent sketch is used throughout every iteration. This is primarily due to compiler constraints (\cref{sec:compiler}), though we also observed empirically that the best results were obtained with a consistent sketch size.

\begin{figure}
    \centering
    \includegraphics[scale=1.0]{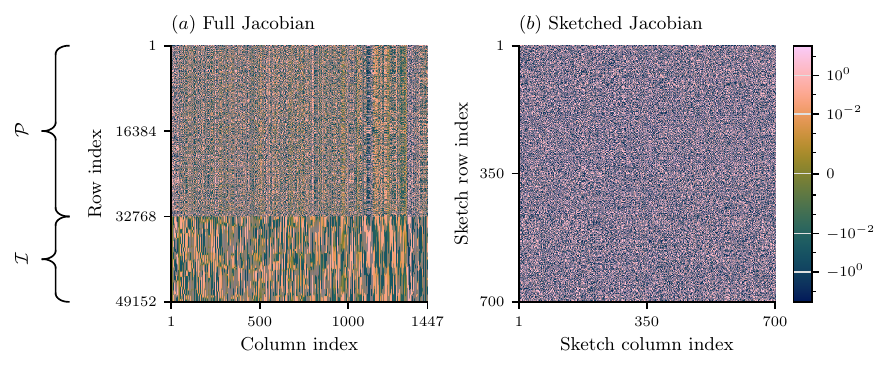}
    \caption{\textbf{Comparison of the full Jacobian (a) with the corresponding doubly-sketched Jacobian (b)}, using CountSketch and SRCT, for an early iteration solving Burgers' equation. The full Jacobian has $N_\mathcal{P} + N_\mathcal{I} = 2^{15} + 2^{14}$
    residuals, and $1447$ trainable parameters. The doubly-sketched Jacobian is a square of size $s=700$, with clear isotropic structure.
    }
    \label{fig:jacobians}
\end{figure}

\subsubsection{Assembling the doubly-sketched Jacobian}
\label{sec:build_jac}

We now walk through \proctwo{Sketch}, shown in \cref{alg:sketch}, which forms the doubly-sketched Jacobian $\tilde{\bm{J}}$, the sketched residual $\tilde{\bm{r}}$, and the full residual $\bm{r}$, without ever forming $\bm{J}$ in full. \cref{fig:jacobians} provides an illustrative example of how the sketched Jacobian compares to its full counterpart. The algorithm begins by looping over the $M$ conditions of the problem -- the \textit{true} residual function $\bm{r}$, which maps over all conditions, is functionally complex: if implemented directly, it would require switch cases depending on the location of the collocation point or the equation being considered, each leading to potentially intensive differential operators. Thus, computing the Jacobian from the true residual function would produce a needlessly large computational graph. However, thanks to CountSketch, and particularly its linearity property \cref{eq:batchable}, we can instead deal with the Jacobians for each condition, $\{ \bm{J}_m\}_{m=1}^{M}$, individually; then, we can simply sum the sketches of these Jacobians together. Each of these Jacobians incurs a scalar multiplication of $\alpha_m = \sqrt{ w_m / |\mathcal{X}_m|}$, applying the normalisation and condition weight of \cref{eq:objective}. As this scaling is applied at accumulation, the routine (featuring all automatic differentiation parts) is agnostic to the current weights $\bm{w}$, and so requires no change between iterations.
Each condition is processed in macro-batches of size $b$, so that even the block $\bm{J}_m$ need never reside in memory in full. Within each macro-batch, residuals and their Jacobians are computed in sub-batches of size $b'$ by reverse-mode automatic differentiation; as $b' \ll d_{\bm\theta}$, this is far cheaper than forward-mode (\cref{tab:sketch_timings}), and, being independent, the sub-batches are dispatched in parallel. Each sub-batch Jacobian is column-sketched by the SRCT, $\hat{\bm{J}}_{\text{batch}} = \bm{J}_{\text{batch}}\,\bm{\Omega}\,\bm{S}$, before the row-compression $\bm{C}$ is applied and accumulated. The full residual $\bm{r}$ is assembled alongside, uncompressed, since the decrease ratio is evaluated in the full space (\cref{sec:subproblem}).

\begin{figure}[t]
\begin{algbox}{Sketch}{Computes sketched Jacobian and residuals}
\label{alg:sketch}
\begin{algIO}
  \algRequire{\bm{\mathcal{F}}}{the $M$ residual conditions}
  \algRequire{\bm{\mathcal{X}}}{the $M$ sets of collocation points}
  \algRequire{\bm{C},\;(\bm{\Omega},\,\bm{S})}{CountSketch and SRCT operators}
  \algRequire{\bm{u}_{\bm{\theta}},\;\bm{\theta}}{the neural network and current parameters}
  \algRequire{\bm{w}}{condition weights}
  \algEnsure{\tilde{\bm{J}} \in \mathbb{R}^{s \times s}}{doubly-sketched Jacobian}
  \algEnsure{\tilde{\bm{r}} \in \mathbb{R}^{s},\;\bm{r} \in \mathbb{R}^{N}}{sketched and full residuals}
\end{algIO}
\begin{algorithmic}[1]
  \Statex \Comment{Hyperparameters: sketch rank $s$, batch size $b$, sub-batch size $b'$}
  \State $\tilde{\bm{J}} \leftarrow \bm{0}^{s \times s}, \quad
         \tilde{\bm{r}} \leftarrow \bm{0}^{s}, \quad
         \bm{r} \leftarrow [\,]$
  \Statex
  \For{$m = 1, \dots, M$}
    \Comment{Loop over $M$ conditions}
    \State $\alpha_m \leftarrow \sqrt{w_m / |\mathcal{X}_m|}$
      \Comment{Condition scaling factor}
    \Statex
    \For{$i = 1, \dots, |\mathcal{X}_m| / b$}
      \Comment{Loop over batches of size $b$; $\bm{J}_m$ never formed in full}
      \State $\bm{J}_{\mathrm{batch}} \leftarrow [\,], \quad
             \bm{r}_{\mathrm{batch}} \leftarrow [\,]$
      \Statex
      \For{$j = 1, \dots, b / b'$}
        \Comment{Loop over sub-batches of size $b'$; parallelise independently}
        \State $\bm{r}_j \leftarrow
               \bm{\mathcal{F}}_m\!\left(\bm{u}_{\bm{\theta}},\,\mathcal{X}_{\mathrm{batch},j}\right)$
          \Comment{Evaluate residuals, $\bm{r}_j \in \mathbb{R}^{b'}$}
        \State $\bm{J}_j \leftarrow \partial_{\bm{\theta}}\,\bm{r}_j$
          \Comment{Reverse-mode Jacobian, $\bm{J}_j \in \mathbb{R}^{b' \times d_{\bm{\theta}}}$;\; $b' \ll d_{\bm{\theta}}$}
        \State $\bm{J}_{\mathrm{batch}} \leftarrow \bm{J}_{\mathrm{batch}} \oplus \bm{J}_j, \quad
               \bm{r}_{\mathrm{batch}} \leftarrow \bm{r}_{\mathrm{batch}} \oplus \bm{r}_j$
          \Comment{Append sub-batch}
      \EndFor
      \Statex
      \State $\hat{\bm{J}}_{\mathrm{batch}} \leftarrow
             \bm{J}_{\mathrm{batch}}\,\bm{\Omega}\,\bm{S}$
        \Comment{Sketch columns, $\hat{\bm{J}}_{\mathrm{batch}} \in \mathbb{R}^{b \times s}$}
      \State $\tilde{\bm{J}} \leftarrow
             \tilde{\bm{J}} + \alpha_m\,\bm{C}\hat{\bm{J}}_{\mathrm{batch}}, \quad
             \tilde{\bm{r}} \leftarrow
             \tilde{\bm{r}} + \alpha_m\,\bm{C}\,\bm{r}_{\mathrm{batch}}$
        \Comment{Accumulate via CountSketch}
      \State $\bm{r} \leftarrow \bm{r} \oplus \alpha_m\,\bm{r}_{\mathrm{batch}}$
        \Comment{Append batch residuals}
    \EndFor
  \EndFor
  \Statex
  \State \Return $\tilde{\bm{J}},\;\tilde{\bm{r}},\;\bm{r}$
\end{algorithmic}
\end{algbox}
\end{figure}

\subsection{Solving the subproblem using \proctwo{SVD}}
\label{sec:subproblem}

A single \proctwo{SVD} of the small square $s \times s$ sketched Jacobian yields inexpensive Levenberg--Marquardt steps, predicted reductions, and the secular map $\lambda\!\leftrightarrow\!\Delta$. This allows one to obtain the steps associated with a set of candidate (or probe) trust-region radii $\{\delta_i\}_i$ at little added cost. An implicit, Jacobian-vector-product solver cannot deliver this factorisation, and would instead need to re-solve from scratch for each predetermined $\lambda$ or $\Delta$.

Sketching the row dimension of the Jacobian yields the sketched residuals $\tilde{\bm{r}}$, while sketching the column dimension means the resulting step $\tilde{\bm{p}}$ lives in a sketched space and must be lifted back to the full parameter space for application. Our smaller, inexact model in the sketched space is written as:

\begin{equation}
    \tilde{m}^\mathsf{LM}_k(\bm{\tilde{p}}; \lambda) := \frac{1}{2} \lVert \tilde{\bm{r}}_k + \tilde{\bm{J}}_k \tilde{\bm{p}} \rVert^2 + \frac{\lambda}{2} \lVert \tilde{\bm{p}} \rVert^2,
    \label{eq:LM_model_approximated}
\end{equation}
with $\tilde{\bm{p}} \in \mathbb{R}^s$, $\tilde{\bm{J}}_k \in \mathbb{R}^{s \times s}$ and $s \ll N$. The analogous trust-region subproblem, with
$\tilde{m}^\mathsf{Q}_k(\bmtilde{p}) := \frac{1}{2} \lVert
\tilde{\bm{r}}_k + \tilde{\bm{J}}_k \tilde{\bm{p}} \rVert^2$,
is then:
\begin{equation}
    \min_{\bmtilde{p}} \ \tilde{m}^\mathsf{Q}_k(\bmtilde{p})
    \qquad \text{subject to} \ \lVert \bmtilde{p} \rVert \le \Delta.
    \label{eq:tr_subproblem_sketched}
\end{equation}
The use of sketching -- namely, the avoidance of ever building the true Gauss--Newton model $m_k^{\textsf{Q}}$ -- means we can never evaluate the ratio in \cref{eq:true_rho} purely from the step $\bm{p}_k$ in full space. We instead compute it approximately, based on the sketched model $\tilde{m}_k^{\textsf{Q}}$, as:
\begin{equation}
  \varrho_k(\bmtilde{p}_k, \bm{p}_k; \bm\theta_k) = \frac{\mathcal{L}(\bm{\theta}_k) - \mathcal{L}(\bm{\theta}_k + \bm{p}_k)}{\tilde{m}^\mathsf{Q}_k(0) - \tilde{m}^\mathsf{Q}_k(\bmtilde{p}_k)}.
    \label{eq:gain_ratio}
    \tag{Ratio $\varrho$}
\end{equation}
Here, the denominator is the predicted decrease in the sketched space, while the numerator is the true reduction in the objective; together these well-approximate the true ratio of \cref{eq:true_rho}.

We solve \cref{eq:LM_model_approximated} using factorisation-based methods, since \cref{sec:finding_step} requires repeatedly re-solving for many candidate $\lambda$ and $\Delta$ from the same sketched Jacobian: a single factorisation reuses itself across this whole array of probes, whereas an iterative solver would need a fresh solve for each. This is made affordable by working in the reduced space of dimension $s$, which also lets us sidestep the cost of Jacobian-vector products and the need for a preconditioner that a Krylov-type method would otherwise require. Given the thin \proctwo{SVD}, $\tilde{\bm{J}}_k=\bm{U}_k\bm{\Sigma}_k\Tp{\bm{V}_k}$, the Levenberg--Marquardt step has the closed-form expression:
\begin{equation}
    \tilde{\bm{p}}_k^{\textsf{LM}}(\lambda) = -\bm{V}_k\,\mathrm{diag}\dfrac{(\sigma_k)_i}{(\sigma_k)_i^2+\lambda}\,\Tp{\bm{U}_k} \bmtilde{r}_k,
    \label{eq:computesolLMproj}
\end{equation}
where $(\sigma_k)_i$ is the $i$-th singular value from $\bm{\Sigma}_k$. This is the method used to obtain steps in \proctwo{DSGNAR}, and is numerically preferable to solving \cref{eq:lm} directly, as it never forms $\Tp{\bm{\tilde{J}}}\bmtilde{J}$, which squares the condition number. This makes iterative methods (Krylov, conjugate gradient) on the already ill-conditioned problem even more prohibitive without a well-chosen preconditioner, discussed in \textcite[Chapter 5]{Conn2000TrustMethods}. Crucially, once the \proctwo{SVD} is computed, evaluating the step for a choice of $\lambda$ is extremely cheap -- costing only $\mathcal{O}(s^2)$ in matrix multiplications -- and, in particular, steps for an array of $\lambda$ choices can be computed quickly in parallel. Further, given a trust-region radius $\Delta$, the corresponding regularisation $\lambda$ such that $\lVert \tilde{\bm{p}}_k^{\textsf{LM}}(\lambda) \rVert = \Delta$ can be obtained efficiently from the \proctwo{SVD} via the secular equation:
\begin{equation}
    \phi(\lambda) := \sqrt{\sum_{i=1}^{s} \frac{(\sigma_k)_i^2\,(\Tp{\bm{U}_k}\tilde{\bm{r}}_k)_i^2}{\big((\sigma_k)_i^2 + \lambda\big)^2}} - \Delta = 0,
    \label{eq:secular}
\end{equation}
iterating Newton's method as in \cite{Conn2000TrustMethods}. It is this ability to re-solve for many $\lambda$ and $\Delta$ from a single \proctwo{SVD} that the algorithm of \cref{sec:algorithm_overview} exploits, probing many candidate radii at once to find a step with a target ratio. Note that even disregarding the conditioning benefits of the \proctwo{SVD}, implicit access to $\Tp{\bm{J}}\bm{J}$ via Jacobian-vector products could not yield such a factorisation.

\subsection{Finding a step with a target ratio $\varrho$}\label{sec:finding_step}
\begin{figure}[t]
\begin{algbox}{LambdaSolve}{Finds $\lambda$ corresponding to target ratio $\varrho^{\star}$}
\label{alg:solve_subproblem}
\begin{algIO}
  \algRequire{\bm{\Omega},\,\bm{S}}{SRCT operators}
  \algRequire{\bm{U},\,\bm{\Sigma},\,\bm{V}^{\top} \in \mathbb{R}^{s \times s}}{\proctwo{SVD} of the sketched Jacobian}
  \algRequire{\tilde{\bm{r}} \in \mathbb{R}^{s},\;\bm{r} \in \mathbb{R}^{N}}{sketched and full residuals}
  \algRequire{\bm{u}_{\bm{\theta}},\;\bm{\theta}}{the neural network and current parameters}
  \algRequire{\Delta_k,\;\varrho^{\star}}{current trust-region radius and target ratio}
  \algEnsure{\Delta^{\star},\;\lambda^{\star}}{selected trust-region radius and regularisation}
\end{algIO}
\begin{algorithmic}[1]
  \Statex \Comment{Hyperparameters: number of probes $q$, Newton iterations $N_{\text{Newton}}$}
  \State $\bm{g} \leftarrow \bm{\Sigma}\,\bm{U}^{\!\top}\,\tilde{\bm{r}}$
    \Comment{$\bm{g} \in \mathbb{R}^s$}
  \State $\delta_i \leftarrow
    \left(\dfrac{\Delta_k}{3}\right)^{\!1 - \frac{i-1}{q-1}}
    \!\left(3\Delta_k\right)^{\frac{i-1}{q-1}}, \quad i = 1, \dots, q$
    \Comment{Geometrically spaced trust-region radius probes}
  \Statex
  \For{$i = 1, \dots, q$}
    \Comment{Loop over probes; parallelise independently}
    \State $\lambda \leftarrow 0$
    \For{$n = 0, 1, \dots, N_{\text{Newton}}$}
      \Comment{Newton steps solving $\|\tilde{\bm{p}}^{\mathrm{LM}}(\lambda)\|_2 = \delta_i$}
      \State $\phi \leftarrow
        \sqrt{\displaystyle\sum_{j=1}^{s}
          \left(\frac{g_j}{\sigma_j^2 + \lambda}\right)^{\!2}} - \delta_i$
      \State $\phi' \leftarrow
        -\dfrac{1}{\phi + \delta_i}
        \displaystyle\sum_{j=1}^{s}
        \frac{g_j^2}{(\sigma_j^2 + \lambda)^3}$
      \State $\lambda \leftarrow \max\!\left(0,\;\lambda - \phi/\phi'\right)$
    \EndFor
    \Statex
    \State $\lambda_i \leftarrow \lambda$
      \Comment{Regularisation parameter for probe $\delta_i$}
    \State $\tilde{\bm{p}}_i \leftarrow
      \displaystyle\sum_{j=1}^{s}
      \frac{g_j}{\sigma_j^2 + \lambda_i}\,\bm{v}_j$
      \Comment{Step in sketch space, $\tilde{\bm{p}}_i \in \mathbb{R}^s$}
    \State $\bm{p}_i \leftarrow \bm{\Omega}\,\bm{S}\,\tilde{\bm{p}}_i$
      \Comment{Lift to full parameter space}
    \State $\varrho_i \leftarrow \varrho(\tilde{\bm{p}}_i,\,\bm{p}_i;\,\bm{\theta})$
      \Comment{Decrease ratio for probe $\delta_i$ as in \eqref{eq:gain_ratio}}
  \EndFor
  \Statex
  \State $\hat{\varrho}_i \leftarrow \min_{1 \le j \le i} \max(0,\,\varrho_j), \quad i = 1, \dots, q$
    \Comment{Enforce $\hat{\varrho}(\delta)$ non-increasing with $\delta$}
  \Statex
  \If{$\hat{\varrho}_q \ge \varrho^{\star}$}
    \State \Return $3\Delta_k,\;\lambda_q$
      \Comment{All probes acceptable; take largest radius}
  \ElsIf{$\hat{\varrho}_1 \le \varrho^{\star}$}
    \State \Return $\tfrac{1}{3}\Delta_k,\;\lambda_1$
      \Comment{No probe acceptable; take smallest radius}
  \Else
    \State $\mathscr{I}(\varrho) \leftarrow
      \proctwo{PCHIP}\!\left(\{\hat{\varrho}_i,\,\log \delta_i\}_i\right)$
      \Comment{Interpolate radius as function of decrease ratio}
    \State $\Delta^{\star} \leftarrow \mathscr{I}(\varrho^{\star})$
      \Comment{Radius achieving target ratio $\varrho^{\star}$}
    \State $\lambda^{\star} \leftarrow \lambda(\Delta^{\star})$
      \Comment{Corresponding regularisation parameter}
    \State \Return $\Delta^{\star},\;\lambda^{\star}$
  \EndIf
\end{algorithmic}
\end{algbox}
\end{figure}

\paragraph{Novelty}
Rather than devising an update strategy for the regularisation $\lambda$ or radius $\Delta$ -- which would be blind to the current landscape -- we instead select each step according to the target decrease ratio $\varrho^\star$. This yields a landscape-aware rule for the regularisation, one that does not depend on the history of previous iterations.
\begin{figure}
    \centering
    \includegraphics[scale=1]{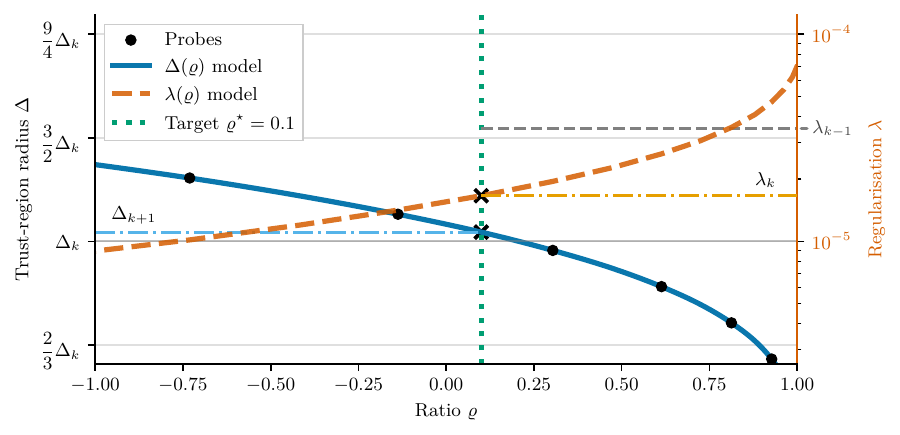}
    \caption{\textbf{The selection of optimisation step}. The model $\Delta(\varrho)$ is built from probe triplets $(\Delta_i, \lambda_i, \varrho_i)$ which are geometrically spaced around the trust-region radius $\Delta_k$. The next iteration's $\Delta_{k+1}$ is selected by interpolating the model at the target ratio $\varrho$. The regularisation corresponding to the target ratio $\varrho$ is chosen as $\lambda_k$ which determines the optimisation step, as in \cref{eq:computesolLMproj} (which is subsequently lifted to the full parameter-space). This visual corresponds to the first optimisation stage, whose goal is to decrease the regularisation over a decrease in the objective, ideally achieving $\lambda_k \ll \lambda_{k-1}$.
    }
    \label{fig:finding_rho}
\end{figure}
As the ratio $\varrho$ is only known once a step has been computed, it cannot be chosen directly beforehand. Traditionally, therefore, it has only been used \emph{post hoc}: to decide whether to increase or decrease the trust-region radius for the next iteration. To instead find a step that achieves some target ratio $\varrho$, probing is required: by interpolating across several candidate steps, one very near to the target can be found.

Since evaluating probes is cheap in \proctwo{DSGNAR} (\cref{sec:subproblem}), we can compute a range of them and read off the radius that attains $\varrho^\star$ by interpolation, illustrated in \cref{fig:finding_rho}. The process is detailed in \proctwo{LambdaSolve} (\cref{alg:solve_subproblem}), where a set of geometrically spaced points around the current trust-region radius is considered. For each trust-region radius probe, we use the already obtained \proctwo{SVD} to cheaply compute the corresponding regularisation $\lambda$ (\cref{sec:subproblem}). Using $\lambda$, the projected step $\tilde{\bm{p}}$ is then computed from \cref{eq:computesolLMproj} and lifted to the full-space step $\bm{p}$. We then use \cref{eq:gain_ratio} to obtain each probe's achieved ratio $\varrho_i$.

The resulting triples $(\Delta_i, \lambda_i, \varrho_i)$ across all probes allow a smooth model to be built (\cref{fig:finding_rho}). We construct this model using a Piecewise Cubic Hermite Interpolating Polynomial (\proctwo{PCHIP}) \cite{Fritsch1980MonotoneInterpolation}. To handle noise, particularly once close to the solution, we force monotonicity in the fit; while not necessarily injective, this still permits straightforward interpolation.

This model allows us to find the trust-region radius and regularisation that approximately achieve the target ratio $\varrho^\star$. If the model does not cover $\varrho^\star$, rather than extrapolate, we select a new set of geometrically spaced probes in the direction of $\varrho^\star$ and repeat.
Due to how this is implemented in efficient compiled code, it is impractical to scan indefinitely until the step corresponding to $\Delta^\star$ is found -- at least, not within the compiled machine-learning frameworks (e.g. JAX \cite{Bradbury2018JAX:Programs}) we use for implementation. Hence, if $\varrho^\star$ is not found within the probe range, we instead test the step associated with the nearest extreme probe. As in \cref{alg:dsgnar}, every such step is still evaluated in full space before acceptance. If the extreme probe with radius $\Delta^\star = \Delta_k / 3$ also fails to yield a decrease, we use $\Delta^\star / 3$ as the centre for the next iteration's probes, covering a region below and overlapping the previous one. We centre probes around $\Delta_k$, rather than around the regularisation parameter $\lambda$, because the trust-region radius is far more stable across iterations (aided by the isotropic parameter embedding of SRCT), whereas $\lambda$ is not. This is illustrated in \cref{solution:wave-double}: for iterations 0--100, the trust-region radius remains consistently close to $1$, while over the same period the regularisation changes by more than 25 orders of magnitude. Because of this stability, the full range of values of $\varrho$ (from less than zero to close to one) is typically obtainable using probes geometrically spaced within $\bigl[\tfrac{1}{3}\Delta_k,\;3\Delta_k\bigr]$. Depending on the architecture, this range can be widened further -- to, say, a factor of $5$ in each direction -- and, depending on GPU budget and the memory cost of a loss evaluation, one may use fewer or more probes.

\paragraph{Accepting the step and updating the radius}\label{sec:accept_reject}
Each candidate step is evaluated in the full parameter space before it is taken. A step is accepted when it yields a smaller objective and a positive ratio. In keeping with the philosophy of \cref{sec:overview}, a step whose achieved ratio \emph{overshoots} the target by more than a small margin is also rejected: an over-aggressive step signals that the trust region is too small, so the radius is expanded, making the subsequent step less damped and giving finer control over the next iteration's probes. On a normal accept, the radius selected by the probe interpolation of \cref{sec:finding_step} is carried forward to centre the next iteration's probes. On a rejection for insufficient decrease, the radius is shrunk. Training stops once the radius falls below the convergence tolerance $\Delta_{\min}$.

\paragraph{Automatic reweighting}\label{sec:adaptive_weights}
A further issue with the objective in \cref{eq:objective} is that the condition weights, $(w_1, \dots , w_M)$, which balance the contributions of each condition (e.g. dynamics $\mathcal{P}$, initial condition $\mathcal{I}$, boundary conditions $\mathcal{B}$), are fixed. In machine learning and PINNs alike, these hyperparameters are notoriously difficult to tune to obtain satisfactory solutions. Each operator $(\bm{\mathcal{P}}, \bm{\mathcal{I}}, \bm{\mathcal{B}})$ has its own scale with complex interactions \cite{Rathore2024ChallengesPerspective}, which has motivated a wide array of proposed dynamic and adaptive weighting schemes \cite{Krishnapriyan2021CharacterizingNetworks,Li2022RevisitingMethod,Cao2025WbPINN:Learning,Wu2026ALearning}. By allowing weights to vary at each iteration $k$, the objective becomes:
\begin{equation}
	\mathcal{L}_k(\bm{\theta}) := \dfrac{1}{2} \lVert \bm{\mathcal{F}}\bm{u}_{\bm\theta} \rVert ^2_{\bm{\mathcal{X}}, \bm{w}_k} = \dfrac{1}{2} \sum_{m=1}^{M} \frac{(\bm{w}_k)_m}{|\mathcal{X}_m|} \sum_{(\bm{x}, t) \in \mathcal{X}_m}  \big[ \bm{\mathcal{F}}_m\bm{u}_{\bm\theta}(\bm{x}, t) \big]^2.\label{eq:variation_objective}
\end{equation}
We provide our own adaptive weighting method, \proctwo{UpdateWeights} (\cref{alg:update_weights}). This is residual-based, and so similar in spirit to \cite{Li2022RevisitingMethod, Anagnostopoulos2024Residual-BasedNetworks}. We do not claim it is intrinsically better than existing work, but the particular use of CountSketch in this paper requires accounting for interactions that do not arise in standard PINN implementations. A second-order method should, in principle, adequately resolve residuals at different scales; however, because CountSketch aggregates residuals from every condition into a shared set of $s$ buckets (\cref{sec:countsketch}), it is essential that all conditions $m=1,\dots,M$ provide values on a comparable scale. The performance of \proctwo{UpdateWeights}, combined with its light computational footprint, makes it well suited to our task.

\paragraph{Automatic stage switching}\label{sec:stage_switch}
The transition from the first (regularisation-decreasing) stage to the second (objective-descent) stage is made automatically by \proctwo{UpdateTargetRatio} (\cref{alg:update_target_ratio}). It monitors the recent history of the regularisation $\lambda$: once a log-linear fit over a sliding window shows a sustained upward trend -- evidence that $\lambda$ has passed its minimum and begun to climb -- the target ratio is raised from its small first-stage value $\varrho^{\textsf{Stage 1}}\le0.2$ to its larger second-stage value $\varrho^{\textsf{Stage 2}}\ge0.5$. At this point, the optimiser shifts from seeking a well-conditioned region to descending towards a nearby minimum. The full procedure, with its window length and trend thresholds, is given in \cref{app:furtheralg}.

\section{Implementation}\label{sec:impdetails}

In this section, we discuss the practical application of \proctwo{DSGNAR} (\cref{alg:dsgnar}) to minimising the objective of \cref{eq:variation_objective}.

\subsection{Environment, machine learning framework and compiler}
\label{sec:compiler}

All experiments in this paper use the machine learning framework \textsc{JAX} \cite{Bradbury2018JAX:Programs} for automatic differentiation. \textsc{JAX} is built on the compiler \textsc{XLA} (Accelerated Linear Algebra) \cite{Sabne2020XLAPerformance}, which performs just-in-time (JIT) compilation. In practice, this means there is an upfront cost to compiling \textsc{JAX} routines, which optimises operations for their specific sizes. Varying the number of residuals per condition across iterations, or the network architecture across time intervals (\cref{sec:time_marching}), would therefore incur further compilation on each change. This would also complicate any implementation of adaptive sketch sizes within this compiled setting, and would require further consideration. We note that the recorded $t_\text{wall}$ times for obtaining solutions omit this compile time, particularly as the compiled routines can be cached to disk.

All experiments are run with \textsc{JAX} version 0.9.1, on \textsc{Ubuntu} version 24.04.4 LTS (Noble Numbat). Initial development was carried out on an \textsc{NVIDIA RTX 2070 SUPER} with 8 GB of GDDR6 memory; later development, and all experiments presented here, used an \textsc{NVIDIA H100 NVL} with 96 GB of HBM3 memory, \textsc{CUDA} version 13.0.

\subsection{Training configuration}
\label{sec:trainconfig}

When applying \proctwo{DSGNAR} (\cref{alg:dsgnar}) to a PDE problem, several initial decisions must be made. For the set of collocation points $\mathcal{X}$, we opt for simple sampling distributions. As detailed in \cref{sec:pinn_setup} and highlighted in \cref{tab:sketch_timings}, unlike most PINN formulations \cite{Wu2023ANetworks}, we are not constrained to a small number of residuals. The residual counts typical of other works remain feasible, and are not essential to the results we observe; however, as laid out in \cref{sec:build_jac}, using more residuals brings clear advantages, such as reducing noise and better capturing important regions. We choose $N^\mathcal{P} = |\mathcal{X}_\mathcal{P}| = 2^{15}$ and $N^\mathcal{I}=N^\mathcal{B}=2^{14}$ in all reported experiments. Smaller numbers may be preferred for GPUs with less memory, or for higher-dimensional problems (where more data is required per point), but any memory issues can be avoided by controlling the batch size. We always sample $\mathcal{X}_\mathcal{P}$ uniformly; depending on the dimension of the problem, $\mathcal{X}_\mathcal{I}$ and $\mathcal{X}_\mathcal{B}$ may use uniform or linear sampling. Each experiment figure explicitly details the sampling method used, chosen primarily for convenience. The only deviation is the lid-driven cavity problem of \cref{app:NSLDC_eq}, which we found to particularly benefit from additional density near the lid: $\mathcal{X}_\mathcal{P}$ is a concatenation of uniform sampling over the full domain and additional uniform sampling over the top ten percent of the domain nearest the lid, with the boundary points sampled likewise.

When comparing our approximate solutions against either exact solutions or results from traditional methods (when no exact solution is available), we evaluate on uniform grids over the relevant space(-time) domain, distinct from the collocation points $\mathcal{X}$ used during training. For high-dimensional problems, such as the 10D Poisson, this is not feasible, and we instead opt for a very dense uniform sample (not to be confused with the training collocation points). Because PINNs do not suffer from overfitting in the conventional sense (\cref{sec:intro}), keeping evaluation points separate from collocation points is not a genuine concern here; nonetheless, we maintain the separation throughout.

The trainable parameters $\bm\theta$ are architecture-dependent, and we opt for standard initialisation methods for each $\bm\theta_0$, detailed in \cref{app:architectures}. For the initial trust-region radius $\Delta_0$, since \proctwo{LambdaSolve} searches for the step matching $\varrho^\star$ regardless of the starting point, this choice has little impact on the final result; we nonetheless choose $\Delta_0 = 1$, as this is often close to $\Delta^\star$ for the first iteration across most architectures. The optimiser halts once the trust-region radius falls below the convergence tolerance $\Delta_{\min}$. For single precision we set $\Delta_\text{min} = 10^{-4}$, and for double precision we set $\Delta_\text{min} = 10^{-8}$. The sketch size is problem- and architecture-dependent, and chosen roughly between $\lfloor d_{\bm\theta}/3 \rfloor$ and $\lfloor d_{\bm\theta}/2 \rfloor$.

\begin{figure}
    \centering
    \includegraphics[scale=1.0]{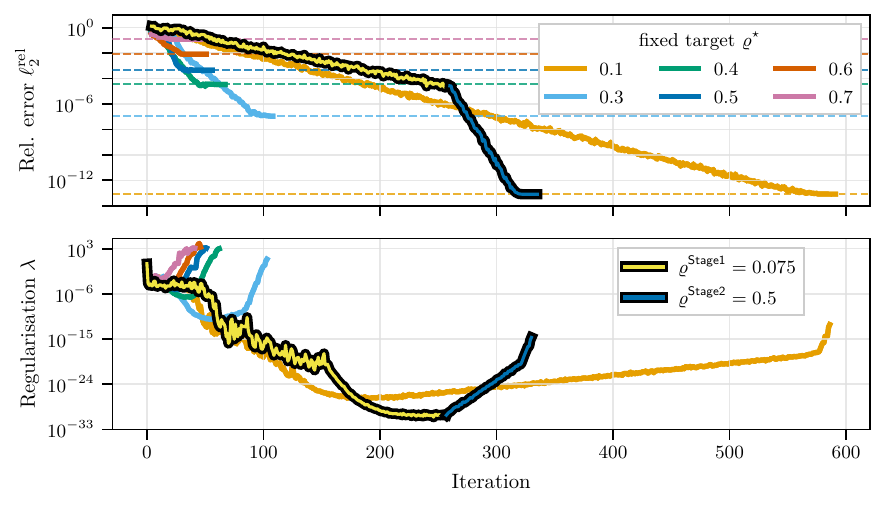}
    \caption{\textbf{Ablation study on the choice of target ratio $\varrho$}, with a comparison against a two-stage approach, when solving Burgers' equation in double precision. We show that smaller choices of the target ratio $\varrho$ provide smaller values of regularisation $\lambda$, and consequently higher-quality solutions. This performance comes at the cost of more iterations; however, the two-stage approach shows that if $\varrho$ switches to a high value once the minimum $\lambda$ is achieved, fewer iterations are required for the same performance.}
    \label{fig:ablation_ratio}
\end{figure}

One important decision, however, is the initial target ratio $\varrho_0$. If adopting the two-stage approach we recommend (\cref{sec:overview}), this will be $\varrho^{\textsf{Stage 1}}$, used for all first-stage iterations. Values around $0.1$ are often ideal: we choose $\varrho^{\textsf{Stage 1}} = 0.1$ for single-precision training and $\varrho^{\textsf{Stage 1}} = 0.075$ for double-precision training, where a lower value corresponds to slower training but potentially a higher-quality solution. For the second stage we conservatively set $\varrho^{\textsf{Stage 2}} = 0.5$; for faster, more aggressive, though potentially more volatile performance, these values can instead be raised to $(\varrho^{\textsf{Stage 1}}, \varrho^{\textsf{Stage 2}}) = (0.2, 0.8)$. We use $24$ probes to build the model $\Delta(\varrho)$ (\cref{sec:finding_step}). As explained in \cref{sec:overview}, the target ratio $\varrho^\star$ strongly shapes the optimisation trajectory: a higher target forces smaller, more conservative steps that nonetheless appear, at least initially, to yield a larger decrease in the objective per iteration; on this highly non-convex landscape, this can substantially change the final minimiser reached -- exactly the effect warned against in \cref{sec:overview}. We quantify this dependence with a dedicated ablation in \cref{fig:ablation_ratio}.

For choices of initial weights $\bm{w}_0$, \proctwo{UpdateWeights} (\cref{alg:update_weights}) alleviates this sensitivity, which is often problematic in PINNs. This is especially highlighted in \cref{solution:kdv-single}, panel (g) (labelled `Weights'), which shows the PDE-condition weight initialised much lower, at $10^{-4}$, than its desired value of $10^{-1}$, yet quickly rising and plateauing at the target. As the optimisation problem associated with the boundary and initial conditions is much simpler, we initialise these weights to $1$, and tend to set the PDE condition, or any other problematic condition, to something small (e.g. $10^{-4}$); these problematic conditions are then gently incorporated over iterations. \proctwo{UpdateWeights} contains one tunable hyperparameter, $\alpha$, which controls how quickly the residuals of different conditions are equalised. We set $\alpha=0.05$ for all experiments, which we found sufficient to equalise residuals adequately given the number of iterations the optimiser typically requires.

\paragraph{Hybrid precision} Single precision requires no special implementation considerations; double precision, however, can (rarely) experience early divergence. We observed that, for poor initialisations of $\bm\theta$, large residuals combined with the optimiser's ability to solve  to the level of noise can lead to early failure. Empirically, this was entirely resolved by carrying out a few initial iterations of the exact same \proctwo{DSGNAR} procedure in single precision. We therefore run the first ten iterations of every double-precision experiment in single precision, to correct any initial problematic elements of $\bm\theta$ and allow drastically mis-scaled weights to align first.

\subsection{Architectures}
\label{sec:impl_architectures}

The literature contains a great deal of recent work developing architectures tailored to PINNs, or even to specific PDE problems. \proctwo{DSGNAR} is architecture-independent, and achieves superior performance over the state of the art even when paired with a simple MLP. However, we observe that some architectures pair better with \proctwo{DSGNAR}: the SIREN architecture (\cref{sec:siren}), in particular, provides considerably more stable trust-region radii and regularisation. We also observed, especially on higher-dimensional problems, that better results are possible with alternative architectures.

For most of the numerical results, a SIREN with four layers and a mildly varying width -- $40$ to $60$ on double-precision problems, to accommodate the added complexity of the PDE -- is employed, to highlight that \proctwo{DSGNAR} is not sensitive, and that the same machinery can be used to tackle most PDE problems with great success. For a select few problems, we show performance with radically different architectures, such as a GaborNet or SPINN, to further display \proctwo{DSGNAR}'s flexibility. Many other architectures, such as LSTMs, were deployed in testing, all yielding strong results. Specific architecture definitions can be found in \cref{app:architectures}. Each network is parameterised by trainable weights $\bm{\theta}$, accepts a single coordinate vector $\bm{x} \in \mathbb{R}^{n_\text{in}}$, and returns an output $\bm{u}_{\bm{\theta}}(\bm{x}) \in \mathbb{R}^{n_\text{out}}$, where $n_\text{in}$ and $n_\text{out}$ are determined by the PDE problem. Each architecture is optionally paired with an input embedding (\cref{sec:fourier}), applied before the main network body.

\subsection{Time marching}\label{sec:time_marching}

This paper considers the standard \ref{eq:simple_objective}, which imparts no inherent sense of time dependence. Instead, it relies on the PDE problem being well-posed, such that violations of the time-dependent dynamics result in a higher loss than a close approximate solution would achieve. This can be viewed as treating the time input variable identically to the spatial inputs: the entire space-time domain is solved in unison, no different to time-independent problems.
However, in some time-dependent settings -- such as large time horizons $T$, or PDEs with stiff dynamics -- solving the whole domain in unison can fail. Time marching addresses this by decomposing the domain into sequential time intervals. As discussed in \cite{Krishnapriyan2021CharacterizingNetworks}, a PINN can instead be trained on many shorter time intervals $[t_n, t_{n+1}]$. At the end of each interval, the network's predicted solution is saved and used as the initial condition for the next interval, $[t_{n+1}, t_{n+2}]$. This process repeats, marching the solution forward in time, and thereby leverages warm starts of the parameters $\bm\theta$. In our results, we use a slightly perturbed version of the previous interval's trainable parameters $\bm\theta$, both to aid faster convergence and to avoid possible divergence from a poor initialisation.

\section{Experiments}
\label{sec:experiments}

We now evaluate the proposed \proctwo{DSGNAR} framework across a diverse set of PDE problems, chosen to test the optimisation philosophy of \cref{sec:method} under markedly different challenges. These include the canonical viscous Burgers' equation (\cref{app:burgers_eq}), a standard benchmark for PINN optimisers featuring a sharp shock; the fourth-order Kuramoto--Sivashinsky equation (\cref{app:KS_eq}), whose chaotic dynamics are challenging even for traditional numerical methods; the ten-dimensional Poisson equation (\cref{app:10dpoisson_eq}), representative of problems beyond the reach of conventional grid-based solvers; and the steady lid-driven cavity Navier--Stokes problem (\cref{app:NSLDC_eq}), a coupled constrained system modelling incompressible fluid flow. Together these problems span many of the principal difficulties encountered in PINNs, namely sharp solution features, long-time dynamics, high dimensionality, and coupled physics systems. Beyond these four representative examples, we report additional benchmarks -- the Wave equation (\cref{app:wave_eq}), the Korteweg--de Vries equation (\cref{app:KdV_eq}), the multi-scale Poisson problem (\cref{app:multiscale_eq}), and a five-dimensional Poisson equation (\cref{app:5dpoisson_eq}) -- whose complete figures and quantitative results are collected in \cref{app:extra_results}.

\begin{figure}[p]
\vspace*{-6em}
\begin{figbox}{Burgers}{Double precision}
\label{solution:burgers-double}
    \includegraphics[scale=0.9, trim={5 0 5 0}, clip]{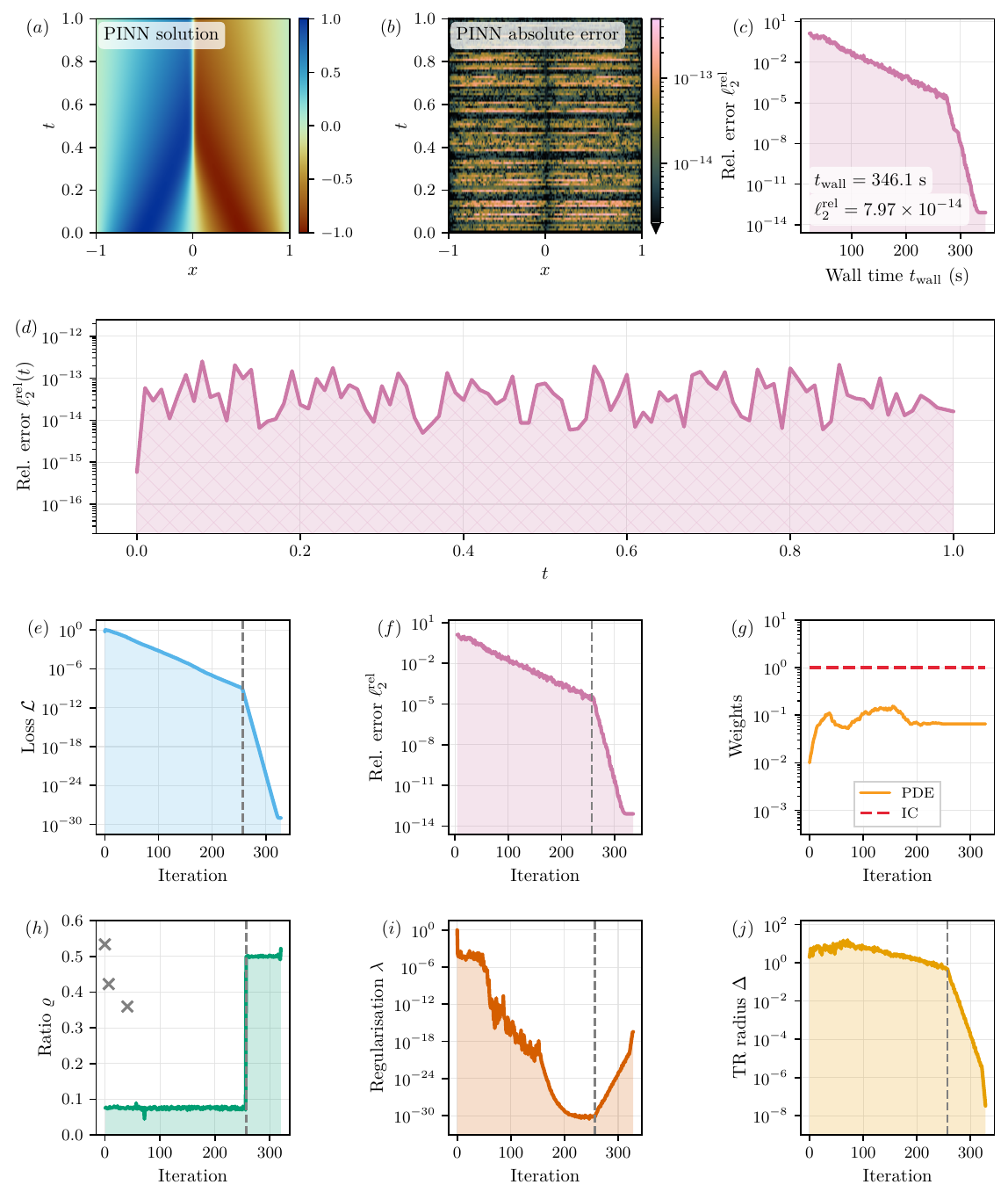}
    \captionof{figboxcap}{
  Results and metrics for the solution of the Burgers equation (see \cref{app:burgers_eq}) in double precision solved with one time-step. For architecture, a SIREN (see \cref{sec:siren}) with hidden dimensions $[60, 60, 60, 60]$ is used ($d_{\bm\theta} = 11{,}285$) with a sketch size of $s = 4000$. For PDE collocation points, we use $N_{\mathcal{\bm{P}}}=2^{15}$ uniformly sampled, and for the initial condition, we use $N_{\mathcal{\bm{I}}}=2^{14}$ linearly spaced. These results prioritise accuracy, whilst computed with remarkable speed, achieving $\ell_2^{\text{rel}} = 7.97 \times 10^{-14}$ in 346.1 seconds.}
\end{figbox}
\end{figure}

\subsection{Evaluation metrics and common optimisation behaviour}
\label{sec:eval_metrics}
A summary of every solve, in both single and double precision where applicable, is provided in \cref{tab:pde_results}, reporting the relative $\ell_2$ error together with the wall-clock training time and comparisons against the current state of the art. To assess solution quality we use the relative error,
\begin{equation}
    \ell_2^{\text{rel}} := \frac{\lVert \bm{u}_{\bm\theta} - \bm{u}^\star \rVert_2}{\lVert \bm{u}^\star \rVert_2},
    \label{eq:rel_l2_error}
\end{equation}
evaluated over the space(-time) domain using points disjoint from the collocation set employed during training. Where practical these evaluation points form a dense uniform grid; otherwise, for higher-dimensional problems, they are taken as a dense uniform sample (\cref{sec:impdetails}). The reference solution $\bm{u}^\star$ is either the exact analytic solution or, where unavailable, a high-fidelity numerical solution generated using a traditional solver. Details for each benchmark are given in \cref{app:PDEs}. For systems with multiple outputs, such as the Navier--Stokes equations, we report $\ell_2^{\text{rel}}$ for each solution component.

Each benchmark is accompanied by a full-page figure, presented either in this section or in \cref{app:extra_results}, with the numerical precision indicated in the upper-right corner. The upper panels visualise the obtained solution, its absolute error relative to the reference solution, and the evolution of $\ell_2^{\text{rel}}$ throughout optimisation. For time-dependent problems we additionally report the error as a function of PDE time, $\ell_2^{\text{rel}}(t)$; for steady problems we instead examine the spatial distribution of the error.

The lower panels report optimisation quantities that are common to every benchmark. Specifically, we show: the objective $\mathcal{L}$ (\cref{eq:variation_objective}); the relative error; the adaptive condition weights updated by \proctwo{UpdateWeights} (\cref{alg:update_weights}); the achieved decrease ratio $\varrho$; the effective regularisation $\lambda$ associated with each accepted step; and the trust-region radius $\Delta$.

These optimisation metrics exhibit remarkably consistent behaviour across all problems. During the first stage of optimisation, the effective regularisation $\lambda$ decreases steadily over many orders of magnitude while the achieved ratio remains close to its conservative target, indicating that the optimiser is primarily seeking a well-conditioned region of parameter space rather than aggressively reducing the objective (\cref{sec:overview}). Once $\lambda$ reaches its minimum, the target ratio is increased automatically, producing the second stage of optimisation in which the loss and solution error decrease rapidly. Throughout this process, the trust-region radius $\Delta$ remains comparatively stable -- typically varying by less than one order of magnitude while $\lambda$ changes by twenty or more. This marked difference motivates centring the probe search around $\Delta$, rather than $\lambda$, as described in \cref{sec:finding_step}. Meanwhile, the adaptive weights converge from their initial values to scales that balance the residuals of the different PDE conditions. We therefore use these common optimisation signatures as a template throughout the remainder of this section, highlighting only those behaviours that are specific to each PDE.

\subsection{Burgers' equation}
\label{sec:res_burgers}

Burgers' equation (\cref{app:burgers_eq}) remains the canonical benchmark for PINN optimisation The double-precision solution is shown in \cref{solution:burgers-double}. Panel (a) recovers the solution throughout the space-time domain, including the narrow internal layer that forms near $x=0$ for viscosity $\nu=0.01$. Panel (b) shows the absolute error against the reference solution, which remains below $10^{-13}$ over almost the entire domain. Importantly, whereas PINN solutions typically exhibit their largest errors along the developing shock, our solution displays some of its \emph{smallest} errors there. This suggests that the dense residual sampling made possible by CountSketch allows the optimiser to resolve precisely the region that is usually the most difficult for PINNs.

The relative error in panel (c) sees a small decrease during the first stage before dropping by approximately nine orders of magnitude during the second stage, ultimately reaching $\ell_2^{\mathrm{rel}} = 7.97\times10^{-14}$ in 346.1 seconds. This behaviour closely follows the optimisation strategy described in \cref{sec:overview}: the first stage primarily seeks a well-conditioned region of parameter space by driving the regularisation down, while the second stage exploits that conditioning to obtain rapid objective reduction.

The optimisation metrics follow the common pattern described previously. The regularisation $\lambda$ (i) decreases to approximately $10^{-30}$ before increasing once the optimiser switches stages, whereas the trust-region radius $\Delta$ (j) remains close to its initial value throughout, only dropping once progress is not possible. Likewise, the loss (e) undergoes its sharpest reduction only after the stage transition, matched closely by the relative error (f), down to the limits of double precision. Approximately 260 iterations are spent reducing the regularisation before only around 70 further iterations are required to converge, illustrating that the majority of the optimisation effort is devoted to locating a highly well-conditioned region rather than directly minimising the objective. The corresponding single-precision solve (\cref{solution:burgers-single}) exhibits the same optimisation behaviour with a smaller network, and is discussed further in \cref{sec:precision_discussion}.

\subsection{Kuramoto--Sivashinsky equation}
\label{sec:res_ks}

\begin{figure}[p]
\vspace*{-6em}
\begin{figbox}{Kuramoto--Sivashinsky}{Double precision}
\label{solution:ks-double}
    \includegraphics[scale=0.9, trim={5 0 5 0}, clip]{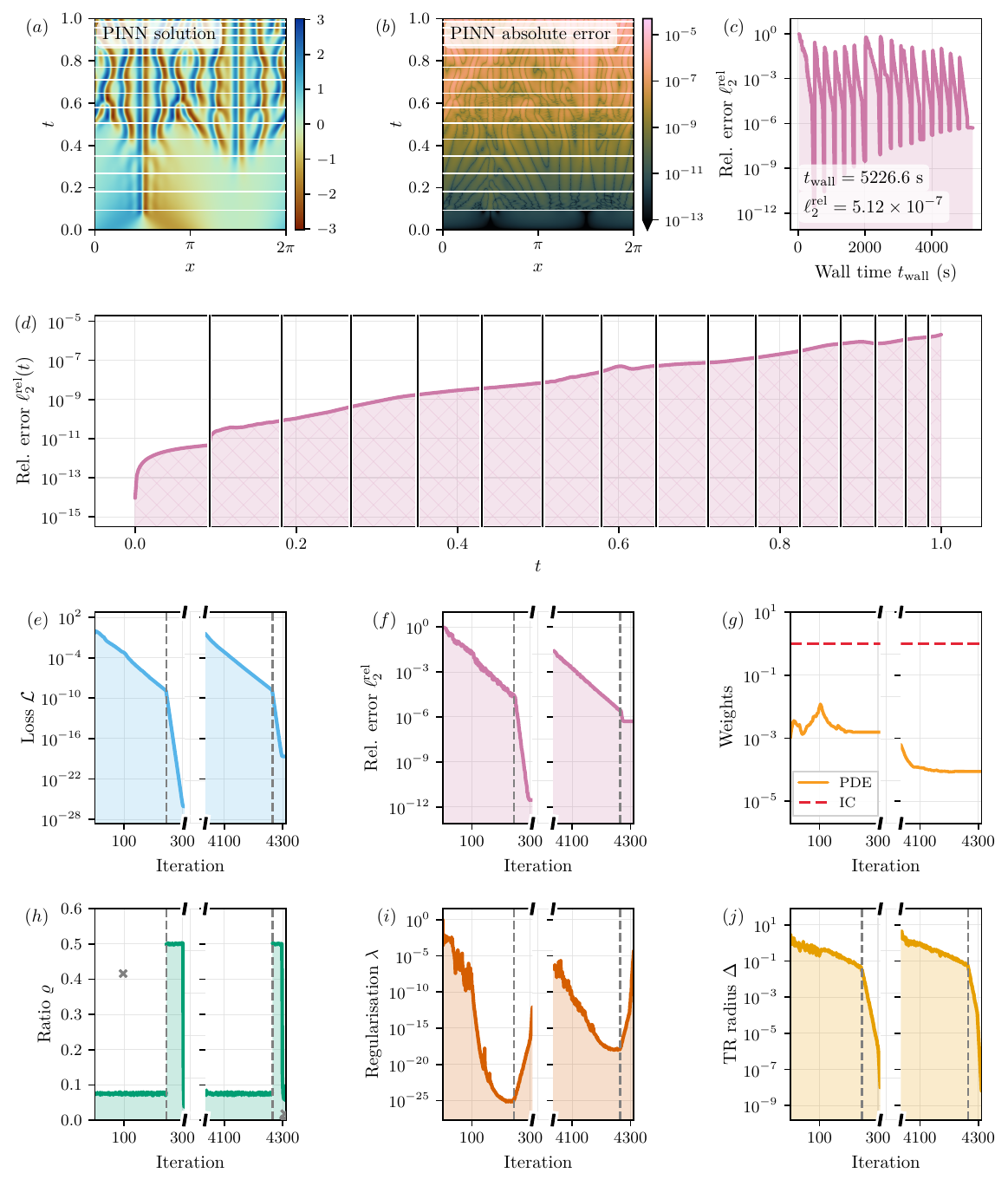}
    \captionof{figboxcap}{
    Results and metrics for the solution of the Kuramoto--Sivashinsky equation (see \cref{app:KS_eq}) in double precision solved with sixteen time-steps. For architecture, a SIREN (see \cref{sec:siren}) with hidden dimensions $[70, 70, 70, 70]$ is used ($d_{\bm\theta} = 15{,}265$) with a sketch size of $s = 5000$. For PDE collocation points, we use $N_{\mathcal{\bm{P}}}=2^{15}$ uniformly sampled, and for the initial condition, we use $N_{\mathcal{\bm{I}}}=2^{14}$ linearly spaced. These results prioritise accuracy, whilst computed with remarkable speed, achieving $\ell_2^{\text{rel}} = 5.12 \times 10^{-7}$ in 5226.6 seconds.
}
\end{figbox}
\end{figure}

\begin{figure}[p]
\vspace*{-6em}
\begin{figbox}{Poisson in 10D}{Double precision}
\label{solution:poisson10-double}
    \includegraphics[scale=0.9, trim={5 0 5 0}, clip]{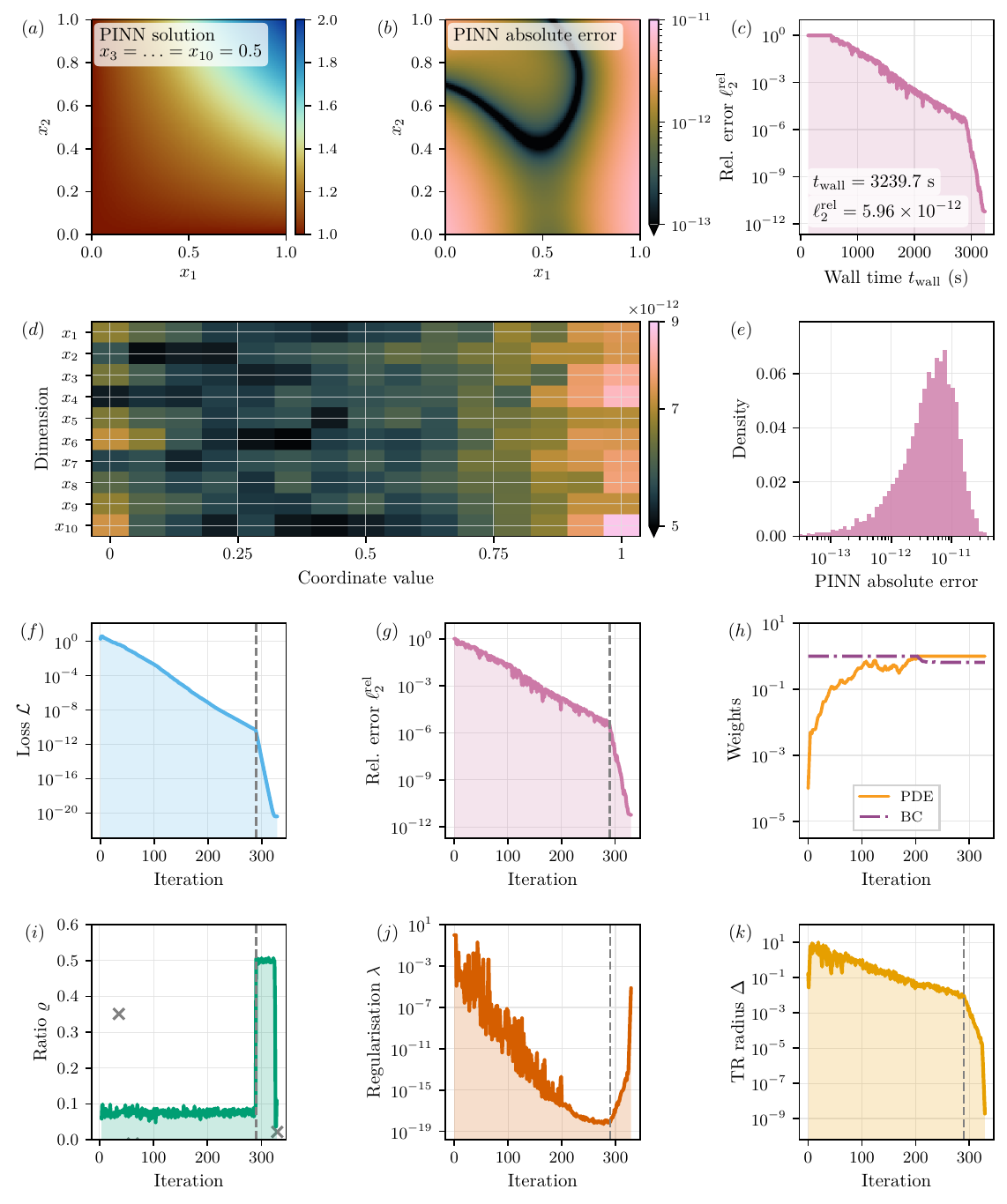}
    \captionof{figboxcap}{
  Results and metrics for the solution of the 10D Poisson equation (see \cref{app:10dpoisson_eq}) in double precision. For architecture, a SPINN (see \cref{sec:spinn}) with hidden dimensions $[20, 20, 20]$ for each branch is used ($d_{\bm\theta} = 17{,}241$) with a sketch size of $s = 5000$. For PDE collocation points, we use $N_{\mathcal{\bm{P}}}=2^{15}$ uniformly sampled, and for the boundary condition, we use $N_{\mathcal{\bm{B}}}=2^{14}$ uniformly sampled. These results prioritise accuracy, whilst computed with remarkable speed, achieving $\ell_2^{\text{rel}} = 5.96 \times 10^{-12}$ in 3239.7 seconds.}
\end{figbox}
\end{figure}

The Kuramoto--Sivashinsky equation (\cref{app:KS_eq}) is a challenging PINN benchmark \cite{Kiyani2025OptimizingNetworks}, where competing second- and fourth-order terms produce chaotic spatio-temporal dynamics. We solve it in double precision over sixteen time intervals using time marching (\cref{sec:time_marching}), requiring a total of 4300 optimisation iterations.

In \cref{solution:ks-double}, panel (a) shows the recovered chaotic solution, with the corresponding absolute error in panel (b). Due to the chaotic nature of the dynamics, small errors are amplified over time, leading to the growth in relative error visible in panel (d). Nevertheless, the solution remains orders of magnitude more accurate than previous PINN results, achieving $\ell_2^{\mathrm{rel}} = 5.12\times10^{-7}$ in 5226.6 seconds.

The bottom panels show the first and final time intervals, each following the same two-stage optimisation behaviour described previously. As the time march progresses, accumulated approximation error produces progressively worse initialisations for later intervals. This is reflected in the optimisation metrics: the minimum regularisation increases over time, indicating that later intervals require stronger damping, as a slightly incorrect initial condition leads to worse data; meanwhile, the loss (e) can still decrease without a corresponding reduction in relative error (f). The condition weights (g) likewise adapt to a different balance between residual terms as the solution evolves through time.

\subsection{High-dimensional Poisson}
\label{sec:res_poisson}

The mesh-free nature of PINNs is particularly valuable in high dimensions, where traditional grid-based methods become impractical due to the curse of dimensionality. We study the ten-dimensional Poisson problem (\cref{app:10dpoisson_eq}), using a separable SPINN architecture (\cref{sec:spinn}) to represent the high-dimensional solution efficiently.
\cref{solution:poisson10-double} shows a two-dimensional slice of the recovered solution in panel (a), with all remaining coordinates fixed. The corresponding absolute error in panel (b) remains uniformly small throughout the domain, with the largest errors occurring near the boundaries as shown in panel (d). The error distribution in panel (e), taken from samples over the entire domain, shows that the absolute error remains below $10^{-10}$ everywhere. The final solution achieves a relative error of $\ell_2^{\mathrm{rel}} = 5.96\times10^{-12}$, improving over previous baselines by more than eight orders of magnitude (\cref{tab:pde_results}).
Despite using a SPINN \cref{sec:spinn}, substantially different architecture to the SIREN-based experiments, the optimisation behaviour remains consistent with the common pattern. The regularisation decreases over the first stage before increasing during the final descent phase, while the trust-region radius remains comparatively stable. This demonstrates that the conditioning mechanism underlying \proctwo{DSGNAR} is not tied to a particular network architecture, but persists even when the parameterisation is substantially changed.

\subsection{Lid-driven cavity (Navier--Stokes)}
\label{sec:res_ns}

Fluid dynamics provides a more demanding test of the framework, as the solution is governed by a coupled system with multiple outputs and constraints. We study the steady lid-driven cavity problem (\cref{app:NSLDC_eq}), a classical benchmark for incompressible Navier--Stokes solvers, in which a moving top wall drives the flow inside a square cavity and produces a primary vortex together with a hierarchy of smaller corner eddies. We solve the equations at Reynolds number $\mathrm{Re}=100$, giving a three-output system consisting of the two velocity components and pressure.

\begin{figure}[p]
\vspace*{-6em}
\begin{figbox}{Lid-driven cavity Navier--Stokes}{Double precision}
\label{solution:ldc-double}
    \includegraphics[scale=0.9, trim={5 0 5 0}, clip]{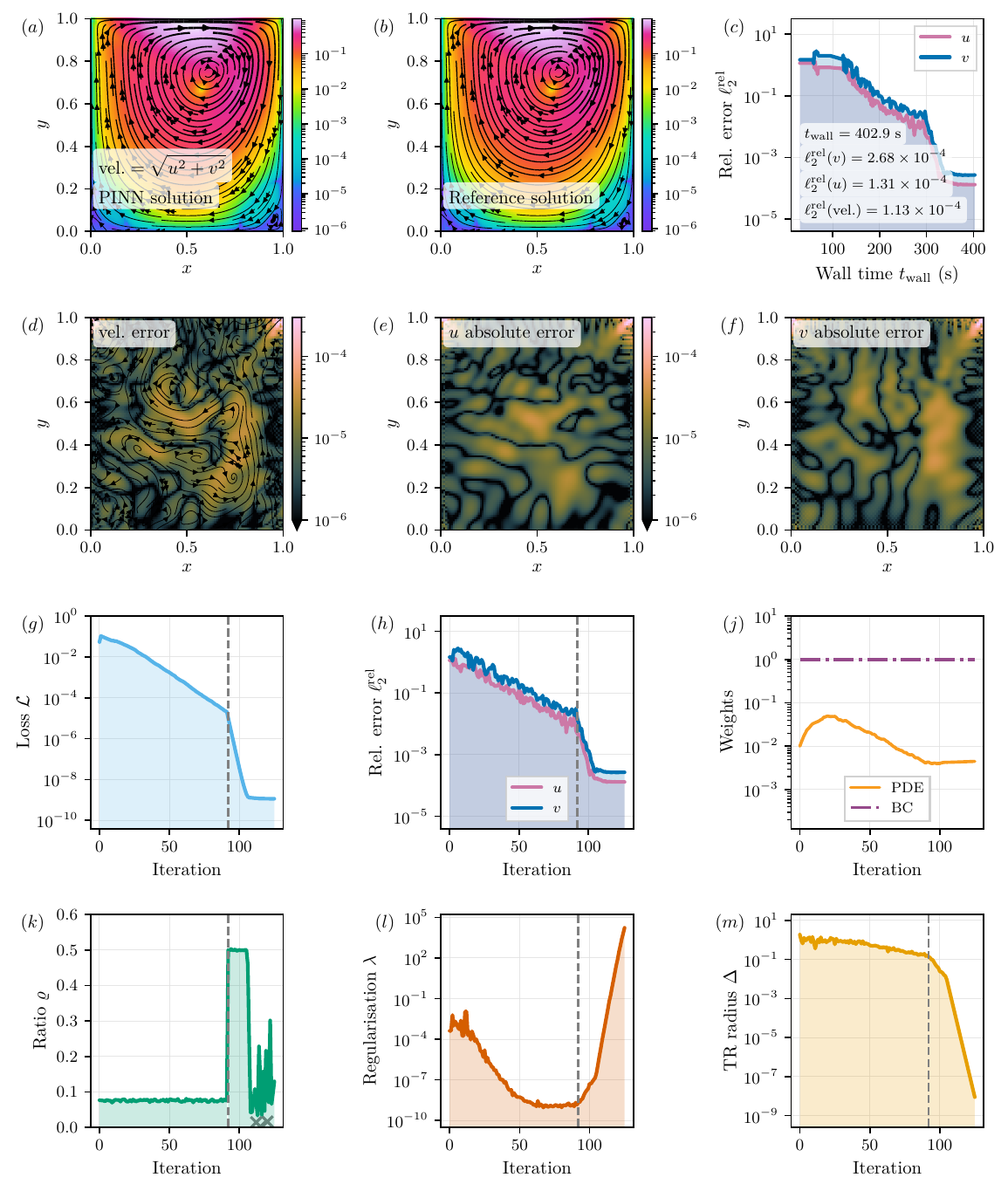}
    \captionof{figboxcap}{
  Results and metrics for the solution of the Lid-driven cavity Navier--Stokes equation (see \cref{app:NSLDC_eq}) in double precision. For architecture, a SIREN (see \cref{sec:siren}) with hidden dimensions $[60, 60, 60, 60]$ for each branch is used ($d_{\bm\theta} = 9{,}219$) with a sketch size of $s = 4000$. For PDE collocation points, we use $N_{\mathcal{\bm{P}}}=2^{15}$ uniformly sampled, and for the boundary condition, we use $N_{\mathcal{\bm{B}}}=2^{14}$ uniformly sampled. These results prioritise accuracy, whilst computed with remarkable speed, achieving $\ell_2^{\text{rel}} = 1.13 \times 10^{-4}$ for the velocity field in 402.9 seconds.}
\end{figbox}
\end{figure}

\cref{solution:ldc-double} shows the recovered velocity field in (a), with the streamlines capturing the primary vortex and secondary corner structures present in the high-fidelity reference solution. The error distributions in (d--f) show that the remaining error is concentrated around the regions of strongest gradients, particularly in the corners. The solution achieves a relative error of $1.13\times10^{-4}$ for the velocity field, obtained in 402.9 seconds.

The optimisation behaviour follows the same general pattern observed throughout the paper. However, unlike other problems where $\lambda$ can often be driven down to the limits of the used precision, the minimum regularisation reached here is considerably higher. This highlights the additional difficulty of the coupled Navier--Stokes system: the standard least-squares PINN objective must simultaneously balance momentum conservation, incompressibility, and boundary conditions, making it harder to identify a region where all residual components can be reduced to the same extent. The trust-region radius remains comparatively stable throughout, reinforcing that the limiting factor is not the step-selection strategy, but the underlying optimisation landscape induced by the PINN formulation itself.

\subsection{Further benchmarks}
\label{sec:res_beyond}

The remaining problems from \cref{tab:pde_results}, with full figures collected in \cref{app:extra_results}, further demonstrate the flexibility of the framework across different PDE problems. We first consider a second Burgers' equation solve (\cref{solution:burgers-extra}) with a different initial condition, where the moving shock is accurately recovered.
The Wave equation is linear and second-order in time, and provides a test of a known weakness of the standard \ref{eq:simple_objective}: that time is not explicitly imparted into the optimisation process. The Wave equation solutions \crefrange{solution:wave-single}{solution:wave-double} in particular show an extremely rapid decrease in the regularisation, while the trust-region radius remains consistent. This highlights the advantage of selecting steps through the target ratio $\varrho^\star$, as this conditioning change would be difficult to follow with traditional strategies that update $\lambda$ based only on parameter history. For this problem, we obtain solutions at the numerical precision limit in both single- and double-precision. The Korteweg--de Vries equation, a dispersive third-order PDE, is solved over four time intervals in \crefrange{solution:kdv-single}{solution:kdv-double}. Each interval follows the same optimisation behaviour observed throughout the paper, with the first stage reducing the regularisation before transitioning into objective descent. As the solution is propagated forward in time, the minimum achievable regularisation increases and the final error grows, reflecting the increasing difficulty of the later time intervals as they are initialised from approximate rather than exact solutions. We investigate the framework on multi-scale problems in \crefrange{solution:multiscale-single}{solution:multiscale-extra}, where Fourier features are leveraged to provide the networks ability to resolve high varied frequencies. In both single- and double-precision we obtain errors close to the numerical limit, and in \cref{solution:multiscale-extra} the intricate multi-scale structure is recovered across the domain. Finally, \cref{solution:poisson5-double} considers the radically different GaborNet architecture (\cref{sec:gabor}). This architecture produces a more rapidly changing optimisation landscape and therefore a less constant trust-region radius than the previous architectures. As the radius moves outside the range covered by the current probes, additional iterations are required to locate steps close to the target ratio -- grey crosses in (i) indicate rejected achieved ratios. Nevertheless, the framework remains effective: only 100 iterations are required to reduce the regularisation to $10^{-35}$ and obtain a solution with $\ell_2^\text{rel}=3.03\times10^{-16}$, in part due to its smoothness -- the remaining absolute error is dominated by floating-point effects, which become clearly visible in (b).

\subsection{Computational cost and scalability}
\label{sec:res_cost}

A key result illustrating the computational trade-off of \proctwo{DSGNAR} is the lid-driven cavity problem in \cref{tab:pde_results}. Here, we solve the same Navier--Stokes system using two sketch sizes, $s=1{,}000$ and $s=4{,}000$, with the network containing $9{,}219$ parameters. The larger sketch captures the local model more accurately and therefore approaches the highest-quality solution possible with our framework. The smaller sketch represents a more aggressive setting, sacrificing some accuracy for a substantial reduction in computational cost, achieving a relative error of $6.34\times10^{-4}$ compared to $1.13\times10^{-4}$, while reducing the solve time from $402.9$ seconds to $80.4$ seconds. This highlights the main computational advantage of the sketched formulation: the sketch size provides a direct control over the balance between accuracy and cost. In problems where the full accuracy of the model is unnecessary, smaller sketches can provide large reductions in runtime while still maintaining accuracy competitive with, or exceeding, existing approaches. As the experiments show errors substantially below many reported PINN results, aggressive sketching provides a practical route towards very fast solves.

\subsection{Single vs. double precision}
\label{sec:precision_discussion}

Double precision is often considered necessary for high-quality PINN solutions, and many recent state-of-the-art results are therefore reported exclusively in double precision. This comes with increased memory requirements and computational cost. Our results demonstrate that this trade-off is not always required. Across four problems in \cref{tab:pde_results}, single-precision training achieves relative errors close to the numerical limit of the chosen precision, while requiring substantially less time. For example, the single-precision Burgers' solve (\cref{solution:burgers-single}) reaches $\ell_2^{\text{rel}} = 4.75\times10^{-7}$ in $9.8$ seconds, using only slightly more than $100$ iterations of \cref{alg:dsgnar}. This improves on the best comparable results in \cref{tab:pde_results} by approximately two orders of magnitude in error while requiring fewer iterations and less computation.

These results suggest that precision should be treated as a practical choice rather than a strict requirement. For applications where $10^{-6}$--$10^{-7}$ relative accuracy is sufficient, single precision provides a substantially cheaper operating point, which has previously been impossible to achieve for PINNs. The effectiveness of \proctwo{DSGNAR} in single precision is highlighted in \cref{tab:pde_results}: on the KdV problem (\cref{app:KdV_eq}), we obtain a relative error of $9.55 \times 10^{-7}$, comfortably outperforming the best reported double-precision result in the literature.

\section{Conclusions and perspectives}
\label{sec:conclusion}

In this paper we introduced \proctwo{DSGNAR}, Doubly-Sketched Gauss--Newton with Adaptive Ratio, an optimisation framework built around a conditioning-first view of PINN training. Rather than chasing the largest immediate decrease in the objective, the method first seeks a region of parameter space where minimal regularisation suffices, and only then descends aggressively once that region is found. This is realised through two coupled ingredients. Firstly, a doubly-sketched Gauss--Newton model compresses both the residual dimension, via CountSketch, and the parameter dimension, via a subsampled randomised cosine transform. This creates a small square matrix whose \proctwo{SVD} can be computed cheaply and reused to evaluate many candidate steps in parallel (\cref{sec:sketched_model}). Secondly, rather than a typical strategy to determine the regularisation $\lambda$ or trust-region radius $\Delta$, \proctwo{DSGNAR} selects each step by probing for a target decrease ratio $\varrho^\star$. This ratio is itself adaptive, raised automatically once minimal regularisation has been achieved, producing the two-stage optimisation behaviour observed throughout \cref{sec:experiments}.

Across a suite of nonlinear, chaotic, multi-scale, high-dimensional, and Navier--Stokes problems (\cref{tab:pde_results}), this combination delivers accuracy far beyond previously reported PINN results while remaining fast. We see a five orders of magnitude improvement over the state of the art on the canonical Burgers' equation, and as much as eight orders of magnitude on a high-dimensional Poisson problem, with relative $\ell_2$ errors as low as $3\times10^{-16}$ in double precision. The same machinery is effective in single precision, reaching $\ell_2^{\text{rel}}=4.75\times10^{-7}$ on Burgers' equation in under ten seconds (\cref{sec:precision_discussion}), suggesting that cheaper single-precision PINNs can now provide highly accurate solves. These results hold with a fixed, small set of hyperparameters across markedly different architectures and problem structures, indicating that \proctwo{DSGNAR} is robust.

Open questions remain. Theoretically, while CountSketch and the SRCT column embedding are each classical subspace embeddings with well-understood guarantees (\cref{sec:countsketch,sec:srct}), these guarantees do not automatically transfer to the doubly-sketched method obtained by composing them; a dedicated subspace-embedding analysis of this combined construction would be desirable. A rigorous analysis of step selection targeting a particular ratio $\varrho^\star$ is left for future work. When training PINNs, linear convergence is rarely, if ever, obtained beyond the initial iterations, with most reported methods instead exhibiting sublinear convergence \cite{Rathore2024ChallengesPerspective,Kiyani2025OptimizingNetworks}; our results (\cref{fig:summary}, \cref{sec:experiments}) instead show  linear convergence throughout the run, until high-accuracy termination is achieved;  theoretical understanding of these improved rates is left for future work.
The coupled Navier--Stokes system of \cref{sec:res_ns} also illustrates that not every problem admits regularisation as small as other problems, consequently leading to a relatively poorer solution -- the bottleneck instead appears to now sit in the underlying PINN objective itself. Since all experiments here use only the basic \ref{eq:simple_objective}, it is therefore possible that alternative PINN formulations, such as FBPINNs \cite{Moseley2023FiniteEquations,Anderson2026ELM-FBPINNs:Method}, could push accuracy further still when combined with \proctwo{DSGNAR}.

Finally, integrating \proctwo{DSGNAR} into an existing scientific machine learning library \cite{Zampini2024PETScML:Learning} would ease adoption by the numerical PDE community and allow more direct comparison against classical solvers on problems of practical interest.

\section*{Acknowledgements} Joseph Webb's work was supported by NAG Ltd. and a EPSRC Case Award, as well as a  Worcester College scholarship  (Paul Thornton)  and an MMSC MSc course bursary (Mathematical Institute, University of Oxford).
Sadok Jerad  and Coralia Cartis acknowledge the support of the Hong Kong Innovation and Technology Commission (InnoHK Project CIMDA). Coralia Cartis’ work was also supported by the EPSRC grant EP/Y028872/1, Mathematical Foundations of Intelligence: An “Erlangen Programme” for AI. The authors gratefully thank Prof Endre S\"ulli for useful discussions and suggestions regarding the numerical experiments.

\printbibliography

\appendix

\section{Further algorithms}\label{app:furtheralg}

\subsection{Residual-based condition re-weighting}
\label{app:reweighting_alg}

\cref{alg:update_weights} presents our condition weight update strategy for $\bm{w}_k$, referenced from \cref{sec:adaptive_weights}. Its purpose is to keep the $M$ conditions of the problem (dynamics $\mathcal{P}$, initial conditions $\mathcal{I}$, boundary conditions $\mathcal{B}$) on a comparable residual scale at every iteration, which matters specifically because of how \proctwo{DSGNAR} builds its sketch: CountSketch (\cref{sec:countsketch}) aggregates residuals from \emph{all} conditions into the same $s$ buckets.

The algorithm first computes the per-condition loss $\ell_m = \|(\bm{r}_k)_m\|^2$ for each of the $M$ conditions, and their mean $\bar\ell_k$ across conditions. Each weight is then rescaled multiplicatively by the ratio of the mean loss to its own condition's loss, raised to a tunable exponent $\alpha$: a condition with an above-average loss ($\ell_m > \bar\ell_k$) has its weight \emph{decreased}, and one with a below-average loss has its weight \emph{increased}, pulling every condition's contribution back towards the mean. The exponent $\alpha$ controls how aggressively this correction is applied: $\alpha = 0$ recovers no re-weighting at all, while larger $\alpha$ more strongly equalises the per-condition losses. We use $\alpha = 0.05$ throughout. A small floor $\varepsilon = 10^{-8}$ guards this ratio against division by zero.

Finally, the whole weight vector is renormalised so that its mean is exactly $1$. Without this step, the multiplicative updates would allow the overall magnitude of $\bm{w}_k$ to drift across iterations.

\begin{figure}[t]
\begin{algbox}{UpdateWeights}{Residual-based condition re-weighting}
\label{alg:update_weights}
\begin{algIO}
  \algRequire{\bm{r}_k}{full residual vector (in structured blocks per condition)}
  \algRequire{\bm{w}_k}{current condition weights}
  \algEnsure{\bm{w}_{k+1}}{updated condition weights}
\end{algIO}
\begin{algorithmic}[1]
  \Statex \Comment{Hyperparameters: re-weighting exponent $\alpha \in \mathbb{R}$}
  \For{$m = 1, \dots, M$}
    \State $\ell_m \leftarrow \|(\bm{r}_k)_m\|^2$
      \Comment{Per-condition loss}
  \EndFor
  \Statex
  \State $\bar{\ell}_k \leftarrow \dfrac{1}{M} \sum_{m=1}^{M} \ell_m$
    \Comment{Mean loss across conditions}
  \Statex
  \For{$m = 1, \dots, M$}
    \State $\hat{w}_m \leftarrow (\bm{w}_k)_m \cdot \left(\dfrac{\bar{\ell}_k}{\ell_m + \varepsilon}\right)^{\!\alpha}$
      \Comment{Scale weight by relative loss}
  \EndFor
  \Statex
  \State $\bm{w}_{k+1} \leftarrow \hat{\bm{w}} \mathbin{/} \left(\dfrac{1}{M}\textstyle\sum_{m=1}^{M} \hat{w}_m\right)$
    \Comment{Normalise: mean weight equals 1}
  \Statex
  \State \Return $\bm{w}_{k+1}$
\end{algorithmic}
\end{algbox}
\end{figure}

\subsection{When to increase the target ratio and enter the second optimisation stage}
\label{app:stage_switch_alg}
\cref{alg:update_target_ratio} presents our method for determining when to trigger the second stage of the optimiser, referenced from \cref{sec:stage_switch}. Recall from \cref{sec:overview} that the first stage deliberately drives the regularisation $\lambda$ down, seeking a well-conditioned region of parameter space, and that the switch to the second (objective-descent) stage should occur once this pursuit has run its course, namely, once $\lambda$ stops decreasing and begins to climb again.

The algorithm works with a sliding window of the $W$ most recent regularisation values, first mapped to a log scale, $\ell_j = \log(\max(\lambda_{k-W+j}, \varepsilon))$, because $\lambda$ evolves over many orders of magnitude (\cref{sec:finding_step}). A line is fit to this window by least squares, giving a slope $\hat s$, and the Pearson correlation coefficient $\hat c$. The two act as complementary checks: $\hat s$ measures whether $\log\lambda$ is climbing \emph{on average}, while $\hat c$ measures how consistently it is doing so. The stage switch is only triggered when both a sustained upward slope ($\hat s > \tau_s$) and a strong, consistent trend ($\hat c > \tau_c$) are observed simultaneously; we use $W = 30$, $\tau_s = 10^{-4}$, and $\tau_c = 0.1$ throughout.

\begin{figure}[t]
\begin{algbox}{UpdateTargetRatio}{Decides if $\varrho$ should be increased for descent}
\label{alg:update_target_ratio}
\begin{algIO}
  \algRequire{\{\lambda_i\}_i}{history of regularisation parameters}
  \algRequire{\varrho_k}{current target decrease ratio}
  \algEnsure{\varrho_{k+1}}{updated target decrease ratio}
\end{algIO}
\begin{algorithmic}[1]
  \Statex \Comment{Hyperparameters: window size $W$, thresholds $\tau_s, \tau_c$}
  \If{$|\{\lambda_i\}| < W$ \kw{or}
  $\varrho_k = \varrho^{\textsf{Stage 2}}$
  }
    \State \Return $\varrho_k$
  \EndIf
  \Statex
  \State $\ell_j \leftarrow \log\bigl(\max(\lambda_{k-W+j},\;\varepsilon)\bigr), \quad j = 1, \dots, W$
    \Comment{Log-scale window of length $W$}
  \State $\hat{s} \leftarrow$ slope of linear fit to $\bm{\ell}$
    \Comment{Positive slope $\Rightarrow$ $\lambda$ is increasing}
  \State $\hat{c} \leftarrow$ Pearson correlation of $\bm{\ell}$
    \Comment{Strength of upward trend}
  \Statex
  \If{$\hat{s} > \tau_s$ \kw{and} $\hat{c} > \tau_c$}
    \State \Return $\varrho^{\textsf{Stage 2}}$
      \Comment{$\lambda$ has passed its minimum; escalate to $\varrho = 0.5$}
  \Else
    \State \Return $\varrho^{\textsf{Stage 1}}$
  \EndIf
  \Statex
\end{algorithmic}
\end{algbox}
\end{figure}

\section{Architectures and features}
\label{app:architectures}

All networks in this work share a common calling convention. For a problem posed on a spatial domain of dimension $n_\text{sp}$, a \emph{time-dependent} network accepts a space--time coordinate $(\bm{x}, t) \in \mathbb{R}^{n_\text{sp}} \times [0, T]$, while a \emph{steady} (time-independent) network accepts a spatial coordinate $\bm{x} \in \mathbb{R}^{n_\text{sp}}$ alone. Writing $n_\text{in}$ for the total input dimension, we have $n_\text{in} = n_\text{sp} + 1$ for time-dependent problems and $n_\text{in} = n_\text{sp}$ for steady ones. Each network is parameterised by trainable weights $\bm{\theta}$ and returns an output $\bm{u}_{\bm{\theta}} \in \mathbb{R}^{n_\text{out}}$, where $n_\text{out}$ is fixed by the problem.

Time is always treated independently of space, such that we can apply spatial input embeddings $\bm{\phi} : \mathbb{R}^{n_\text{sp}} \to \mathbb{R}^{d_\phi}$ , leaving the temporal coordinate
untouched. The vector passed to the network is therefore
\begin{equation}
    \bm{e} =
    \begin{cases}
        \bigl(\bm{\phi}(\bm{x}),\, t\bigr) \in \mathbb{R}^{d_\phi + 1}, & \text{time-dependent},\\[4pt]
        \bm{\phi}(\bm{x}) \in \mathbb{R}^{d_\phi}, & \text{steady},
    \end{cases}
    \label{eq:embedded_input}
\end{equation}
whose dimension we denote $d_e$. Keeping $t$ separate mirrors its role in the PDE formulation of \cref{sec:pinn_setup}, where temporal and spatial operators act differently, and primarily avoids imposing spatial feature maps (such as periodic embeddings) on a direction where they are inappropriate. The choice of embedding $\bm{\phi}$ -- the identity, a periodic mapping, or Fourier features -- is described in \cref{sec:fourier}.

\subsection{Input embeddings}
\label{sec:fourier}

The simplest spatial embedding is the identity, $\bm{\phi}(\bm{x}) = \bm{x}$, in which the raw spatial coordinates are passed straight to the network body. Two richer embeddings are used in this work, both acting per spatial axis after the coordinate is normalised to the unit interval using the per-axis domain bounds,
\begin{equation}
    \tilde{x}_j = \frac{x_j - (x_{\min})_j}{(x_{\max})_j - (x_{\min})_j} \in [0, 1],
    \qquad j = 1, \dots, n_\text{sp}.
    \label{eq:coord_norm}
\end{equation}
In every case the temporal coordinate, when present, is appended to the embedded spatial vector \emph{without} transformation, in keeping with \cref{eq:embedded_input}.

\paragraph{Periodic embedding}
To impose exact periodicity in space, each normalised coordinate is mapped to a cosine--sine pair,
\begin{equation}
    \bm{\phi}(\bm{x}) = \bigl( \cos(2\pi \tilde{x}_j),\ \sin(2\pi \tilde{x}_j) \bigr)_{j=1}^{n_\text{sp}} \in \mathbb{R}^{2 n_\text{sp}},
    \label{eq:periodic_embed}
\end{equation}
so that the network -- and all of its spatial derivatives -- are periodic across the domain by construction. This is the embedding underlying the hard-constraint treatment of \cref{sec:hard_constraints}.

\paragraph{Fourier features}
More generally, $M$ Fourier features per spatial axis are formed from a chosen set of integer modes $\{m_1, \dots, m_M\}$,
\begin{equation}
    \bm{\phi}(\bm{x}) = \Bigl( \cos(2\pi m_k \tilde{x}_j),\ \sin(2\pi m_k \tilde{x}_j) \Bigr)_{\substack{j=1,\dots,n_\text{sp} \\ k=1,\dots,M}} \in \mathbb{R}^{2 M n_\text{sp}},
    \label{eq:fourier_features}
\end{equation}
which recovers the periodic embedding \cref{eq:periodic_embed} when $M = 1$ and $m_1 = 1$. Unlike random Fourier features \cite{Tancik2020FourierDomains}, the modes here are fixed and deterministic; the network learns only how to combine them through its body. The choice of modes can vary. One can choose (linear) modes $m_k = k$, giving $\{1, 2, \dots, M\}$, which populate the low-frequency bands. For problems exhibiting features across disparate length scales, it may be favourable to use geometric (dyadic) modes $m_k = 2^{k-1}$, giving $\{1, 2, 4, \dots, 2^{M-1}\}$, which span a wide range of spatial frequencies with only $M$ features per axis.

\subsection{Hard constraints in PINNs}
\label{sec:hard_constraints}

A limitation of training PINNs with the loss of \cref{eq:simple_objective} is that boundary conditions are only softly enforced, so the learned solution may not adequately satisfy them; choosing (or adaptively setting) the associated weights can be challenging, leading to poor convergence and accuracy \cite{Wang2022WhenPerspective}. Following \textcite{Lagaris1998ArtificialEquations}, one can instead approximate the solution by a network that enforces boundary conditions \emph{analytically} -- a strategy known as \emph{hard constraints}. Periodic boundaries are naturally handled by the periodic spatial embedding $\bm{\phi}$ of \cref{sec:fourier} \cite{Dong2021ANetworks}, as illustrated in
\cref{ex:hard_constraints}.

\begin{figure}[t]
\begin{example}{Analytically enforcing periodic boundaries}
\label{ex:hard_constraints}

Consider Burgers' equation with periodic boundaries on $x \in [-1, 1]$:
\begin{equation*}
    u(-1, t) = u(1, t) \quad \text{and} \quad u_x(-1, t) = u_x(1, t).
\end{equation*}
A standard network $u_{\bm\theta} : \mathbb{R}^2 \to \mathbb{R}$ acting on $(x, t)$ must learn to satisfy these conditions. Instead, applying the periodic spatial embedding $(x, t) \mapsto (\cos(\pi x), \sin(\pi x), t)$ forces the output $u_{\bm\theta}(\cos(\pi x), \sin(\pi x), t)$ to be periodic on $[-1, 1]$ by construction, as are all higher-order partial derivatives with respect to $x$. The boundary condition $\bm{\mathcal{B}}$ can therefore be dropped, leaving only $M = 2$ conditions:
\begin{equation*}
\bm{\mathcal{F}} u = \begin{pmatrix}
u_{t} + u u_x - \nu u_{xx} \\
u(x, 0) - u_0(x)
\end{pmatrix} = \bm{0}.
\end{equation*}
Note that the time coordinate $t$ is left untouched by the embedding, in keeping with \cref{eq:embedded_input}.
\end{example}
\end{figure}

\begin{figure}[t]
\begin{example}{Fourier input features}
\label{ex:fourier}
Consider a steady one-dimensional problem on $x \in [0, 1]$ whose solution contains both slowly- and rapidly-varying components, such as the multi-scale Poisson problem of \cref{app:multiscale_eq}. With $M = 4$ Fourier features and the geometric mode set $\{1, 2, 4, 8\}$, a coordinate $x$ (here already normalised, so $\tilde{x} = x$) is embedded as
\begin{equation*}
    \bm{\phi}(x) = \bigl(x, 
        \cos 2\pi x,\ \sin 2\pi x,\
        \cos 4\pi x,\ \sin 4\pi x,\
        \cos 8\pi x,\ \sin 8\pi x,\
        \cos 16\pi x,\ \sin 16\pi x
    \bigr) \in \mathbb{R}^{9},
\end{equation*}
and the network body sees these eight features alongside the single coordinate $x$. Because the high-frequency modes are supplied directly as inputs, the network need only learn how to weight them, rather than synthesise them through successive nonlinearities -- the difficulty that otherwise stalls
plain $\tanh$ networks on multi-scale solutions.

\end{example}
\end{figure}

\subsection{MLP}
\label{sec:mlp}

The standard multi-layer perceptron (MLP) is a fully connected feed-forward network with $H$ hidden layers of widths $d_1, \dots, d_H$, using the hyperbolic tangent $\tanh$ as its activation function. Writing $\bm{h}^{(0)} = \bm{e}$ for the embedded input of \cref{eq:embedded_input} (with $\bm{\phi}$ the identity when no spatial embedding is used), the forward pass reads:
\begin{equation}
    \bm{h}^{(\ell)} = \tanh\!\left( a_\ell\!\left( \bm{W}^{(\ell)}\bm{h}^{(\ell-1)} + \bm{b}^{(\ell)} \right) \right), \qquad \ell = 1, \dots, H,
\end{equation}
where $\bm{W}^{(\ell)} \in \mathbb{R}^{d_\ell \times d_{\ell-1}}$, $\bm{b}^{(\ell)} \in \mathbb{R}^{d_\ell}$, and $a_\ell \in \mathbb{R}$ is a trainable per-layer activation scale, initialised to~$1$. The output is a final affine map onto the $L$ solution components, $\bm{u}_{\bm\theta} = \bm{W}^{(H+1)}\bm{h}^{(H)} + \bm{b}^{(H+1)} \in \mathbb{R}^{L}$, with $\bm{W}^{(H+1)} \in \mathbb{R}^{L \times d_H}$. Weights are initialised orthogonally with a $\tanh$ gain of $5/3$; output weights use LeCun
initialisation~\cite{LeCun1998EfficientBackProp}.

\paragraph{Trainable parameters}
\begin{equation}
    \bm{\theta} = \{\bm{W}^{(\ell)}, \bm{b}^{(\ell)}\}_{\ell=1}^{H+1} \cup \{a_\ell\}_{\ell=1}^{H},
\end{equation}
flattened into a single vector.

\subsection{SIREN}
\label{sec:siren}

The sinusoidal representation network (SIREN), introduced by \textcite{Sitzmann2020ImplicitFunctions}, replaces the standard nonlinearity with a scaled sine activation. Writing $\bm{h}^{(0)} = \bm{e}$ for the embedded input of \cref{eq:embedded_input}, each of the $H$ hidden layers takes the form:
\begin{equation}
    \bm{h}^{(\ell)} = \sin\!\left( \omega_\ell \left( \bm{W}^{(\ell)}\bm{h}^{(\ell-1)} + \bm{b}^{(\ell)} \right) \right), \qquad \ell = 1, \dots, H,
\end{equation}
where $\omega_\ell > 0$ is a per-layer frequency parameter. A principled initialisation scheme is given in~\cite{Sitzmann2020ImplicitFunctions}: first-layer weights are drawn from $\mathcal{U}\!\left[-1/d_0,\, 1/d_0\right]$, and hidden-layer weights from $\mathcal{U}\!\left[-\sqrt{6/d_{\ell-1}}/\omega_\ell,\, \sqrt{6/d_{\ell-1}}/\omega_\ell\right]$, preserving the distribution of activations across layers. The output is a final affine map onto the $L$ solution components, identical to the MLP above.

In our implementation each $\omega_\ell$ is trained as $\exp(\log \omega_\ell)$, enforcing positivity while allowing the network to tune its own frequency.

\paragraph{Trainable parameters}
\begin{equation}
    \bm{\theta} = \{\bm{W}^{(\ell)}, \bm{b}^{(\ell)}\}_{\ell=1}^{H+1} \cup \{\log\omega_\ell\}_{\ell=1}^{H},
\end{equation}
flattened into a single vector (each $\omega_\ell>0$ is learned via its log). SIRENs are well suited to PDE problems because sinusoidal activations naturally preserve the smoothness and high-frequency structure of solutions and their derivatives, which $\tanh$ networks can struggle to represent~\cite{Sitzmann2020ImplicitFunctions}.

\subsection{GaborNet}
\label{sec:gabor}

GaborNet is a multiplicative filter network~\cite{Fathony2021MultiplicativeNetworks} in which the feature layer is formed from learnable Gabor wavelets rather than sinusoids or standard activations. A Gabor filter is a sinusoid modulated by a Gaussian envelope,
\begin{equation}
    g(\bm{e};\, \bm{\mu},\, \bm{\omega},\, \sigma) = \exp\!\left(-\dfrac{\lVert \bm{e} - \bm{\mu} \rVert^2}{2\sigma^2}\right) \cos(\bm{\omega}^\mathsf{T}\bm{e} + \varphi),
\end{equation}
acting on the embedded input $\bm{e} \in \mathbb{R}^{d_e}$ of \cref{eq:embedded_input}, with trainable filter centres $\bm{\mu} \in \mathbb{R}^{d_e}$, frequency vectors $\bm{\omega} \in \mathbb{R}^{d_e}$, bandwidth $\sigma > 0$, and phase $\varphi \in \mathbb{R}$. Given $F$ filters, each parameterised by $(\bm{\mu}_f, \bm{\omega}_f, \sigma_f, \varphi_f)$, the feature vector is
\begin{equation}
    \bm\psi(\bm{e}) = \bigl(g(\bm{e};\, \bm{\mu}_f, \bm{\omega}_f, \sigma_f, \varphi_f)\bigr)_{f=1}^F \in \mathbb{R}^F,
\end{equation}
and the network output is
$\bm{u}_{\bm\theta} = \bm{W}\bm\psi(\bm{e}) + \bm{b} \in \mathbb{R}^{L}$,
with $\bm{W} \in \mathbb{R}^{L \times F}$.

\paragraph{Trainable parameters}
\begin{equation}
    \bm{\theta} = \{\bm{\mu}_f, \bm{\omega}_f, \sigma_f, \varphi_f\}_{f=1}^{F} \cup \{\bm{W}, \bm{b}\},
\end{equation}
flattened into a single vector.

\subsection{SPINN}
\label{sec:spinn}

The separable physics-informed neural network (SPINN), introduced by \textcite{Cho2023SeparableNetworks}, addresses the exponential scaling of conventional PINNs in high dimensions by factorising the solution approximation into a sum of rank-one products for each input dimension. Each of the $n_\text{in}$ input coordinates is given its own branch network $f_{\bm{\theta}_i} : \mathbb{R}^{d_{\phi,i}} \to \mathbb{R}^R$ of rank $R$, acting on the per-axis embedding $\bm{\phi}_i$ of that coordinate (a Fourier embedding from \cref{sec:fourier} for spatial axes; the identity for the time axis, so that time remains independent of space). The network output is the rank-$R$ combination
\begin{equation}
    \bm{u}_{\bm{\theta}}(\bm{x}, t) = \bm{W}\!\left(
        \bigodot_{i=1}^{n_\text{in}} f_{\bm{\theta}_i}\!\bigl(\bm{\phi}_i\bigr)\right)
    + \bm{b} \in \mathbb{R}^{L},
\end{equation}
where $\bigodot$ denotes the elementwise (Hadamard) product of the branch outputs and $\bm{W} \in \mathbb{R}^{L \times R}$. Each branch is a small MLP with $\tanh$ activations.

This separable structure allows residuals to be evaluated on a tensor grid of $n_\text{in}$ one-dimensional arrays rather than a full $n_\text{in}$-dimensional set of collocation points, reducing the number of network forward passes from $\mathcal{O}(N^{n_\text{in}})$ to $\mathcal{O}(N)$. SPINN is therefore particularly effective for the high-dimensional Poisson problems considered in this work (\cref{tab:pde_results}), where conventional point-wise evaluation
would be computationally prohibitive.

\paragraph{Trainable parameters}
\begin{equation}
    \bm{\theta} = \{\bm{\theta}_i\}_{i=1}^{n_\text{in}} \cup \{\bm{W}, \bm{b}\},
\end{equation}
flattened into a single vector, where each $\bm{\theta}_i$ collects the weights and biases of the $i$-th branch MLP.

\section{Considered PDEs}\label{app:PDEs}

The problems below correspond to those reported in \cref{tab:pde_results},
presented in the same order.

\subsection{\textbf{Burgers' Equation}}\label{app:burgers_eq}

We consider the viscous Burgers' equation:
\begin{equation}\label{eq:burgers}
    \frac{\partial u}{\partial t} + u\,\frac{\partial u}{\partial x}
    - \nu\,\frac{\partial^2 u}{\partial x^2} = 0,
\end{equation}
on a spatially-periodic space-time domain $(x, t) \in [-1, 1] \times [0, 1]$, with
viscosity $\nu = 0.01$.
The initial condition is $u(x, 0) = -\sin(\pi x)$.
Periodic boundary conditions are enforced analytically by mapping the
network input $(x, t) \mapsto (\cos(\pi x), \sin(\pi x), t)$, which
eliminates $\bm{\mathcal{B}}$ from the residual system (see \cref{ex:hard_constraints}).
The resulting operator has $M=2$ conditions:
\begin{equation}
    \bm{\mathcal{F}}\,u
    = \left(\begin{array}{l@{\quad}l}
        u_t + u\,u_x - \nu\,u_{xx} & \forall\,(x,t)\in[-1,1]\times[0,1] \\[4pt]
        u(x, 0) + \sin(\pi x) & \forall\, x\in[-1,1]
      \end{array}\right)
    = \begin{pmatrix} \mathcal{P}\, u \\ \mathcal{I}\, u \end{pmatrix}
    = \bm{0}.
\end{equation}
The reference solution is computed by a Fourier pseudo-spectral discretisation in space ($1024$ modes), advanced in time with an eighth-order Dormand--Prince adaptive Runge--Kutta integrator (\texttt{DOP853}) at absolute and relative tolerances of $10^{-15}$ \cite{Trefethen2000SpectralMATLAB,hairer1993ode1}.

\subsection{\textbf{Korteweg--De Vries Equation}}\label{app:KdV_eq}

We consider the KdV equation:
\begin{equation}\label{eq:KdV_eq}
    u_t + \eta\,u\,u_x + \mu^2\,u_{xxx} = 0,
\end{equation}
on a spatially-periodic space-time domain $(x, t) \in [-1, 1] \times [0, 1]$, with parameters $\eta = 1$ and
$\mu = 0.022$.
The initial condition is $u(x, 0) = \cos(\pi x)$, and periodic boundary
conditions are again enforced analytically (see \cref{ex:hard_constraints}).
The resulting operator has $M=2$ conditions:
\begin{equation}
    \bm{\mathcal{F}}\,u
    = \left(\begin{array}{l@{\quad}l}
        u_t + \eta\,u\,u_x + \mu^2\,u_{xxx} & \forall\,(x,t)\in[-1,1]\times[0,1] \\[4pt]
        u(x, 0) - \cos(\pi x) & \forall\, x\in[-1,1]
      \end{array}\right)
    = \begin{pmatrix} \mathcal{P}\, u \\ \mathcal{I}\, u \end{pmatrix}
    = \bm{0}.
\end{equation}
The reference solution is computed by a Fourier pseudo-spectral discretisation in space ($512$ modes), advanced with the classical fourth-order Runge--Kutta scheme at a step of $5\times10^{-6}$, chosen to respect the dispersive stability restriction $\Delta t \lesssim \Delta x^3/\mu^2$ \cite{Trefethen2000SpectralMATLAB}.

\subsection{\textbf{Wave Equation}}\label{app:wave_eq}

We consider the one-dimensional wave equation:
\begin{equation}
    u_{tt} - c^2\,u_{xx} = 0,
\end{equation}
with $c = 2$ on a space-time domain $(x, t) \in [0,1]^2$, with zero-Dirichlet boundaries and a standing-wave initial condition:
\begin{equation}
    u(x,0) = \sin(\pi x) + \tfrac{1}{2}\sin(4\pi x), \qquad
    u_t(x,0) = 0.
\end{equation}
The resulting operator has $M=4$ conditions:
\begin{equation}
    \bm{\mathcal{F}}\,u
    = \left(\begin{array}{l@{\quad}l}
        u_{tt} - c^2\,u_{xx} & \forall\,(x,t)\in[0,1]^2 \\[4pt]
        u(x, 0) - \sin(\pi x) - \tfrac{1}{2}\sin(4\pi x) & \forall\, x\in[0,1] \\[4pt]
        u_t(x, 0) & \forall\, x\in[0,1] \\[4pt]
        u(\{0,1\}, t) & \forall\, t\in[0,1]
      \end{array}\right)
    = \begin{pmatrix}
        \mathcal{P}\, u \\ \mathcal{I}_1\, u \\ \mathcal{I}_2\, u \\ \mathcal{B}\, u
      \end{pmatrix}
    = \bm{0}.
\end{equation}
The exact solution is
\begin{equation}
    u(x, t) = \sin(\pi x)\cos(2\pi t)
            + \tfrac{1}{2}\sin(4\pi x)\cos(8\pi t).
\end{equation}

\subsection{\textbf{Kuramoto--Sivashinsky Equation}}\label{app:KS_eq}

We consider the one-dimensional Kuramoto--Sivashinsky (KS) equation, a
canonical model for spatio-temporal chaos:
\begin{equation}\label{eq:ks}
    u_t + \alpha\,u\,u_x + \beta\,u_{xx} + \gamma\,u_{xxxx} = 0,
\end{equation}
on a spatially-periodic space-time domain $(x, t) \in [0, 2\pi] \times [0, 1]$, with
\begin{equation}
    \alpha = \frac{100}{16}, \qquad
    \beta  = \frac{100}{16^2}, \qquad
    \gamma = \frac{100}{16^4}.
\end{equation}
The initial condition is $u_0(x) = \cos(x)(1 + \sin(x))$. Periodic boundary conditions are enforced
analytically (see \cref{ex:hard_constraints}). The resulting operator has $M=2$ conditions:
\begin{equation}
    \bm{\mathcal{F}}\,u
    = \left(\begin{array}{l@{\quad}l}
        u_t + \alpha\,u\,u_x + \beta\,u_{xx} + \gamma\,u_{xxxx} & \forall\,(x,t)\in[0,2\pi]\times[0,1] \\[4pt]
        u(x,0) - \cos(x)(1 + \sin(x)) & \forall\, x\in[0,2\pi]
      \end{array}\right)
    = \begin{pmatrix} \mathcal{P}\, u \\ \mathcal{I}\, u \end{pmatrix}
    = \bm{0}.
\end{equation}
The reference solution is computed by a Fourier spectral discretisation in space, advanced with the fourth-order exponential time-differencing Runge--Kutta scheme (ETDRK4) of \textcite{cox2002etd}, whose coefficients are evaluated by contour integration over $64$ Cauchy points \cite{Kassam2006Fourth-OrderPDEs}; the nonlinear term is dealiased by $3/2$ zero-padding.

\subsection{\textbf{Multi-Scale Poisson}}\label{app:multiscale_eq}

We consider a two-dimensional multi-scale Poisson problem (see \textcite{Anderson2026ELM-FBPINNs:Method}):
\begin{equation}
    -\Delta u(x_1, x_2) = f(x_1, x_2), \qquad (x_1, x_2) \in (0,1)^2,
\end{equation}
on the domain $\Omega = [0,1]^2$  with a right-hand side
designed to excite multiple spatial frequencies simultaneously. We have homogeneous Dirichlet boundary conditions $u = 0$ on $\partial[0,1]^2$,
and right-hand side
\begin{equation}
    f(x_1, x_2)
    = \frac{2}{n}\sum_{w \in \mathcal{W}} (w\pi)^2
      \sin(w\pi x_1)\sin(w\pi x_2),
\end{equation}
where $\mathcal{W} = \{4, 8, 12, 16\}$ and $n = |\mathcal{W}| = 4$.
The resulting operator has $M=2$ conditions:
\begin{equation}
    \bm{\mathcal{F}}\,u
    = \left(\begin{array}{l@{\quad}l}
        -\Delta u(x_1,x_2) - f(x_1,x_2) & \forall\,(x_1,x_2)\in(0,1)^2 \\[4pt]
        u(x_1,x_2) & \forall\,(x_1,x_2)\in\partial[0,1]^2
      \end{array}\right)
    = \begin{pmatrix} \mathcal{P}\, u \\ \mathcal{B}\, u \end{pmatrix}
    = \bm{0}.
\end{equation}
Since the problem is steady, there is no initial-condition operator $\bm{\mathcal{I}}$.
The exact solution is
\begin{equation}
    u_\star(x_1, x_2)
    = \frac{1}{n}\sum_{w \in \mathcal{W}}
      \sin(w\pi x_1)\sin(w\pi x_2).
\end{equation}

\subsection{\textbf{High-Dimensional Poisson}}\label{app:highdpoisson_eq}

We consider two high-dimensional Poisson problems following
\textcite{Guzman-Cordero2025ImprovingRandomization}.

\paragraph{\textbf{5D Poisson}.\label{app:5dpoisson_eq}}
The problem is $-\Delta u(\bm{x}) = f(\bm{x})$ on $[0,1]^5$, with
right-hand side
\begin{equation}
    f(\bm{x}) = \pi^2 \sum_{i=1}^{5} \cos(\pi x_i),
\end{equation}
and Dirichlet boundary conditions $u(\bm{x}) = \sum_{i=1}^5 \cos(\pi x_i)$
for $\bm{x} \in \partial[0,1]^5$.
The resulting operator has $M=2$ conditions:
\begin{equation}
    \bm{\mathcal{F}}\,u
    = \left(\begin{array}{l@{\quad}l}
        -\Delta u(\bm{x}) - \pi^2\displaystyle\sum_{i=1}^{5}\cos(\pi x_i) & \forall\,\bm{x} \in (0,1)^5 \\[6pt]
        u(\bm{x}) - \displaystyle\sum_{i=1}^{5}\cos(\pi x_i) & \forall\,\bm{x} \in \partial[0,1]^5
      \end{array}\right)
    = \begin{pmatrix} \mathcal{P}\, u \\ \mathcal{B}\, u \end{pmatrix}
    = \bm{0}.
\end{equation}
The exact solution is $u_\star(\bm{x}) = \sum_{i=1}^5 \cos(\pi x_i)$.

\paragraph{\textbf{10D Poisson}.\label{app:10dpoisson_eq}}
The problem is $-\Delta u(\bm{x}) = 0$ on $[0,1]^{10}$ with harmonic
boundary conditions
\begin{equation}
    u(\bm{x}) = \sum_{i=1}^{5} x_{2i-1}\,x_{2i}, \qquad
    \bm{x} \in \partial[0,1]^{10}.
\end{equation}
The resulting operator has $M=2$ conditions:
\begin{equation}
    \bm{\mathcal{F}}\,u
    = \left(\begin{array}{l@{\quad}l}
        -\Delta u(\bm{x}) & \forall\,\bm{x} \in (0,1)^{10} \\[4pt]
        u(\bm{x}) - \displaystyle\sum_{i=1}^{5} x_{2i-1}\,x_{2i} & \forall\,\bm{x} \in \partial[0,1]^{10}
      \end{array}\right)
    = \begin{pmatrix} \mathcal{P}\, u \\ \mathcal{B}\, u \end{pmatrix}
    = \bm{0}.
\end{equation}
The exact solution is $u_\star(\bm{x}) = \sum_{i=1}^5 x_{2i-1}\,x_{2i}$.

\subsection{\textbf{Lid-Driven Cavity Flow}}\label{app:NSLDC_eq}

We consider the steady incompressible Navier--Stokes equations in the
unit square $\Omega = [0,1]^2$ at Reynolds number $\mathrm{Re} = 100$:
\begin{align}
    u\,u_x + v\,u_y + p_x - \tfrac{1}{\mathrm{Re}}(u_{xx} + u_{yy}) &= 0, \\
    u\,v_x + v\,v_y + p_y - \tfrac{1}{\mathrm{Re}}(v_{xx} + v_{yy}) &= 0, \\
    u_x + v_y &= 0.
\end{align}
The network outputs three fields: $x$-velocity $u$, $y$-velocity $v$, and
pressure $p$. Boundary conditions are imposed on all four walls:
\begin{equation}
    u(x,y) = \begin{cases} 4x(1-x) & y = 1\ \text{(moving lid)},\\ 0 & \text{otherwise,} \end{cases}
    \qquad v(x,y) = 0 \quad \forall\,(x,y)\in\partial\Omega.
\end{equation}
Since only gradients of $p$ appear in the momentum equations, the pressure
field is determined only up to an additive constant; one has the option to pin it via $p(0,0) = 0$, but we instead allow an arbitrary additive constant so that we have one fewer condition to solve for. The resulting operator has
$M=5$ conditions
\begin{equation}
    \bm{\mathcal{F}}\,\bm{u}
    = \left(\begin{array}{l@{\quad}l}
        u\,u_x + v\,u_y + p_x - \frac{1}{\mathrm{Re}}(u_{xx}+u_{yy})
            & \forall\,(x,y)\in\Omega \\[3pt]
        u\,v_x + v\,v_y + p_y - \frac{1}{\mathrm{Re}}(v_{xx}+v_{yy})
            & \forall\,(x,y)\in\Omega \\[3pt]
        u_x + v_y
            & \forall\,(x,y)\in\Omega \\[3pt]
        u(x,y) - \mathbb{I}_{\{y=1\}}\cdot 4x(1-x)
            & \forall\,(x,y)\in\partial\Omega \\[3pt]
        v(x,y)
            & \forall\,(x,y)\in\partial\Omega
    \end{array}\right)
    = \begin{pmatrix}
        \mathcal{P}_1\, \bm{u} \\ \mathcal{P}_2\, \bm{u} \\ \mathcal{P}_3\, \bm{u} \\ \mathcal{B}_1\, \bm{u} \\ \mathcal{B}_2\, \bm{u}
    \end{pmatrix}
    = \bm{0},
\end{equation}
where $\mathcal{P}_1, \mathcal{P}_2, \mathcal{P}_3$ denote the $x$-momentum, $y$-momentum, and continuity equations respectively, and $\mathcal{B}_1, \mathcal{B}_2$ the boundary conditions on $u$ and $v$; as the problem is steady, there is no initial-condition operator $\bm{\mathcal{I}}$.
The reference solution is computed with the \textsc{Dedalus} spectral framework \cite{Burns2020Dedalus}, using a Chebyshev discretisation in both directions ($100\times100$) and a fourth-order semi-implicit backward-difference time-stepper (\texttt{SBDF4}), marched to steady state at $\mathrm{Re}=100$ with a regularised lid velocity $u(x,1)=4x(1-x)$.

\section{Results and metrics for comprehensive suite of PDEs}
\label{app:extra_results}

Below are a set of figures showing results and metrics for the rest of the PDEs in \cref{tab:pde_results}, with some extra problems to showcase \proctwo{DSGNAR}'s capabilities.

\begin{figure}[p]
\vspace*{-6em}
\begin{figbox}{Burgers}{Single precision}
\label{solution:burgers-single}
    \includegraphics[scale=0.9, trim={5 0 5 0}, clip]{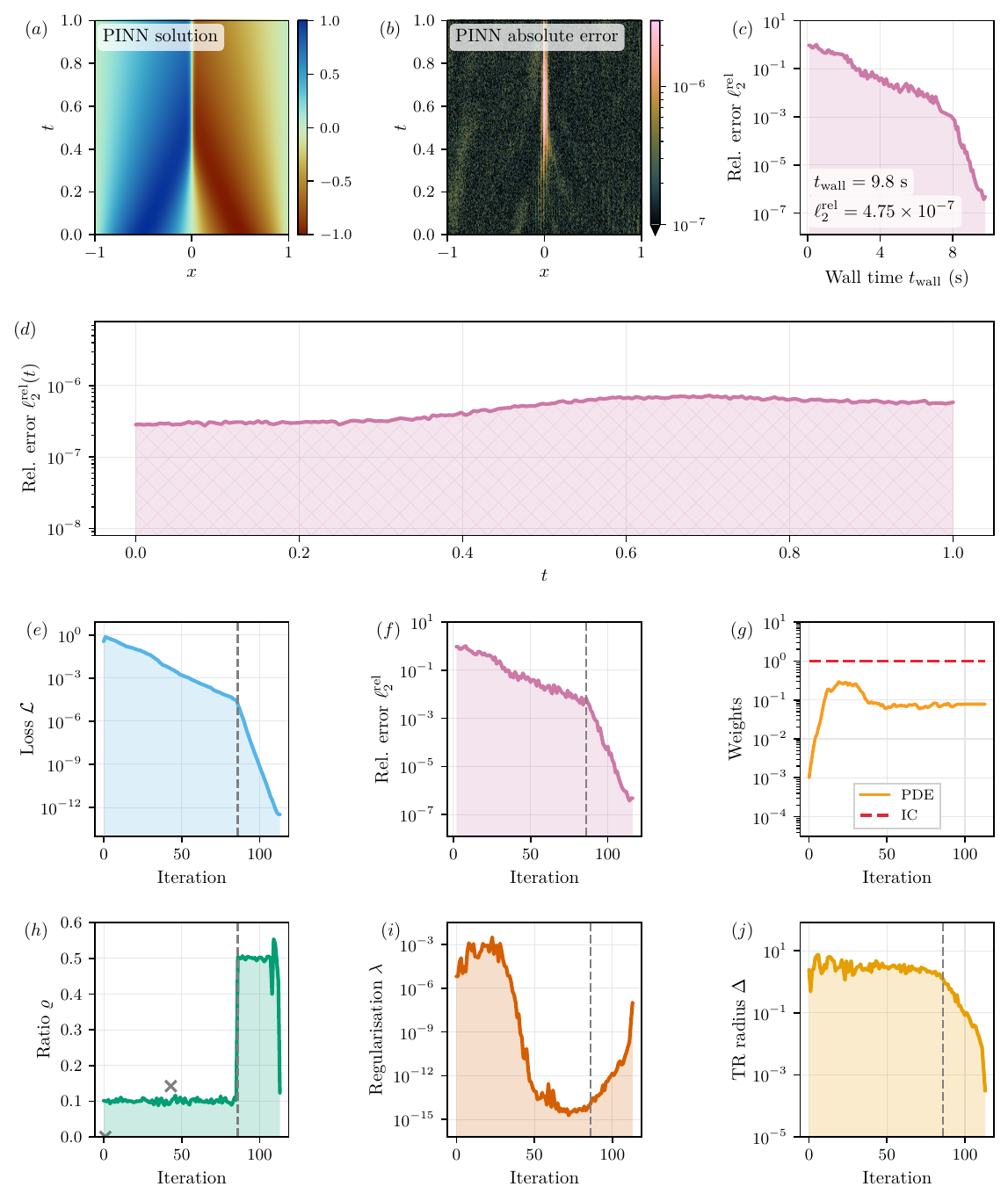}
    \captionof{figboxcap}{
  Results and metrics for the solution of the Burgers equation (see \cref{app:burgers_eq}) in single precision solved with one time-step. For architecture, an MLP with hidden dimensions $[16, 16, 16, 16, 16, 16]$ is used ($d_{\bm\theta} = 1{,}447$) with a sketch size of $s = 700$. For PDE collocation points, we use $N_{\mathcal{\bm{P}}}=2^{15}$ uniformly sampled, and for the initial condition, we use $N_{\mathcal{\bm{I}}}=2^{14}$ linearly spaced. These results prioritise compute time, whilst obtaining near machine error, achieving $\ell_2^{\text{rel}} = 4.75 \times 10^{-7}$ in 9.8 seconds.}
\end{figbox}
\end{figure}

\begin{figure}[p]
\vspace*{-6em}
\begin{figbox}{Burgers with different initial condition}{Double precision}
\label{solution:burgers-extra}
    \includegraphics[scale=0.9, trim={5 0 5 0}, clip]{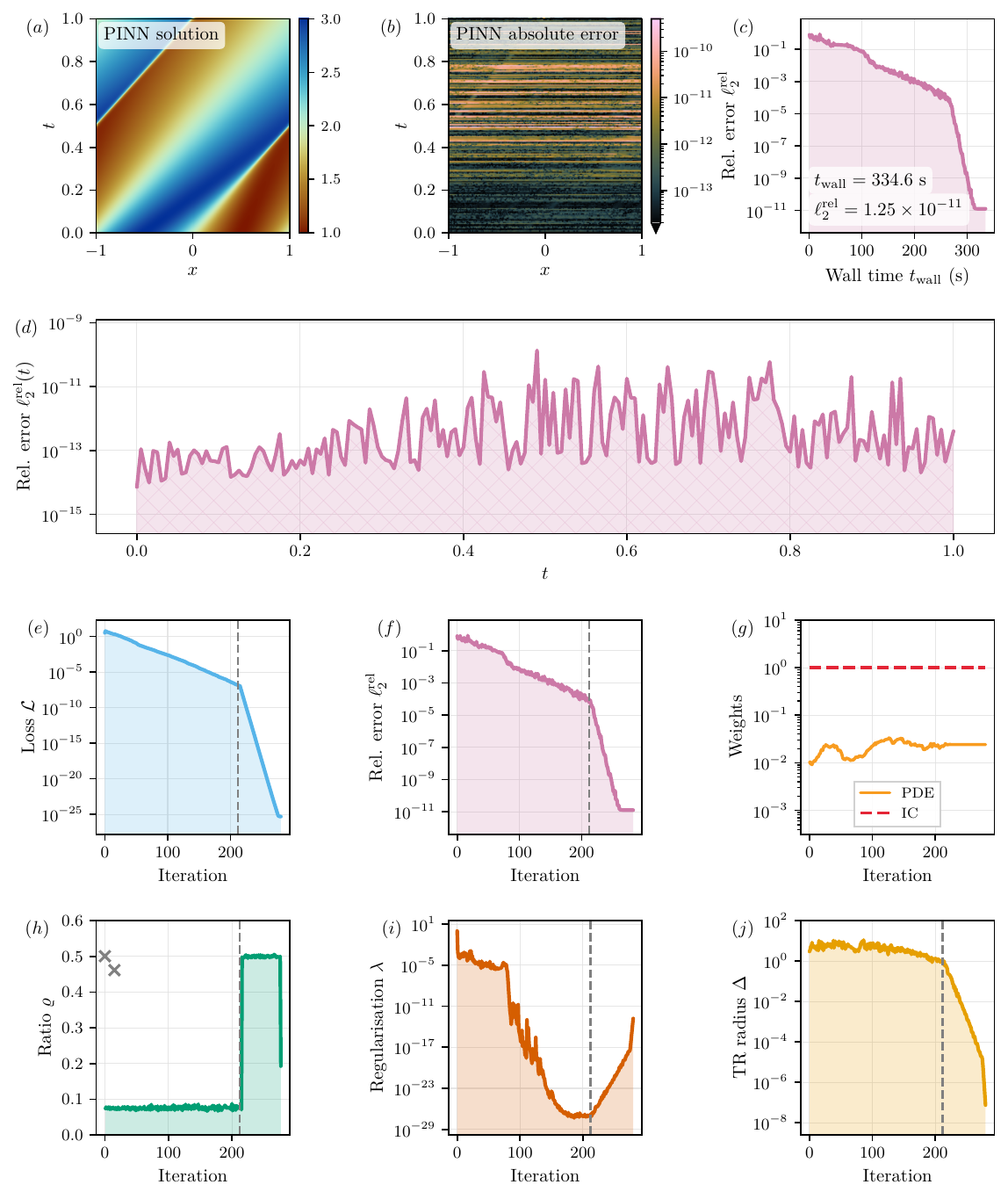}
    \captionof{figboxcap}{
  Results and metrics for the solution of the Burgers equation (see \cref{app:burgers_eq}) in double precision solved with one time-step. This problem differs from the other double-precision Burgers' equation in \cref{solution:burgers-double} by its initial condition, $u(x, 0) = -\sin(\pi x) + 2$. All other aspects of architecture and implementation are kept the same. These results prioritise accuracy, whilst computed with remarkable speed, achieving $\ell_2^{\text{rel}} = 1.25 \times 10^{-11}$ in 334.6 seconds.}
\end{figbox}
\end{figure}

\begin{figure}[p]
\vspace*{-6em}
\begin{figbox}{Wave}{Single precision}
\label{solution:wave-single}
    \includegraphics[scale=0.9, trim={5 0 5 0}, clip]{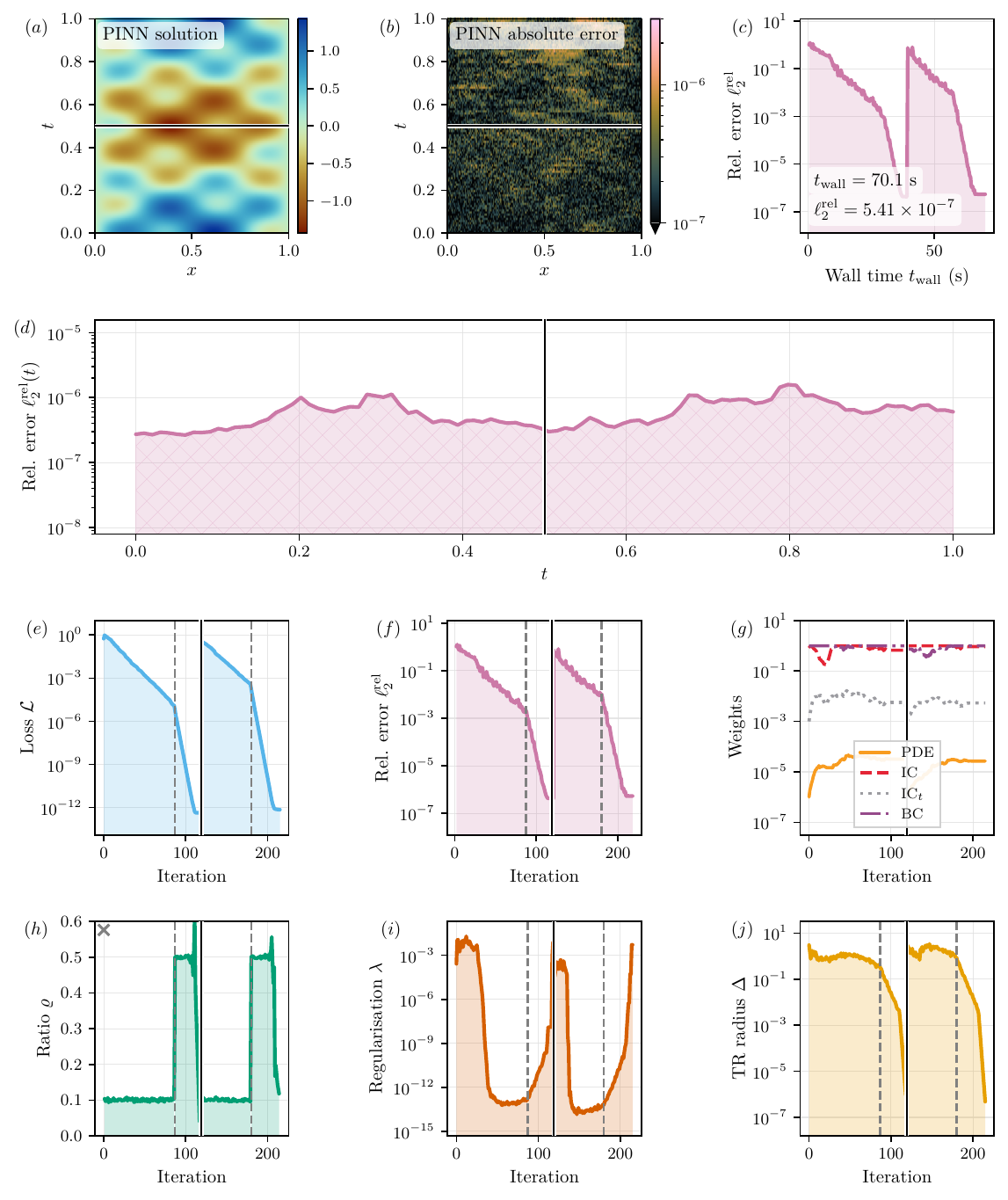}
    \captionof{figboxcap}{
  Results and metrics for the solution of the Wave equation (see \cref{app:wave_eq}) in single precision solved with two time-steps. For architecture, a SIREN (see \cref{sec:siren}) with hidden dimensions $[40, 40, 40, 40]$ is used ($d_{\bm\theta} = 5{,}085$) with a sketch size of $s = 2000$. For PDE collocation points, we use $N_{\mathcal{\bm{P}}}=2^{15}$ uniformly sampled, and for the initial condition, we use $N_{\mathcal{\bm{I}}}=2^{14}$ linearly spaced. These results prioritise accuracy, whilst computed with remarkable speed, achieving $\ell_2^{\text{rel}} = 5.41 \times 10^{-7}$ in 70.1 seconds.}
\end{figbox}
\end{figure}

\begin{figure}[p]
\vspace*{-6em}
\begin{figbox}{Wave}{Double precision}
\label{solution:wave-double}
    \includegraphics[scale=0.9, trim={5 0 5 0}, clip]{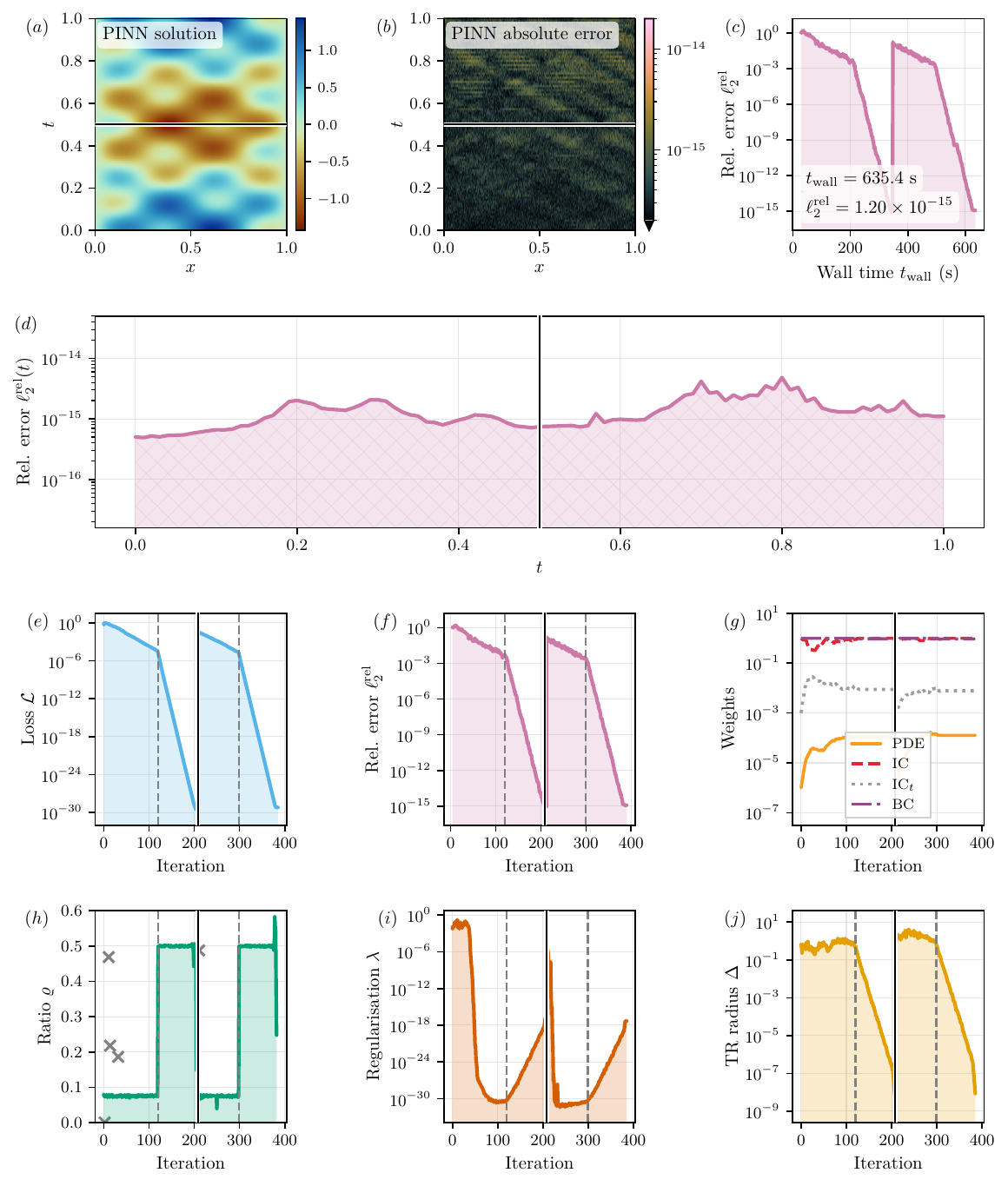}
    \captionof{figboxcap}{
  Results and metrics for the solution of the Wave equation (see \cref{app:wave_eq}) in double precision solved with two time-steps. For architecture, a SIREN (see \cref{sec:siren}) with hidden dimensions $[60, 60, 60, 60]$ is used ($d_{\bm\theta} = 11{,}225$) with a sketch size of $s = 4000$. For PDE collocation points, we use $N_{\mathcal{\bm{P}}}=2^{15}$ uniformly sampled, and for the initial condition, we use $N_{\mathcal{\bm{I}}}=2^{14}$ linearly spaced. These results prioritise accuracy, whilst computed with remarkable speed, achieving $\ell_2^{\text{rel}} = 1.20 \times 10^{-15}$ in 635.4 seconds.}
\end{figbox}
\end{figure}

\begin{figure}[p]
\vspace*{-6em}
\begin{figbox}{Korteweg--De Vries}{Single precision}
\label{solution:kdv-single}
    \includegraphics[scale=0.9, trim={5 0 5 0}, clip]{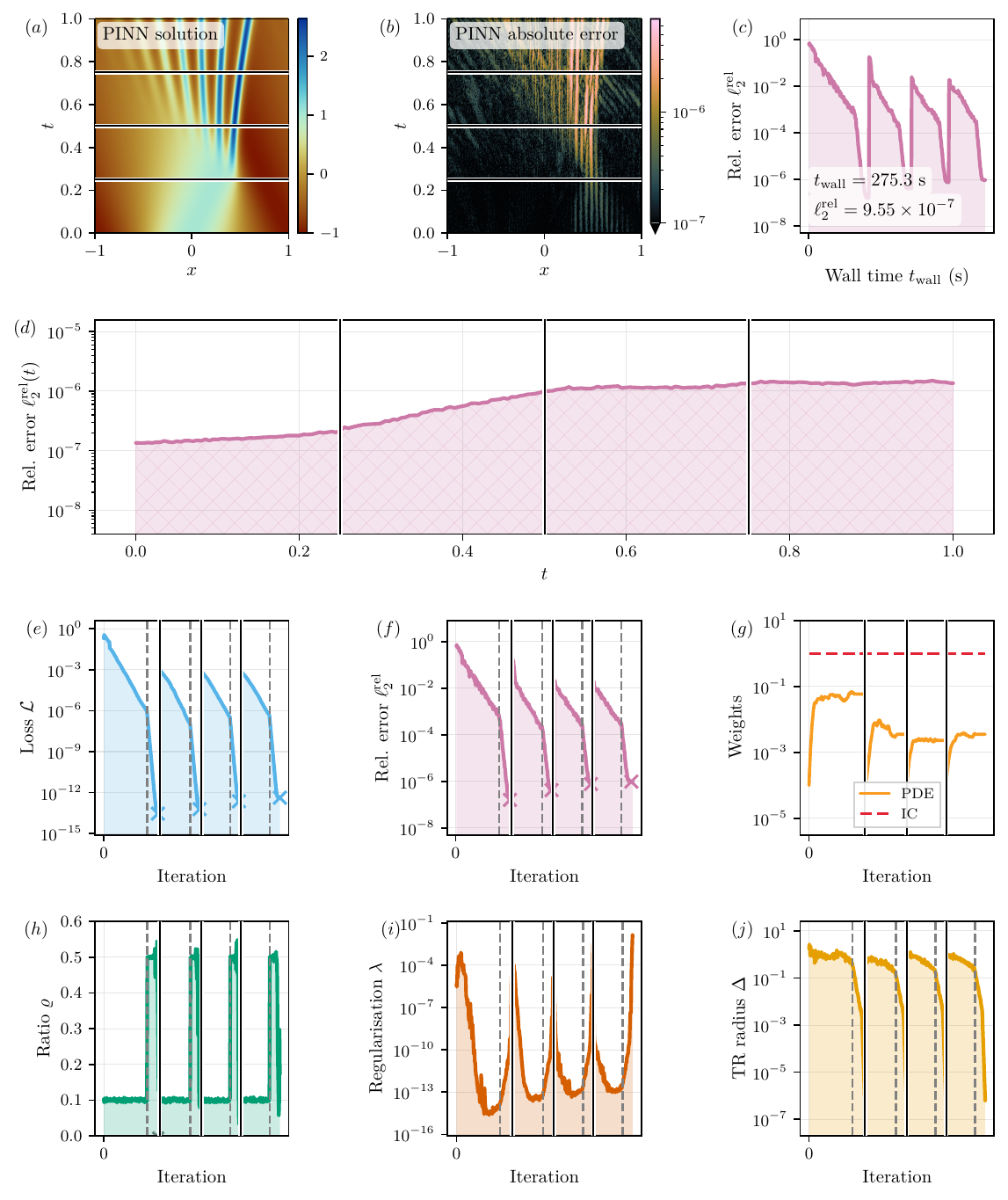}
    \captionof{figboxcap}{
  Results and metrics for the solution of the Korteweg--De Vries equation (see \cref{app:KdV_eq}) in single precision solved with four time-steps. For architecture, a SIREN (see \cref{sec:siren}) with hidden dimensions $[40, 40, 40, 40]$ is used ($d_{\bm\theta} = 5{,}121$) with a sketch size of $s = 3000$. For PDE collocation points, we use $N_{\mathcal{\bm{P}}}=2^{15}$ uniformly sampled, and for the initial condition, we use $N_{\mathcal{\bm{I}}}=2^{14}$ linearly spaced. These results prioritise accuracy, whilst computed with remarkable speed, achieving $\ell_2^{\text{rel}} = 9.55 \times 10^{-7}$ in 275.3 seconds.}
\end{figbox}
\end{figure}

\begin{figure}[p]
\vspace*{-6em}
\begin{figbox}{Korteweg--De Vries}{Double precision}
\label{solution:kdv-double}
    \includegraphics[scale=0.9, trim={5 0 5 0}, clip]{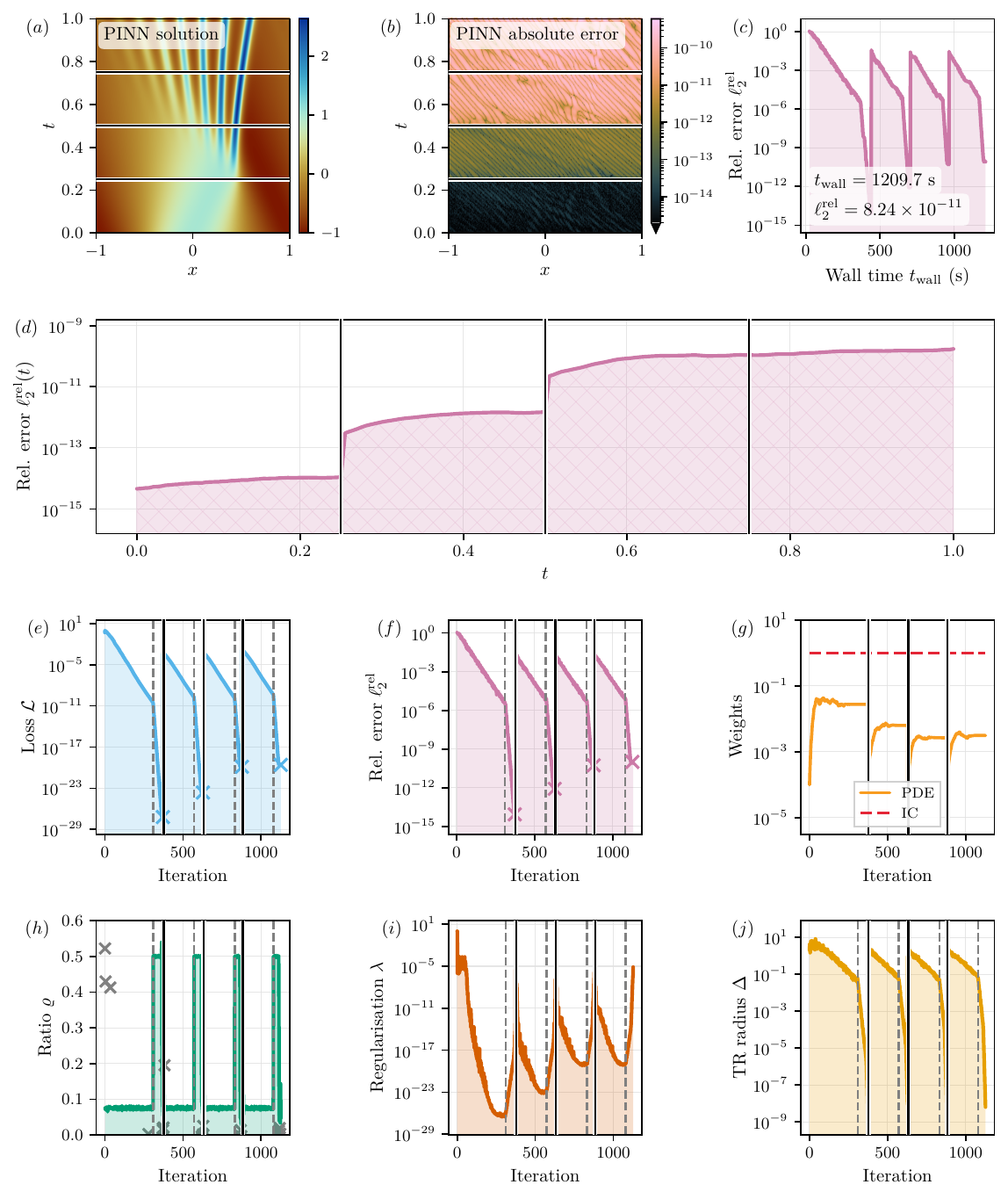}
    \captionof{figboxcap}{
  Results and metrics for the solution of the Korteweg--De Vries equation (see \cref{app:KdV_eq}) in double precision solved with four time-steps. For architecture, a SIREN (see \cref{sec:siren}) with hidden dimensions $[60, 60, 60, 60]$ is used ($d_{\bm\theta} = 11{,}285$) with a sketch size of $s = 4000$. For PDE collocation points, we use $N_{\mathcal{\bm{P}}}=2^{15}$ uniformly sampled, and for the initial condition, we use $N_{\mathcal{\bm{I}}}=2^{14}$ linearly spaced. These results prioritise accuracy, whilst computed with remarkable speed, achieving $\ell_2^{\text{rel}} = 8.24 \times 10^{-11}$ in 1209.7 seconds.}
\end{figbox}
\end{figure}

\begin{figure}[p]
\vspace*{-6em}
\begin{figbox}{Multi-scale Poisson}{Single precision}
\label{solution:multiscale-single}
    \includegraphics[scale=0.9, trim={5 0 5 0}, clip]{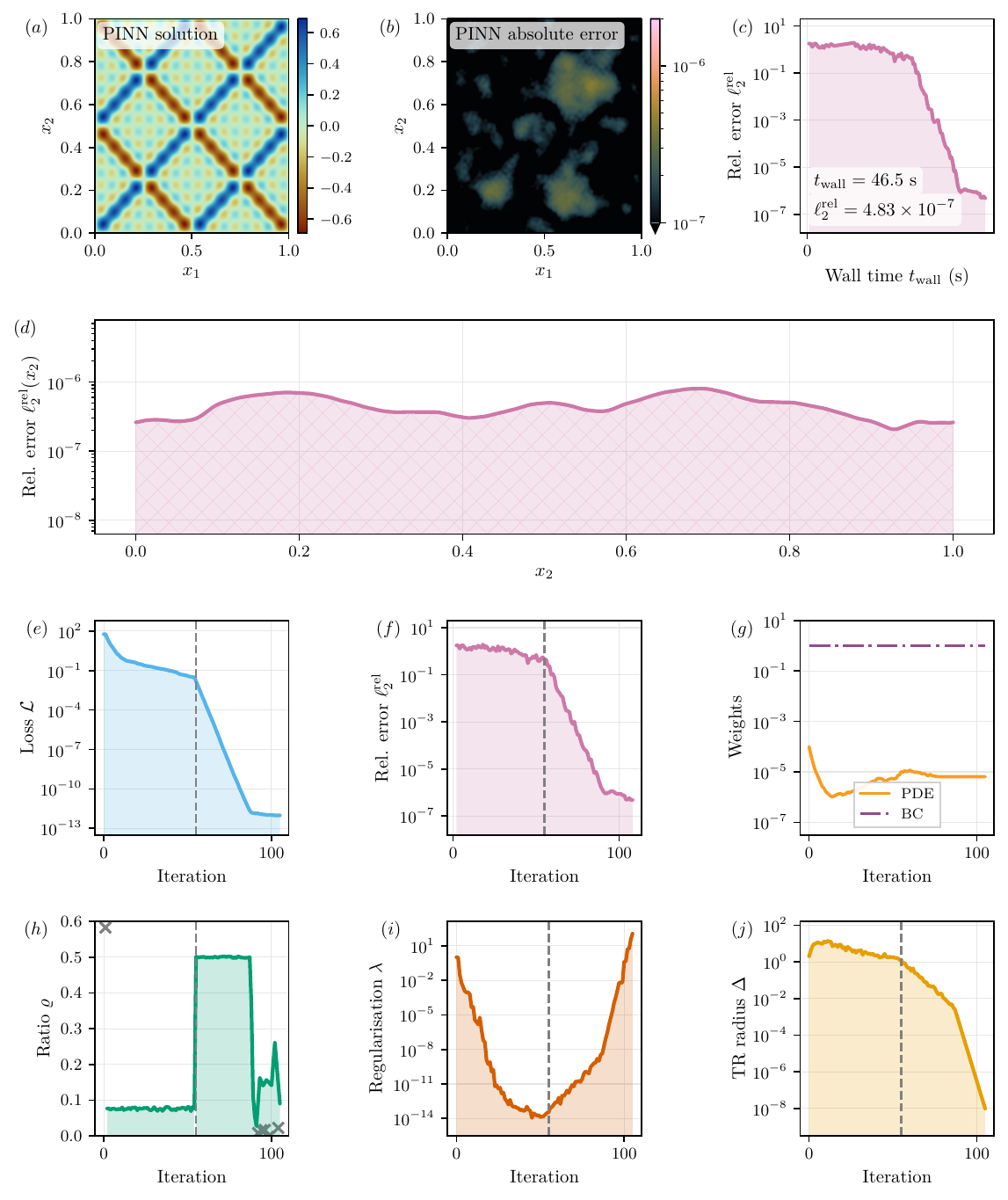}
    \captionof{figboxcap}{
  Results and metrics for the solution of the Multi-scale Poisson equation (see \cref{app:multiscale_eq}) in single precision. For architecture, an MLP with hidden dimensions $[30, 30, 30, 30]$ and three Fourier nodes are utilised (see \cref{sec:fourier}) is used ($d_{\bm\theta} = 3{,}215$) with a sketch size of $s = 1200$. For PDE collocation points, we use $N_{\mathcal{\bm{P}}}=2^{15}$ uniformly sampled, and for the boundary condition, we use $N_{\mathcal{\bm{B}}}=2^{14}$ uniformly sampled. These results prioritise compute time, whilst obtaining near machine error, achieving $\ell_2^{\text{rel}} = 4.83 \times 10^{-7}$ in 46.5 seconds.}
\end{figbox}
\end{figure}

\begin{figure}[p]
\vspace*{-6em}
\begin{figbox}{Multi-scale Poisson}{Double precision}
\label{solution:multiscale-double}
    \includegraphics[scale=0.9, trim={5 0 5 0}, clip]{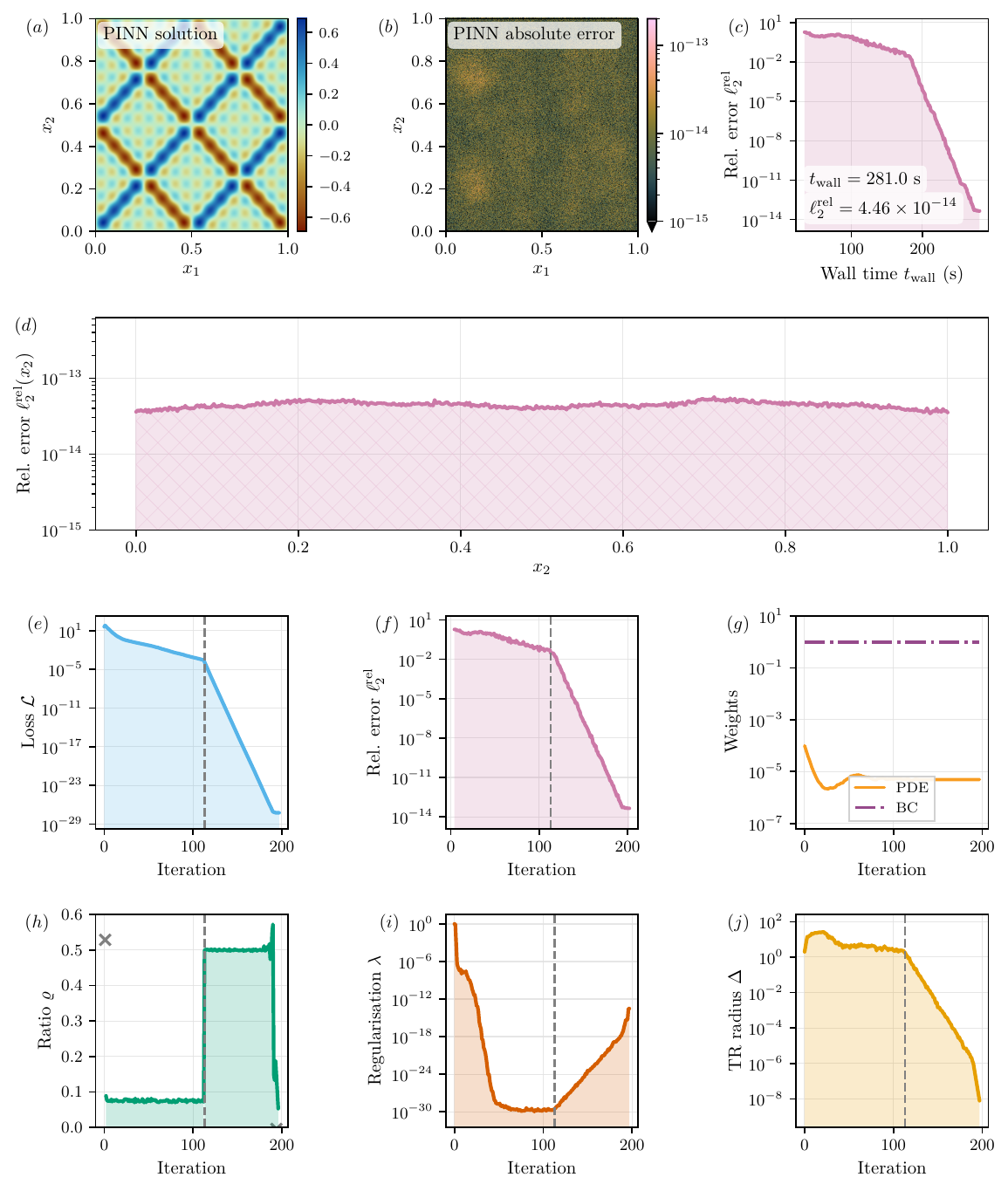}
    \captionof{figboxcap}{
  Results and metrics for the solution of the Multi-scale Poisson equation (see \cref{app:multiscale_eq}) in double precision. For architecture, a SIREN (see \cref{sec:siren}) with hidden dimensions $[60, 60, 60, 60]$ and three Fourier nodes are utilised (see \cref{sec:fourier}) is used ($d_{\bm\theta} = 11{,}825$) with a sketch size of $s = 4000$. For PDE collocation points, we use $N_{\mathcal{\bm{P}}}=2^{15}$ uniformly sampled, and for the boundary condition, we use $N_{\mathcal{\bm{B}}}=2^{14}$ uniformly sampled. These results prioritise accuracy, whilst computed with remarkable speed, achieving $\ell_2^{\text{rel}} = 4.46 \times 10^{-14}$ in 281.0 seconds.}
\end{figbox}
\end{figure}

\begin{figure}[p]
\vspace*{-6em}
\begin{figbox}{Multi-scale Poisson with different parameters}{Double precision}
\label{solution:multiscale-extra}
    \includegraphics[scale=0.9, trim={5 0 5 0}, clip]{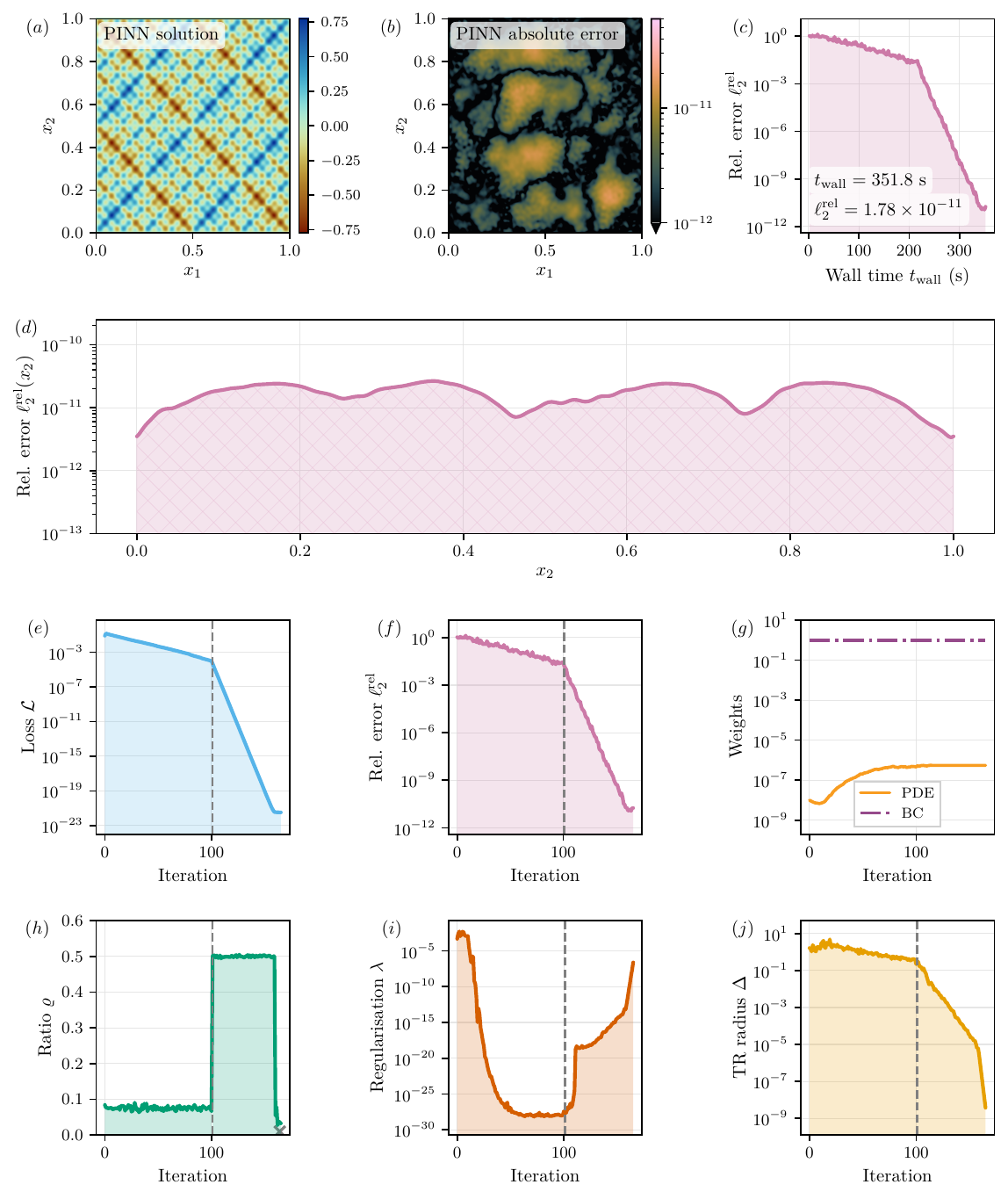}
    \captionof{figboxcap}{
  Results and metrics for the solution of the Multi-scale Poisson equation where $\mathcal{W} = \{4, 8, 16, 32\}$ (see \cref{app:multiscale_eq}) in double precision. For architecture, a SIREN (see \cref{sec:siren}) with hidden dimensions $[60, 60, 60, 60]$ and four Fourier nodes are utilised (see \cref{sec:fourier}) is used ($d_{\bm\theta} = 12{,}065$) with a sketch size of $s = 4000$. For PDE collocation points, we use $N_{\mathcal{\bm{P}}}=2^{15}$ uniformly sampled, and for the boundary condition, we use $N_{\mathcal{\bm{B}}}=2^{14}$ uniformly sampled. These results prioritise accuracy, whilst computed with remarkable speed, achieving $\ell_2^{\text{rel}} = 1.78 \times 10^{-11}$ in 351.8 seconds.}
\end{figbox}
\end{figure}

\begin{figure}[p]
\vspace*{-6em}
\begin{figbox}{Poisson in 5D}{Double precision}
\label{solution:poisson5-double}
    \includegraphics[scale=0.9, trim={5 0 5 0}, clip]{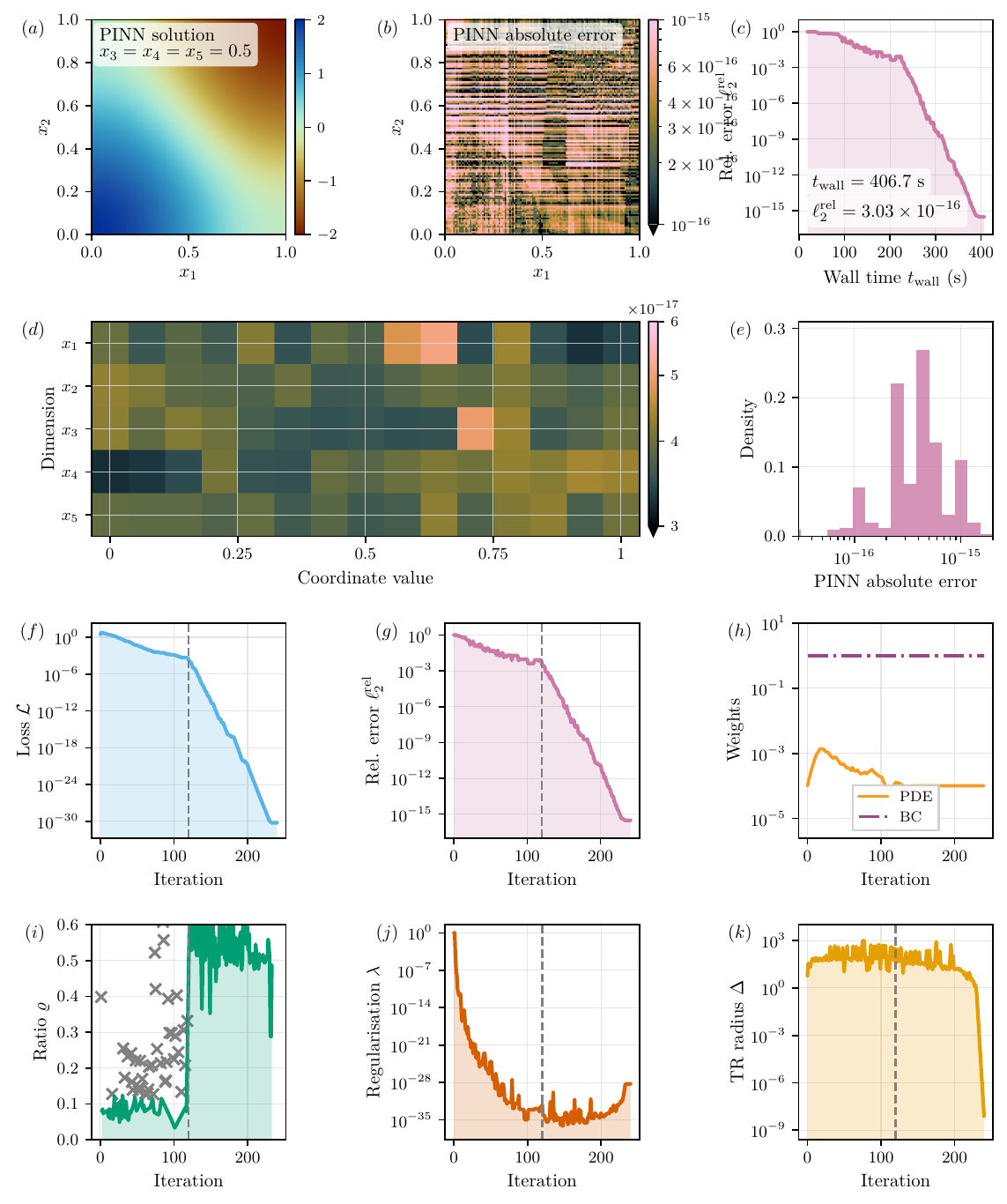}
    \captionof{figboxcap}{
  Results and metrics for the solution of the 5D Poisson equation (see \cref{app:5dpoisson_eq}) in double precision. For architecture, a GaborNet (see \cref{sec:gabor}) with $2^{10}$ features is used ($d_{\bm\theta} = 13{,}313$) with a sketch size of $s = 4000$. For PDE collocation points, we use $N_{\mathcal{\bm{P}}}=2^{15}$ uniformly sampled, and for the boundary condition, we use $N_{\mathcal{\bm{B}}}=2^{14}$ uniformly sampled. These results prioritise accuracy, whilst computed with remarkable speed, achieving $\ell_2^{\text{rel}} = 3.03 \times 10^{-16}$ in 406.7 seconds.}
\end{figbox}
\end{figure}

\end{document}